\documentclass[10pt,twocolumn,letterpaper]{article}

\usepackage{iccv}
\usepackage{times}
\usepackage{epsfig}
\usepackage{graphicx}
\usepackage{amsmath}
\usepackage{amssymb}
\usepackage{booktabs}
\usepackage{epsfig}
\usepackage{multirow}
\usepackage{dsfont}
\usepackage{cite}
\usepackage{float}
\usepackage{color, colortbl, xcolor}
\usepackage{soul}
\usepackage{float}
\usepackage{url}
\usepackage{afterpage}
\usepackage{rotate}
\usepackage{tikz}
\usepackage{makecell}
\usepackage{multirow}
\usepackage{algorithm}
\usepackage{algorithmicx}
\usepackage{algpseudocode}
\usepackage{mathtools}
\usepackage{etoolbox}
\usepackage{lipsum}
\usepackage{bbm}
\usepackage{listings} 
\usepackage{tabularx}
\usepackage{amsbsy}
\usepackage{pifont} 
\usepackage{subcaption}
\usepackage{bbold}
\usepackage{xcolor}
\usepackage{array}
\usepackage{bbding}
\usepackage{color, colortbl}
\usepackage{caption,setspace}
\usepackage{mathrsfs}
\usepackage{hhline}      

\usepackage[normalem]{ulem}

\algdef{SE}[SUBALG]{Indent}{EndIndent}{}{\algorithmicend\ }%
\algtext*{Indent}
\algtext*{EndIndent}
\usetikzlibrary{shadows}
\usetikzlibrary{plotmarks}
\usetikzlibrary{shapes}

\newcolumntype{Y}{>{\centering\arraybackslash}X}
\algnewcommand{\nComment}[1]{\Statex \Comment{#1}}
\definecolor{LightCyan}{rgb}{0.88,1,1}

\makeatletter
\newcommand{\thickhline}{%
	\noalign {\ifnum 0=`}\fi \hrule height 1pt
	\futurelet \reserved@a \@xhline
}

\usepackage[breaklinks=true,colorlinks,bookmarks=false]{hyperref}
\usepackage{cleveref}  
\iccvfinalcopy 

\def\iccvPaperID{4147} 

\ificcvfinal\fi

\crefname{section}{Sec.}{Secs.}
\Crefname{section}{Section}{Sections}
\Crefname{table}{Table}{Tables}
\crefname{table}{Tab.}{Tabs.}
\crefname{figure}{Fig.}{Figs.}
\newcommand*\circled[1]{\tikz[baseline=(char.base)]{
            \node[shape=circle,draw,inner sep=.5pt] (char) {#1};}}
\definecolor{darkcyan}{rgb}{0,.79,.75}

\definecolor{city_inlier}{rgb}{0,0,1}
\definecolor{city_outlier}{rgb}{0, 0.75, 0.95}

\definecolor{ood_inlier}{rgb}{0.1, 0.54, 0.1}
\definecolor{ood_outlier}{rgb}{0.5, 0.8, 0.5}

\newcommand{\mycircle}[1]{\tikz{\node[draw=#1, color=white, fill=#1, circle,minimum
width=0.25cm,minimum height=0.25cm,inner sep=0pt, circular drop shadow, very thick] at (0,0) {};}}

\newcommand{\mytriangle}[1]{\tikz{\node[draw=#1, color=white, fill=#1,isosceles
triangle,isosceles triangle stretches,shape border rotate=90,minimum
width=0.25cm,minimum height=0.25cm,inner sep=0pt, drop shadow, very thick] at (0,0) {};}}

\newcommand{\OodCoco}{\protect\mytriangle{ood_outlier}}
\newcommand{\InlierCoco}{\protect\mycircle{ood_inlier}}
\newcommand{\OodCity}{\protect\mytriangle{city_outlier}}
\newcommand{\InlierCity}{\protect\mycircle{city_inlier}}
\begin{document}
\title{Residual Pattern Learning for Pixel-wise Out-of-Distribution Detection in Semantic Segmentation}



\author{
\parbox{\linewidth}{\centering Yuyuan Liu\textsuperscript{\rm 1 $\dagger$} $\quad$ Choubo Ding\textsuperscript{\rm 1 $\dagger$} $\quad$ Yu Tian\textsuperscript{\rm 2} $\quad$   Guansong Pang\textsuperscript{\rm 3} $\quad$  Vasileios Belagiannis\textsuperscript{\rm 4} \\  $\quad$ Ian Reid\textsuperscript{\rm 1} $\quad$  Gustavo Carneiro\textsuperscript{\rm 5, 1} \\   \vspace{10pt}
\small \textsuperscript{\rm 1} Australian Institute for Machine Learning, University of Adelaide 
 $\quad$  \textsuperscript{\rm 2} Harvard University \\
\textsuperscript{\rm 3} Singapore Management University  $\quad$  
\textsuperscript{\rm 4} Friedrich-Alexander-Universität Erlangen-Nürnberg $\quad$  
\textsuperscript{\rm 5} University of Surrey \\ \vspace{3pt}
\url{https://github.com/yyliu01/RPL}}
}
\twocolumn[{%
\renewcommand\twocolumn[1][]{#1}%
\maketitle
\begin{center}
    \centering
    \captionsetup{type=figure}
    \vspace{-10pt}
    \includegraphics[width=\textwidth]{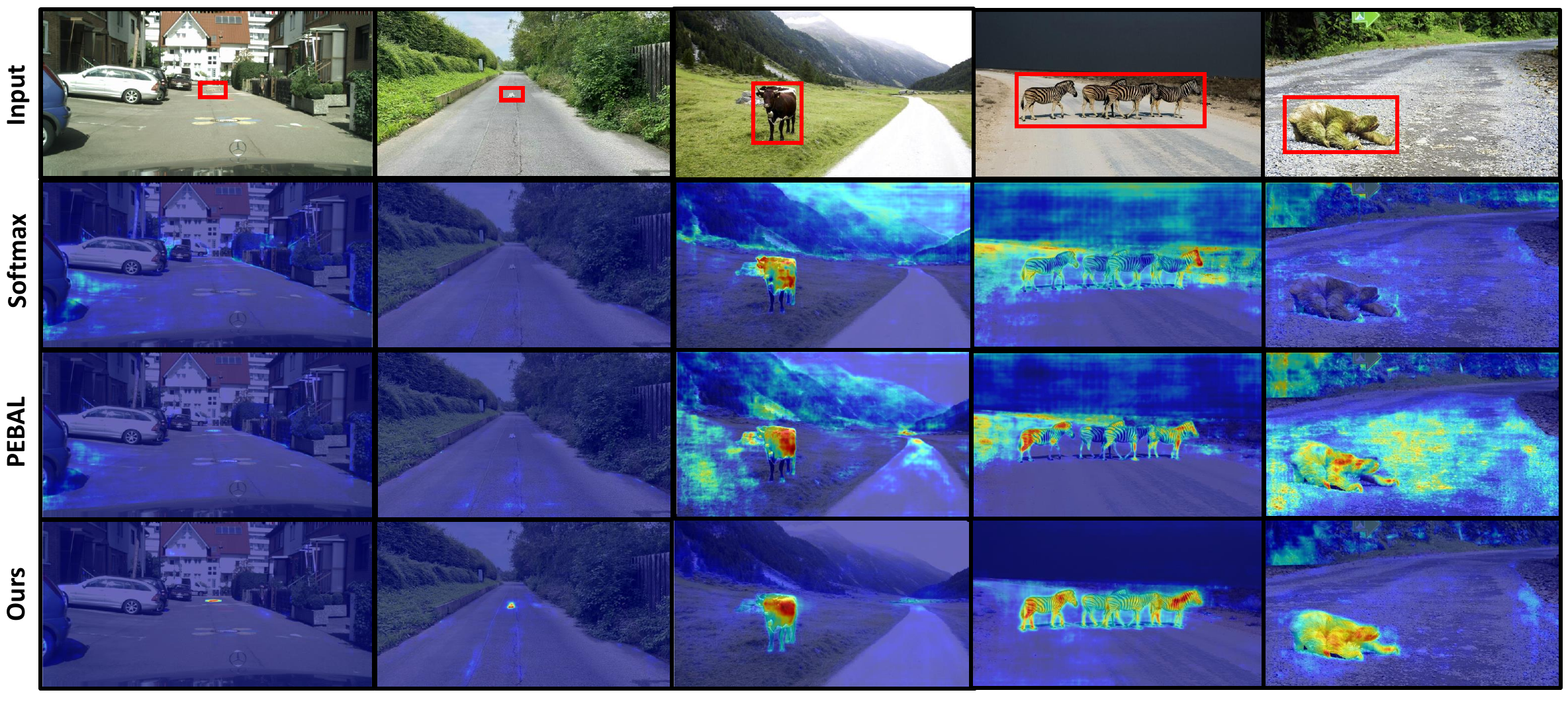}
    \captionof{figure}{\textbf{Visualisations for pixel-wise anomaly scores}. 
    The top row shows real-world images along with \textcolor{red}{bounding boxes} around anomalous objects with respect to the training categories. Subsequent rows show the anomaly score map for different methods, namely SoftMax~\cite{hendrycks2016baseline}, PEBAL~\cite{tian2021pixel}, and our approach. Note that our method successfully detect all anomalous objects without incorrectly deeming in-distribution pixels to be anomalous, while the previous SOTA PEBAL~\cite{tian2021pixel} either fails to detect the anomaly (column 2) or mis-detects inliers as anomalies (columns 3-5).
    } 
    \label{fig:first_page} 
\end{center}

}]
\maketitle

\begin{abstract}
\vspace{-10 pt}
Semantic segmentation models classify pixels into a set of known (``in-distribution'') visual classes. When deployed in an open world, the reliability of these models depends on their ability to not only  classify in-distribution pixels but  also to detect out-of-distribution (OoD) pixels. 
Historically, the poor OoD detection performance of these models has motivated the design of methods based on model re-training using synthetic training images that include OoD visual objects.
Although successful, these re-trained methods have two issues: 1) their in-distribution segmentation accuracy may drop during re-training, and 2) their OoD detection accuracy does not generalise well to new contexts outside the training set (e.g., from city to country context). 
In this paper, we mitigate these issues with: (i) a new residual pattern learning (RPL) module that assists the segmentation model to detect OoD pixels with minimal deterioration to inlier segmentation accuracy; and (ii) a novel context-robust contrastive learning (CoroCL) that enforces RPL to robustly detect OoD pixels in various contexts. Our approach improves by around 10\% FPR and 7\% AuPRC previous state-of-the-art in Fishyscapes, Segment-Me-If-You-Can, and RoadAnomaly datasets. 
\vspace{-15pt}
\end{abstract}

\section{Introduction}
\label{sec:intro}

Semantic segmentation is a fundamental computer vision task that classifies each image pixel into a set of in-distribution visual classes (or inliers)~\cite{hao2020brief}. 
Despite its success in closed-set benchmarks, many real-world applications need to be able to cope with open-world scenarios, in which closed-world segmentation models will mistakenly estimate out-of-distribution (OoD) visual objects (or anomalies) as one of the in-distribution classes. For some use-cases, such as autonomous driving, such errors could be catastrophic~\cite{hao2020brief,yurtsever2020survey}.
One possible way to mitigate this risk is based on the development of pixel-wise OoD detection methods that work together with an in-distribution segmentation model to produce an OoD map on top of the closed-set segmentation mask~\cite{jung2021standardized,  mukhoti2018evaluating, hendrycks2016baseline}. 
Generally, this can be achieved by measuring the in-distribution segmentation uncertainty using 
classification entropy~\cite{di2021pixel}, logits~\cite{jung2021standardized} or
posterior distribution~\cite{hendrycks2016baseline, neal2012bayesian} by \textit{freezing} the inlier segmentation model. Even though those approaches do not jeopardize the inlier performance, they tend to produce low uncertainty for hard OoD pixels that share similar patterns with in-distribution objects, leading to unsatisfactory performance when dealing with complex scenes, as shown in~\cref{fig:first_page}, row 2 (labelled as 'Softmax').

Different from the OoD detection methods above, state-of-the-art (SOTA) OoD pixel detectors are based on methods that \textit{re-train} the inlier segmentation models with OoD data~\cite{chan2021entropy,tian2021pixel,grcic2022densehybrid}.
Such OoD data is obtained from Outlier Exposure (OE)~\cite{hendrycks2018deep}, which introduces additional OoD images~\cite{chan2021entropy} or synthetically mixes OoD objects to training images~\cite{tian2021pixel,grcic2022densehybrid}. 
Then, the closed-set model is re-trained to optimise the OE pixels via entropy maximisation~\cite{chan2021entropy} or abstention learning~\cite{tian2021pixel}. 
Such re-training boosts 
improves the pixel-wise OoD detection accuracy, but unlike the approaches that \textit{freeze the inlier segmentation}~\cite{jung2021standardized, hendrycks2016baseline}, it can  worsen the in-distribution segmentation accuracy. 
For example, pixels from minority categories (e.g., ``fence'', ``sidewalks'') can be mis-classified as majority classes (e.g., ``road''). 
\textbf{We argue that the ability to detect anomalies should be achieved with minimal detriment to the closed-set segmentation accuracy.} 
Another critical issue that affects OoD pixel detectors is their narrow context reliability~\cite{tian2021pixel,chan2021entropy,grcic2022densehybrid,di2021pixel,jung2021standardized},
where context is defined by the in-distribution pixels that represent the scene surroundings, 
such as city (column 1) and country (columns 2-5) contexts in~\cref{fig:first_page}. 
This is because entropy or energy-based optimisation algorithms~\cite{chan2021entropy,di2021pixel,tian2021pixel} train pixel-wise anomaly scores independently, but ignore the relationships between OoD and contexts, leading to over-confident minimisation of inlier uncertainty and a weak maximisation of outlier pixels.
Therefore, an anomaly detector can incorrectly classify inliers with unfamiliar patterns as outliers in new contexts, or misclassify OoD pixels within small anomalies as inliers.  
For example, although the SOTA PEBAL~\cite{tian2021pixel} (\cref{fig:first_page}, row 3) barely detects the tiny anomaly in the city context of column 1, it ignores the outlier in column 2 and mis-detects anomalies in many inliers in columns 3-5, when the context changes from city to country.  
Given that such context changes are common and unpredictable in real-world scenarios,
\textbf{it is important to develop a pixel-wise OoD detector that can generalise to various contexts.} 

In this paper, we address the points above with a new pixel-wise OoD detection module, called Residual Pattern Learning (RPL), which consists of an external module 
attached to a \textbf{frozen} closed-set segmentation network. 
RPL is trained to learn the residual pattern of anomalies based on intermediate features
and induce the (frozen) segmentation classifier to produce high uncertainty for the potentially anomalous regions.
By not \textit{re-training} the segmentation model, RPL guarantees high accuracy in detecting OoD pixels while minimising the impact on the closed-set segmentation performance.
The effective training of RPL depends on our proposed positive energy loss that focuses only on the energy score of anomalies to deal 
with the imbalanced distribution of inlier and outlier samples in pixel-wise OoD detection.
Moreover, we address open-world context robustness with our new context-robust contrastive learning (CoroCL), which optimises the pixel-wise embeddings by exploring the relationship between anomalies and inliers in different contexts.
To summarise, our contributions are:
\begin{enumerate}
    \item The residual pattern learning (RPL) module that induces the closed-set segmentation model to detect potential anomalies, which is a novel perspective for pixel-wise OoD detection; 
    \item A context-robust contrastive learning (CoroCL) that is designed to generalise the detection of OoD pixels 
    to new contexts (see~\cref{fig:first_page}, row 4); and 
    \item A novel positive energy loss function that deals with the imbalanced distribution of inliers and outliers by focusing only on the energy score of anomalies.
\end{enumerate}
Our approach is shown to be the most  accurate method in various scene contexts, yielding improvements of around  10\% FPR and 7\% AuPRC over the SOTA results in Fishyscapes, Segment-Me-If-You-Can, and RoadAnomaly datasets. Our approach is also shown to be the most stable method in various scene contexts and to be easily integrated to other SOTAs, such as Meta-OoD~\cite{chan2021entropy} and PEBAL~\cite{tian2021pixel}. 





\section{Related Work}
\label{sec:related_work}

\noindent \textbf{Semantic segmentation} refers to pixel-wise classification methods that are useful for applications like autonomous driving~\cite{hao2020brief,yurtsever2020survey} or scene understanding~\cite{hong2022goss}.
Influenced by fully convolutional network (FCN)~\cite{long2015fully}, current approaches provide better segmentation details by maintaining the high-level image representations~\cite{yuan2020object, ronneberger2015u, krevso2020efficient}, or construct a feature pyramid to merge the multi-scale representations to learn global image contexts~\cite{lin2017feature, zhao2017pyramid}.
DeepLab methods~\cite{chen2017deeplab,chen2017rethinking,chen2018encoder} utilise dilated convolution to filter the incoming features and enlarge the receptive field. 
Our model is based on  DeepLabV3+~\cite{chen2018encoder} to enable a fair comparison with recently proposed anomaly detection approaches.

\noindent\textbf{Pixel-wise OoD Detection}
methods are based on approaches that
\textit{freeze}~\cite{jung2021standardized,di2021pixel} or \textit{re-train} the segmentation model~\cite{tian2021pixel,chan2021entropy,grcic2022densehybrid}. 
Most of the early anomaly detection approaches~\cite{mukhoti2018evaluating} freeze the segmentation model, and OoD pixels are classified based on posterior distribution measures (e.g., entropy)~\cite{hendrycks2016baseline, lakshminarayanan2017simple}  or on distance measures (e.g., Mahalanobis)~\cite{lee2018simple}. Those methods have low computational cost and do not affect the original inlier segmentation accuracy, but show inaccurate OoD detection. Recently, Synboost~\cite{di2021pixel} improved OoD detection accuracy with an extra FCN network to classify anomalies based on the segmentation model's output and a generative model~\cite{wang2018high}. Unfortunately, the incorrect information from segmentation output introduces confirmation bias to the training process, leading to poor generalisation. 
Instead of working with the segmentation model's output, our RPL module is integrated with the segmentation network. 
Given the richer information content of  intermediate features, compared with the output logit~\cite{di2021pixel} or classification probabilities~\cite{mukhoti2018evaluating}, our approach can detect potential anomalies more effectively. 

Promising results are shown by approaches that re-train the segmentation model~\cite{hendrycks2018deep}. 
Meta-OoD~\cite{chan2021entropy} introduces OoD images from an outlier dataset (e.g., COCO~\cite{lin2014microsoft}) to the inlier training set to boost the entropy results in outlier regions. To better simulate anomalies, PEBAL~\cite{tian2021pixel}, Densehybrid~\cite{grcic2022densehybrid} and~\cite{bevandic2019simultaneous} randomly crop anomalous objects from the outlier exposure (OE) datasets into inlier images, where PEBAL achieves SOTA performance with pixel-wise abstention learning~\cite{liu2019deep} and energy regularisation~\cite{liu2020energy}. However, these re-training approaches can degrade the closed-set segmentation accuracy, which motivated us to introduce a method 
that produces pixel-wise anomalous scores with minimal deterioration to the inlier segmentation performance.

\noindent \textbf{Supervised Contrastive Learning}~\cite{khosla2020supervised, wang2021exploring,li2022targeted} has shown promising results for classification tasks by leveraging labelled data to learn well structured feature representations. These methods build an additional projector to extract image embeddings and push apart  embedding pairs from different classes while pulling together same class pairs. Such optimisation allows the model to explore the relationship among the input representations, instead of only relying on individual samples' classification results for training~\cite{wang2021exploring,tian2020makes}. 
Motivated by this, our proposed Context-robust Contrastive Learning (CoroCL) aims to produce consistent representations for RPL by understanding the relationship between the pixel-wise contexts and anomaly embeddings.

\section{Methodology}
\label{sec:methodology}

\begin{figure*}[t!]
    \centering
    \includegraphics[width=\textwidth]{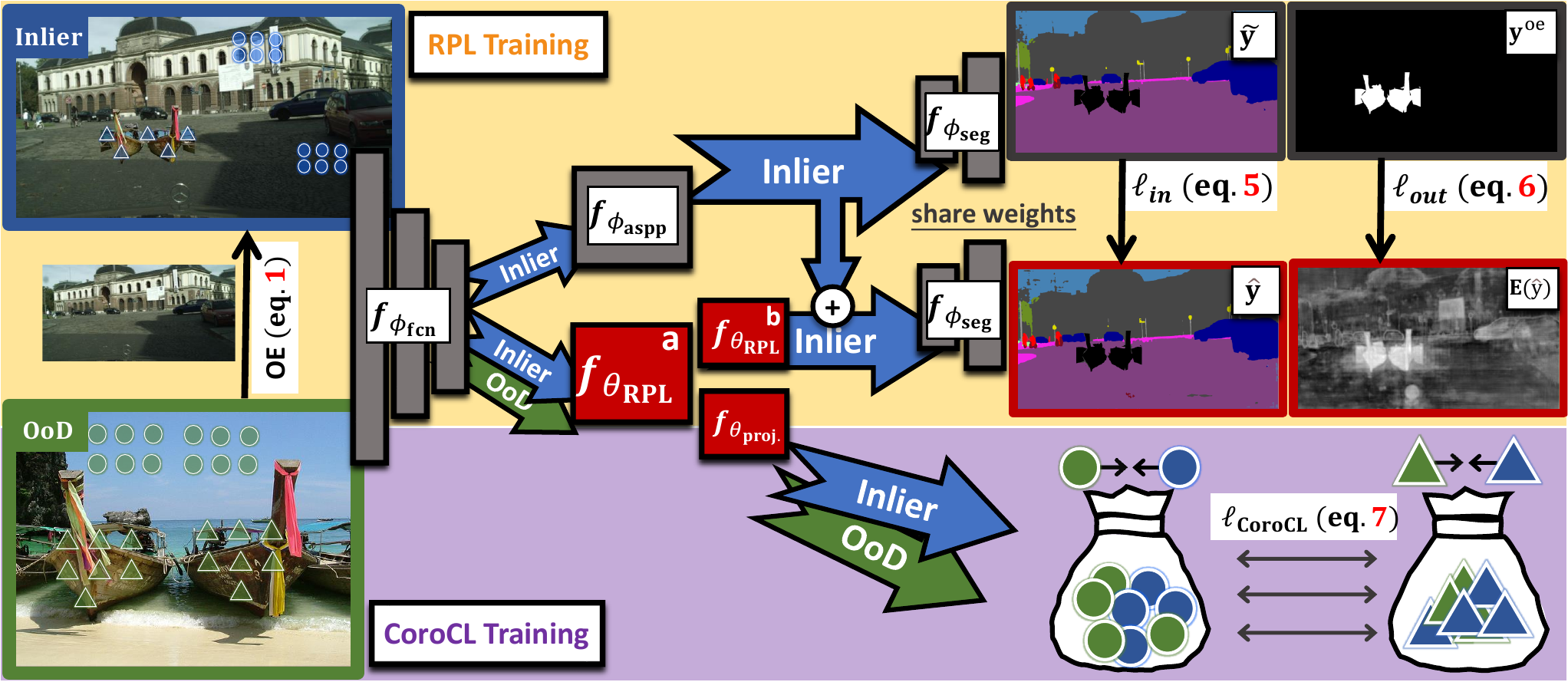}
    \caption{
    The \textbf{RPL module is trained} to approximate the inlier segmentation $\tilde{\mathbf{y}}$ with  $\hat{\mathbf{y}}=f_{\phi_{\text{seg}}}(.)$, which uses as input the model's intermediate features and RPL's output, and to find the OoD pixels in  $\mathbf{y}^{oe}$.
    \textbf{The training of CoroCL} 
    pulls together embedding pairs that both belong to in-distribution (\InlierCoco, \InlierCity) or OoD pixels (\OodCoco, \OodCity), while pushing in-distribution embeddings apart from OoD embeddings.
    }
    \label{fig:draft}
    \vspace{-15 pt}
\end{figure*}

Before we describe our approach and the training process, we introduce our dataset for the pixel-wise OoD detection. We assume to have the inlier dataset (e.g., Cityscapes~\cite{cordts2016cityscapes}) with $\mathcal{D}^{in} = \{(\mathbf{x}^{in}_i, \mathbf{y}^{in}_i)\}_i^{|\mathcal{D}^{in}|}$, where $\mathbf{x}^{in} \in \mathcal{X} \subset \mathbb{R}^{H\times W \times 3}$ is the input image with resolution $H\times W$, and $\mathbf{y}^{in} \in \mathcal{Y}^{in} \subset \{1,...,C\}^{H\times W}$ is the segmentation map, with $C$ closed-set categories. We also have an outlier dataset (e.g., COCO~\cite{lin2014microsoft}) with $\mathcal{D}^{out} = \{(\mathbf{x}^{out}_i, \mathbf{y}^{out}_i)\}_i^{|\mathcal{D}^{out}|}$, where
$\mathbf{x}^{out} \in \mathcal{X}$, and 
$\mathbf{y}^{out} \in \mathcal{Y}^{out} \subset \{0,P\}^{H\times W}$ is a binary mask distinguishing foreground and background objects, where $P>C$ to ensure the OoD are distinguishable from the inlier categories. Following previous anomaly detection approaches~\cite{tian2021pixel, grcic2022densehybrid}, we define the Outlier Exposure (OE) that grabs anomalous regions in $\mathcal{D}^{out}$ and copies them to the inlier images from $\mathcal{D}^{in}$ during training without any overlap policies, as follows:
 \begin{equation}
 \begin{aligned}
   \mathbf{x}^{oe} &\ =\ (1-\mathbf{m})\odot\mathbf{x}^{in} + \mathbf{m} \odot \mathbf{x}^{out} \\
   \mathbf{y}^{oe} &\ =\ (1-\mathbf{m})\odot\mathbf{y}^{in} + \mathbf{m} \odot \mathbf{y}^{out},
 \end{aligned}
 \label{eq:define_oe}
 \end{equation}
 where 
 $\mathbf{m}=\mathbb{I}(\mathbf{y}^{out}=P)$ ($\mathbb{I}(.)$ is an indicator function) is a binary map that is equal to $1$ for the outlier pixels 
 , but equal to $0$, otherwise. 
This allows us to define the $OE$ (mix-content) dataset with $\mathcal{D}^{oe}=\{(\mathbf{x}^{oe}_i, \mathbf{y}^{oe}_i, \mathbf{m}_i)\}_{i=1}^{|\mathcal{D}^{oe}|}$. 

\subsection{Residual Pattern Learning (RPL) for Pixel-wise Anomaly Objects}


As depicted in~\cref{fig:draft}, the original segmentation model consists of a fully convolutional network (e.g., ResNet~\cite{he2016deep}) denoted by a backbone $f_{\phi_{\text{fcn}}}:\mathcal{X} \to \mathcal{Z}$, which transforms the image to a representation 
$\mathbf{z} \in \mathcal{Z} \subset \mathbb{R}^{H'\times W \times C'}$, 
and the feature extraction block (e.g., ASPP~\cite{chen2018encoder}) defined by $f_{\phi_{\text{aspp}}}:\mathcal{Z} \to \mathcal{K}$ embeds $\mathbf{z}$ into the space of $\mathcal{K} \subset \mathbb{R}^{K}$.
Then, the segmentation head $f_{\phi_{\text{seg}}}:\mathcal{K} \to [0,1]^{H \times W \times C}$ maps from $\mathcal{K}$ to $C$ segmentation channels (more details are shown in Supplementary Sec.\textcolor{red}{A}), producing the inlier segmentation:
\begin{equation}
  \tilde{\mathbf{y}} = f_{\phi_{\text{seg}}}(f_{\phi_{\text{aspp}}}(f_{\phi_{\text{fcn}}}(\mathbf{x}))).
  \label{eq:inlier}
\end{equation}

\begin{figure}[t!]
    \centering
    \begin{subfigure}[b]{0.23\textwidth}
         \centering
         \includegraphics[width=\textwidth]{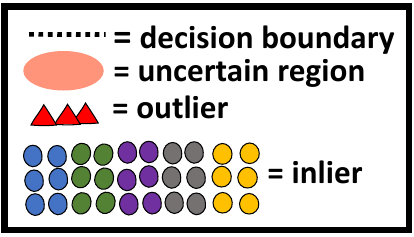}
         \caption*{}
     \end{subfigure} 
     \hfill
    \begin{subfigure}[b]{0.23\textwidth}
         \centering
         \includegraphics[width=\textwidth]{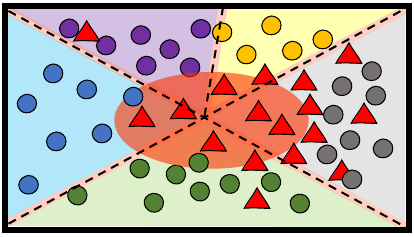}
         \caption{}
         \label{fig:freeze_sketch}
     \end{subfigure} 
     \hfill
    \centering
         \begin{subfigure}[b]{0.23\textwidth}
         \centering
         \includegraphics[width=\textwidth]{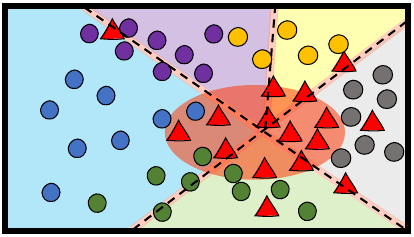}
         \caption{}
         \label{fig:re-train_sketch}
     \end{subfigure}
     \hfill
    \centering
         \begin{subfigure}[b]{0.23\textwidth}
         \centering
         \includegraphics[width=\textwidth]{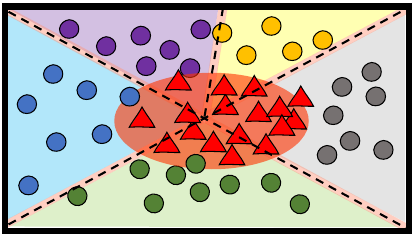}
         \caption{}
         \label{fig:ours_sketch}
     \end{subfigure}
     
    \vspace{-10 pt}
    \caption{Sketch anomaly detection models. \textbf{(a)} Approaches~\cite{jung2021standardized,di2021pixel} that \textit{freeze the inlier segmentation model} maintain inlier categories segmentation accuracy,
    but fail when hard OoD pixels share patterns with inliers. \textbf{(b)} \textit{Re-training} approaches~\cite{tian2021pixel,chan2021entropy,grcic2022densehybrid} fine-tune inlier decision boundaries to detect anomalies but can mis-classify inliers. \textbf{(c)} Our RPL pushes the OoD pixels toward the uncertainty region by minimising the impact to the inlier decision boundary. Note that the inliers inside the uncertain region are the false positive anomalies detected by OoD methods.}
    \label{fig:sketch}
    \vspace{-10 pt}
\end{figure}

\textit{Re-training} approaches~\cite{tian2021pixel,grcic2022densehybrid,chan2021entropy} optimise the segmentation network to recognise the anomaly patterns via
the entropy~\cite{grcic2022densehybrid, chan2021entropy} or energy~\cite{tian2021pixel} from $f_{\phi_{\text{seg}}}(.)$, as depicted in~\cref{fig:re-train_sketch}. Such re-training yields promising results by boosting anomaly uncertainty, but they also shift the inlier's decision boundaries and produce mis-classified inlier predictions. 
On the other hand, as depicted in~\cref{fig:freeze_sketch}, approaches that \textit{freeze the inlier segmentation}~\cite{di2021pixel,jung2021standardized,hendrycks2016baseline,mukhoti2018evaluating} and use segmentation output information (e.g., logits~\cite{jung2021standardized}, prediction~\cite{hendrycks2016baseline} or entropy~\cite{di2021pixel}) have shown poor performance when dealing with hard outliers. 
Our RPL in Fig.~\ref{fig:ours_sketch} is designed to address the issues above by placing an external residual module, denoted by
$f_{\theta_{\text{rpl}}}:\mathcal{Z} \to \mathcal{K}$, between the FCN layers $f_{\phi_{\text{fcn}}}(.)$ and the segmentation head $f_{\phi_{\text{seg}}}(.)$. 
Note that the false positive anomaly detection (i.e., inliers located within the uncertain regions) is a side effect of the production of OoD masks that affects all OoD detectors and is orthogonal to the problem of inlier segmentation accuracy.

The RPL result is added to the ASPP output and decoded by the segmentation head with 
\begin{equation}
    \hat{\mathbf{y}}=f_{\phi_{\text{seg}}}\Big(f_{\phi_{\text{aspp}}}\big(f_{\phi_{\text{fcn}}}(\mathbf{x})\big)+f_{\theta_{\text{rpl}}}\big(f_{\phi_{\text{fcn}}}(\mathbf{x})\big)\Big).
    \label{eq:outlier}
\end{equation}
Our training differs from previous SOTA methods given that we \textbf{freeze the entire segmentation model} (i.e., $\phi_{\text{fcn}}, \phi_{\text{aspp}}$ and $\phi_{\text{seg}}$) during the training that only optimises the RPL block $\theta_{\text{rpl}}$. Therefore, the decision boundary for in-distribution categories is always fixed, so the accuracy of the close-set semantic segmentation prediction, i.e.,  $\tilde{\mathbf{y}}$ from~\eqref{eq:inlier}, will suffer minimal degradation.

\subsection{RPL Training}
\label{sec:training process}

 
Intuitively, RPL is trained to approximate the inlier pixel classification from the segmentation model and at the same time to detect OoD pixels. 
 Hence, the training depends on the closed-set segmentation $\tilde{\mathbf{y}}$ from~\eqref{eq:inlier} and on our model segmentation $\hat{\mathbf{y}}$ from~\eqref{eq:outlier} to optimise the following function:
 \begin{equation}
     \begin{aligned}
       \ell_{\text{RPL}}(\mathcal{D}^{oe}, \theta_{\text{rpl}}) = \ell_{in} (\mathcal{D}^{oe}, \theta_{\text{rpl}}) + \alpha \times \ell_{out} (\mathcal{D}^{oe}, \theta_{\text{rpl}}),
     \end{aligned}
     \label{eq:rpl_all}
 \end{equation}
where $\alpha$ weights the contribution of the second loss function (we set $\alpha=0.05$ following~\cite{lis2019detecting} for all the experiments). 
The first term in~\eqref{eq:rpl_all} is the loss function for RPL to recognise the inlier features, as defined below:
\begin{equation}
\resizebox{\hsize}{!}{$
 \begin{aligned}
   \ell_{in}  (\mathcal{D}^{oe}, \theta_{\text{rpl}}) = \sum_{(\mathbf{x}^{oe},\mathbf{y}^{oe},\mathbf{m}) \in \mathcal{D}^{oe}}\sum_{\omega \in \Omega}   (1-\mathbf{m}(\omega)) \Big{(} & \ell_{ce}(\mathbf{\mathsf{OneHot}(\tilde{y}}({\omega})),\mathbf{\hat{y}}({\omega})) + \\
   &\ell_{reg}\big{(}H(\mathbf{\tilde{y}}({\omega})), H(\mathbf{\hat{y}}({\omega}))\big{)}\Big{)},
 \end{aligned}
 $}
\label{eq:l_in}
 \end{equation}
where $\Omega$ is the image lattice of size $H \times W$, $\mathbf{m}$ represents the anomaly binary mask from~\eqref{eq:define_oe} at pixel $\omega$, $\mathsf{OneHot}(\tilde{\mathbf{y}})$ returns the one-hot representation of $\tilde{\mathbf{y}}$, $\ell_{ce}(.)$ denotes the cross-entropy loss, $H$ represents the Shannon entropy, and the dis-similarity regularisation is defined as $\ell_{reg}(a,b)=\|\frac{(a-b)}{t}\|_2^2$. 
The second term in~\eqref{eq:rpl_all} is our proposed positive energy that maximises the energy at OoD pixels, as in:
\begin{equation}
\resizebox{\hsize}{!}{$
    \begin{aligned}
      \ell_{out}(\mathcal{D}^{oe}, \theta_{\text{rpl}}) &\ = \  \sum_{(\mathbf{x}^{oe}_i,\mathbf{y}^{oe}_i,\mathbf{m}_i) \in \mathcal{D}^{oe}} \sum_{\omega \in \Omega} \:  \max(-\mathbf{m}_{i}({\omega}) E(\mathbf{\hat{y}}_i({\omega})), 0),
    \end{aligned}
    $}
\label{eq:l_out}
\end{equation}
where $E(x)=- \log \sum_{i=1}^C \exp{(x_i)}$, and $C$ is the number of inlier categories. 
The positive energy in~\eqref{eq:l_out} has two advantages over previous OoD energy-based methods for detecting OoD pixels~\cite{liu2020energy,tian2021pixel}: 1) the hinge loss in~\cite{liu2020energy,tian2021pixel} depends on two hard-to-tune hyperparameters to minimise the inlier energy and maximise the outlier energy, while our loss in~\eqref{eq:l_out} does not have any hyperparameters; 2) given that there are many more in-distribution than OoD pixels, the hinge loss  in~\cite{liu2020energy,tian2021pixel} has a poor convergence for anomalous pixels, whereas our loss in~\eqref{eq:l_out} mitigates such weak optimisation for outliers by exclusively focusing on them.

\subsection{Training RPL with Context-robust Contrastive Learning (CoroCL)}

Most OoD detection methods fail to distinguish between unfamiliar inlier pixels and potentially OoD pixels when the context changes because those approaches lack understanding of the relationship between contexts and the anomalous objects. 
With the goal of exploring such relationship, 
our proposed CoroCL (as displayed in~\cref{fig:draft}) involves a new contrastive learning loss function and a projection layer, \textbf{built on top of the RPL block},  
represented by $f_{\theta_{\text{proj}}}:\mathcal{K} \to \mathbb{R}^{I \times R}$ 
that takes the output from RPL and project it to a space of $I$ embeddings of $R$ dimensions.
To explain the projection layer, we need to subdivide the RPL model as $\theta_{\text{rpl}} = \{ \theta_{\text{rpl}}^a,\theta_{\text{rpl}}^b\}$, where $\theta_{\text{rpl}}^a$ denotes the RPL main layers and $\theta_{\text{rpl}}^b$ represents the last convolutional layers of $\theta_{\text{rpl}}$, with the proposed projection layer receiving the features from $f_{\theta_{\text{rpl}}^a}(.)$.
CoroCL trains the main layers of the RPL module and projection layer to pull the embeddings belonging to the same category (inliers or OoD) closer together and push away the embeddings from different categories. 
We target the pixel-wise embedding learning~\cite{wang2021dense,wang2021exploring}, but to enable the recognition of anomalies across different contexts, our approach uses the samples via both $OE$ (mix-content) dataset $\mathcal{D}^{oe}$ and the outlier dataset $\mathcal{D}^{out}$.
Let us assume that we have the embeddings from the $OE$ dataset denoted by $\mathcal{R}^{oe}= \{ \mathbf{r}|\mathbf{r}=f_{\theta_{\text{proj}}}(f_{\theta_{\text{rpl}}^a}(f_{\phi_{\text{fcn}}}(\mathbf{x}^{oe}))), (\mathbf{x}^{oe},\mathbf{y}^{oe},\mathbf{m}) \in \mathcal{D}^{oe} \}$ that combines anomalies with the training set domain, 
and the embeddings from the outlier dataset represented by $\mathcal{R}^{out}= \{ \mathbf{r}|\mathbf{r}=f_{\theta_{\text{proj}}}(f_{\theta_{\text{rpl}}^a}(f_{\phi_{\text{fcn}}}(\mathbf{x}^{out}))), (\mathbf{x}^{out},\mathbf{y}^{out}) \in \mathcal{D}^{out} \}$. 
Given that validation images are
mainly based on the driving scenes (e.g., $\mathcal{D}^{\text{in}}$), we use $\mathcal{R}^{oe}$ to 
form the anchor set $\mathcal{A} = \{ \mathbf{r}_i | \mathbf{r}_i \sim \mathbf{r}, \mathbf{r} \in \mathcal{R}^{oe}, \mathbf{r}_i \in \mathbb{R}^{R}  \}$ by random sampling  $\mathcal{R}^{oe}$ to extract a subset of of the embeddings from $\mathbf{r}\in \mathcal{R}^{oe}$ (more details is shown in Supplementary Sec.\textcolor{red}{E}). 
To make the model robust to various contexts, we form the contrastive set for CoroCL with 
$\mathcal{C} = \{ \mathbf{r}_i | \mathbf{r}_i \sim \mathbf{r}, \mathbf{r} \in (\mathcal{R}^{oe} \bigcup \mathcal{R}^{out}), \mathbf{r}_i \in \mathbb{R}^{R}  \}$ by random sampling  $\mathcal{R}^{oe}$ and $\mathcal{R}^{out}$ to extract a subset of the embeddings from $\mathbf{r}\in (\mathcal{R}^{oe} \bigcup \mathcal{R}^{out})$. 
CoroCL is defined by:
\begin{equation}
\scalebox{0.85}{$
    \begin{aligned}
    &\ell_{\text{CoroCL}}(\mathcal{D}^{oe},\mathcal{D}^{out},\theta_{\text{proj}},\theta_{\text{rpl}}^a) = \\  
    & \sum_{\mathbf{r}_i \in \mathcal{A}}
     \sum_{\mathbf{p}\in\mathcal{P}(\mathbf{r}_i)}
     -\log \frac{\exp(\mathbf{r}_i \cdot \mathbf{p}/\tau)}{\exp(\mathbf{r}_i\cdot \mathbf{p}/\tau) +
     \sum_{\mathbf{n} \in \mathcal{N}(\mathbf{r}_i)} \exp(\mathbf{r}_i\cdot \mathbf{n}/\tau)},
    \end{aligned}
    $}
    \label{eq:CoroCL}
\end{equation}
where $\mathcal{P}(\mathbf{r}_i)=\{ \mathbf{p} | \mathbf{p}=\mathbf{r}_j, \mathbf{r}_j \in \mathcal{C}, \mathbf{m}_i = \mathbf{m}_j \}$ is the set of positive embeddings for the anchor embedding, with the class (inlier or outlier) of the anchor denoted by $\mathbf{m}_i$ and of the positive embedding represented by $\mathbf{m}_j$,
$\mathcal{N}(\mathbf{r}_i)=\{ \mathbf{n} | \mathbf{n}=\mathbf{r}_j, \mathbf{r}_j \in \mathcal{C} , \mathbf{m}_i \ne \mathbf{m}_j \}$ is the set of negative embeddings for the anchor embedding, with the class (inlier or outlier) of the anchor denoted by $\mathbf{m}_i$ and of the negative embedding represented by $\mathbf{m}_j$. 

The overall loss function to train the RPL module is:
\begin{equation}
    \begin{aligned}
    \theta_{\text{rpl}}^*,\theta_{\text{proj}}^* = \arg\min_{\theta_{\text{rpl}},\theta_{\text{proj}}} & \ell_{\text{RPL}}(\mathcal{D}^{oe},\theta_{\text{rpl}}) + \\ & \ell_{\text{CoroCL}}(\mathcal{D}^{oe},\mathcal{D}^{out},\theta_{\text{rpl}}^a,\theta_{\text{proj}}),
    \end{aligned}
    \label{eq:overall}
\end{equation}
where $\theta_{\text{RPL}}^a \subset \theta_{\text{RPL}}$ denotes the main layers of $\theta_{\text{RPL}}$.
Note that the loss function~\eqref{eq:overall} is applied only to the RPL module and projection layer instead of the original segmentation model, so that our model is trained to recognise anomaly patterns in various contexts with minimal impact to the performance of closed-set segmentation categories. 

\noindent \textbf{Inference.} 
During inference,
the inlier segmentation map is represented by $\tilde{\mathbf{y}}$ from~\eqref{eq:inlier}, and the OoD pixels are estimated from the energy score $E(.)$ of~\eqref{eq:l_out} on the segmentation map $\hat{\mathbf{y}}$ from~\eqref{eq:outlier}, for each pixel position $\omega$ of a test image $\mathbf{x}$. 
We also apply Gaussian smoothing to produce the final energy map, following  previous works~\cite{tian2021pixel, jung2021standardized}.

 

\section{Experiments}

\begin{table*}[t!]
\caption{\textbf{Comparision with SOTA methods on Fihsyscapes and SMIYC validation sets.} All approaches are based on the \textbf{DeepLabv3+} architecture with WiderResNet38 backbone. Best results are in \textbf{boldface}, and the worst results from previous SOTAs caused by the lack of context robustness are marked in \textcolor{cyan}{cyan}.}
\vspace{-5 pt}
\label{tab:valid}
\centering
\resizebox{\linewidth}{!}{
\begin{tabular}{!{\vrule width 1.5pt}r!{\vrule width 1.5pt}ccc|ccc!{\vrule width 1.5pt}ccc|ccc!{\vrule width 1.5pt}} 
\specialrule{1.5pt}{0pt}{0pt}
\multicolumn{1}{!{\vrule width 1.5pt}c!{\vrule width 1.5pt}}{\multirow{3}{*}{Methods}} & \multicolumn{6}{c!{\vrule width 1.5pt}}{Fishyscapes (validation set)}         & \multicolumn{6}{c!{\vrule width 1.5pt}}{SMIYC~(validation set)}                       \\ 
\cline{2-13}
                         & \multicolumn{3}{c|}{Static} & \multicolumn{3}{c!{\vrule width 1.5pt}}{L\&F} & \multicolumn{3}{c|}{Anomaly} & \multicolumn{3}{c!{\vrule width 1.5pt}}{Obstacle}  \\ 
\cline{2-13}
                         & FPR $\downarrow$  & AuPRC $\uparrow$    & AUROC $\uparrow$      & FPR $\downarrow$     & AuPRC $\uparrow$   & AUROC $\uparrow$ & FPR$\downarrow$  & AuPRC $\uparrow$   & $F1^*$ $\uparrow$        & FPR $\downarrow$   & AuPRC   $\uparrow$ & $F1^*$ $\uparrow$        \\ 
\specialrule{1.5pt}{0pt}{0pt}
Maximum softmax~\cite{hendrycks2016baseline} {\scriptsize \textcolor{lightgray}{[baseline]}}                & 23.31 & 26.77 & 93.14       & 10.36   & 40.34 & 90.82 & 60.2     &   40.4    &  42.6             &  3.8      & 43.4      &   53.7            \\

Mahalanobis~\cite{lee2018simple} {\scriptsize \textcolor{lightgray}{[baseline]}}              & 11.7   & 27.37 & 96.76        & 11.24 & 56.57 & 96.75  & 86.4     &   22.5    &  31.7             &  26.1      & 25.9      &   27.7            \\

SML~\cite{jung2021standardized} {\scriptsize \textcolor{lightgray}{[ICCV'21]}}                & 12.14 & 66.72 & 97.25      &  33.49  & 22.74 & 94.97  &  84.13   & 21.68      & 28.00              & 91.31       & 18.60      & 28.39              \\
Synboost~\cite{di2021pixel} {\scriptsize \textcolor{lightgray}{[CVPR'21]}}                & 25.59 & 66.44 & 95.87       & 31.02   & 60.58 & 96.21 & 30.9     &   68.8    &  65.6             &  2.8      & 81.4      &   73.2            \\
Meta-OoD~\cite{chan2021entropy} {\scriptsize \textcolor{lightgray}{[ICCV'21]}}                & 13.57 & 72.91 & 97.56       & 37.69 \cellcolor{LightCyan}   & 41.31\cellcolor{LightCyan}  & 93.06 \cellcolor{LightCyan}  & 17.43 & 80.13 & 74.3               & 0.41 & 94.14 & 88.4               \\
DenseHybrid~\cite{grcic2022densehybrid} {\scriptsize \textcolor{lightgray}{[ECCV'22]}}      & 4.17  & 76.23 & 99.07         & 5.09  & 69.79 & 99.01      & 52.65 \cellcolor{LightCyan}  & 61.08 \cellcolor{LightCyan}  & 53.72 \cellcolor{LightCyan}             & 0.71 & 89.49 & 81.05              \\
PEBAL~\cite{tian2021pixel}  {\scriptsize \textcolor{lightgray}{[ECCV'22]}}                    & 1.52  & 92.08 & 99.61       & 4.76 & 58.81 & 98.96  & 36.74 & 53.10 & 57.99              & 7.92\cellcolor{LightCyan}  & 10.45\cellcolor{LightCyan}  & 22.10 \cellcolor{LightCyan}               \\ 
\hline
RPL+CoroCL {\scriptsize \textcolor{black}{[Ours]}}                & \textbf{0.85}  & \textbf{92.46} & \textbf{99.73}       & \textbf{2.52}    & \textbf{70.61} & \textbf{99.39} & \textbf{7.18} & \textbf{88.55} & \textbf{82.90}         & \textbf{0.09} & \textbf{96.91} & \textbf{91.75}         \\
\specialrule{1.5pt}{0pt}{0pt}
\end{tabular}}
\vspace{-15 pt}
\end{table*}

\begin{table}[t!]
\centering
\caption{\textbf{Comparison with SOTA on RoadAnomaly validation set.} All approaches are based on the \textbf{DeepLabv3+} architecture and best results are in \textbf{boldface}.}
\vspace{-5 pt}
\label{tab:road}
\resizebox{\linewidth}{!}{
\begin{tabular}{!{\vrule width 1.5pt}r!{\vrule width 1.5pt}c|c|c!{\vrule width 1.5pt}} 
\specialrule{1.5pt}{0pt}{0pt}
\multicolumn{1}{!{\vrule width 1.5pt}c!{\vrule width 1.5pt}}{\multirow{2}{*}{Method}} & \multicolumn{3}{c!{\vrule width 1.5pt}}{RoadAnomaly}  \\ 
\cline{2-4}
                        & FPR $\downarrow$      & AuPRC $\uparrow$ & AuROC $\uparrow$         \\ 
\specialrule{1.5pt}{0pt}{0pt}
Maximum softmax~\cite{hendrycks2016baseline} {\scriptsize \textcolor{lightgray}{[baseline]}}               & 68.15    & 22.38 & 75.12          \\
Gambler~\cite{liu2019deep} {\scriptsize \textcolor{lightgray}{[NIPS'19]}}                & 48.79    & 31.45 & 85.45          \\
SynthCP~\cite{xia2020synthesize}    {\scriptsize \textcolor{lightgray}{[ECCV'20]}}              & 64.69    & 24.86 & 76.08          \\
Synboost~\cite{di2021pixel} {\scriptsize \textcolor{lightgray}{[CVPR'21]}}                & 59.72    & 41.83 & 85.23          \\
SML~\cite{jung2021standardized}{\scriptsize \textcolor{lightgray}{[ICCV'21]}}                      & 49.74    & 25.82 & 81.96          \\ 
GMMSeg~\cite{liang2022gmmseg} {\scriptsize \textcolor{lightgray}{[NIPs'22]}}  &  47.90 &  34.42 & 84.71  \\
PEBAL~\cite{tian2021pixel}{\scriptsize \textcolor{lightgray}{[ECCV'22]}}                   & 44.58    & 45.10 & 87.63          \\
\hline
RPL+CoroCL {\scriptsize \textcolor{black}{[Ours]}}                     & \textbf{17.74}    & \textbf{71.60} & \textbf{95.72}          \\

\specialrule{1.5pt}{0pt}{0pt}
\end{tabular}}
\vspace{-15 pt}
\end{table}

\begin{table*}[htb!]
\caption{\textbf{Comparision with SOTAs} on Fihsyscapes and SMIYC official test benchmarks\textcolor{red}{\protect\footnotemark[1]}$^,$\textcolor{red}{\protect\footnotemark[2]}. All approaches are based on the \textbf{DeepLabv3+} architecture with WiderResNet38 backbone. Best results are in \textbf{boldface}, and the worst results caused by the lack of context robustness are marked in \textcolor{cyan}{cyan}. 
Additional experiments with methods fine-tuned with extra training sets (e.g., Vistas~\cite{neuhold2017mapillary}, Wilddash2~\cite{zendel2018wilddash}) are in Supplementary Sec.\textcolor{red}{B}. * denotes the results reported in the paper.} 
\vspace{-5 pt} 
\label{tab:test}
\centering
\resizebox{\linewidth}{!}{
\begin{tabular}{!{\vrule width 1.5pt}r!{\vrule width 1.5pt}cc|cc!{\vrule width 1.5pt}cc|cc!{\vrule width 1.5pt}cc!{\vrule width 1.5pt}} 
\specialrule{1.5pt}{0pt}{0pt}
\multicolumn{1}{!{\vrule width 1.5pt}c!{\vrule width 1.5pt}}{\multirow{3}{*}{Methods}}  & \multicolumn{4}{c!{\vrule width 1.5pt}}{Fishyscapes (test)}               & \multicolumn{4}{c!{\vrule width 1.5pt}}{SMIYC (test)     }                                                 & \multicolumn{2}{c!{\vrule width 1.5pt}}{\multirow{2}{*}{\textbf{Overall}}}  \\ 
\cline{2-9}
                         & \multicolumn{2}{c|}{Static} & \multicolumn{2}{c!{\vrule width 1.5pt}}{L\&F} & \multicolumn{2}{c|}{Anomaly} & \multicolumn{2}{c!{\vrule width 1.5pt}}{Obstacle} & \multicolumn{2}{c!{\vrule width 1.5pt}}{}                          \\ 
\cline{2-11}
\rule{0pt}{10pt}
                         & FPR $\downarrow$   & AuPRC $\uparrow$                  & FPR $\downarrow$   & AuPRC $\uparrow$              & FPR $\downarrow$   & AuPRC $\uparrow$                   & FPR $\downarrow$   & AuPRC $\uparrow$                                & $\overline{\text{\small FPR}}$ $\downarrow$    & $\overline{\text{\small AuPRC}}$ $\uparrow$                                    \\ 
\specialrule{1.5pt}{0pt}{0pt}
Resynthesis~\cite{lis2019detecting}{\scriptsize \textcolor{lightgray}{[ICCV'19]}}              & 27.13 & 29.6                & 48.05 & 5.70            & 25.93 & 52.28                & 4.70  & 37.71                  & 26.45 & 31.32                              \\
Embedding~\cite{blum2021fishyscapes}{\scriptsize \textcolor{lightgray}{[IJCV'19]}}                & 20.25 & 44.03               & 30.02 & 3.55            & 70.76 & 37.52                & 46.38 & 0.82                  & 41.85 & 21.48                                \\
Synboost~\cite{di2021pixel}{\scriptsize \textcolor{lightgray}{[CVPR'19]}}                 & 18.75 & 72.59               & 15.79 & 43.22           & \cellcolor{LightCyan} 61.86 & \cellcolor{LightCyan} 56.44                & 3.15  & 71.34                    & 24.89 & 60.90                               \\
Meta-OoD~\cite{chan2021entropy}{\scriptsize \textcolor{lightgray}{[ICCV'21]}}                 & 8.55  & 86.55               & 35.14\cellcolor{LightCyan} & 29.96 \cellcolor{LightCyan}          & 15.00 & \textbf{85.47}                & 0.75  & 85.07                  & 14.86 & 71.76                                 \\

DenseHybrid~\cite{grcic2022densehybrid}{\scriptsize \textcolor{lightgray}{[ECCV'22]}}              & 5.51  & 72.27               & 6.18  & 43.90           & 62.25\cellcolor{LightCyan} & 42.05\cellcolor{LightCyan}                & 6.02  & 80.79                 & 19.99      & 59.75       \\
GMMSeg~\cite{liang2022gmmseg}* {\scriptsize \textcolor{lightgray}{[NIPs'22]}}  & 15.96 & 76.02 & 6.61 & \textbf{55.63} & - & - & - & - & - & - \\
PEBAL~\cite{tian2021pixel}{\scriptsize \textcolor{lightgray}{[ECCV'22]}}                    & 1.73~ & 92.38               & 7.58  & 44.17           & 40.82 & 49.14                & 12.68\cellcolor{LightCyan} & 4.98\cellcolor{LightCyan}                       & 15.70 & 47.67                                \\ 
\hline
\rule{0pt}{10pt} 
RPL+CoroCL {\scriptsize \textcolor{black}{[Ours]}}                     & \textbf{0.52}  &\textbf{95.96}               &\textbf{ 2.27}  & 53.99           & \textbf{11.68}  &  83.49                 & \textbf{0.58}   & \textbf{85.93}                        & \textbf{3.76}  & \textbf{79.84}                                \\
\specialrule{1.5pt}{0pt}{0pt}
\end{tabular}}

\vspace{-10 pt}
\end{table*}

In this section, we first explain the experimental setup, then compare our method with the current SOTA, present the inlier segmentation accuracy, and show the ablation study for RPL and CoroCL.

\noindent \textit{\textbf{Training set.}} We establish our experiments using the urban driving dataset \textbf{Cityscapes}~\cite{cordts2016cityscapes},  which consists of $2,975$ images for training, $500$ for validation and $1,525$ for testing. Every image in the dataset has resolution $2,048 \times 1,024$, and there are $19$ classes in total. Following  previous works~\cite{chan2021entropy, tian2021pixel}, we rely on \textbf{COCO}~\cite{lin2014microsoft} for the outlier exposure (OE). COCO contains objects captured in various contexts and images have at least 480 pixels for height or width. We only consider COCO samples that have visual classes that do not overlap with Cityscapes classes, which allowed us to form the dataset $\mathcal{D}^{out}$ with  $46,751$ images.

\noindent \textit{\textbf{Validation sets.}} We evaluate our approach on multiple benchmarks.  
\textbf{Fishyscapes}~\cite{blum2021fishyscapes} has high-resolution images for anomaly detection, with two validation subsets: FS-L\&F and FS-Static. FS-L\&F consists of $1000$ images from the LostAndFound dataset~\cite{pinggera2016lost} with refined labels, and FS-Static contains $30$ images with synthetic anomalous objects. \textbf{RoadAnomaly}~\cite{lis2019detecting} is another large anomaly validation set that contains real-world anomalous objects with $60$ images from the Internet. 
\textbf{Segment-Me-If-You-Can} (SMIYC)~\cite{chan2021segmentmeifyoucan} has two subsets, including  \textit{AnomalyTrack} that contains large anomaly objects on different contexts (partially overlapped with~\cite{lis2019detecting}), and \textit{ObstacleTrack} which contains small obstacles on the road ahead. There are $10$ and $30$ images for validation, and $100$ and $327$ images for online test in AnomalyTrack and ObstacleTrack, respectively. 
In addition, Chan et al.~\cite{chan2021segmentmeifyoucan} eliminated the mislabeled normal frames from LostAndFound~\cite{pinggera2016lost} and proposed the SMIYC-L\&F to simulate real-world driving scene situations.


\noindent \textbf{Implementation Details.} Our experiments are based on the DeepLabv3+ architecture with WiderResNet38 as the backbone. We present the results with other backbones (e.g., resnet101, resnet50) in Supplementary Sec.\textcolor{red}{D}. Following~\cite{chan2021entropy, chan2021segmentmeifyoucan, tian2021pixel, grcic2022densehybrid}, our training is based on the pre-trained checkpoints from Cityscapes~\cite{cordts2016cityscapes}, where WiderResNet38 is from~\cite{zhu2019improving}. 
During training, we set the initial learning rate to be $7.5e^{-5}$ and utilise poly learning rate decay with $(1-\frac{\text{iter}}{\text{max\_iter}})^{0.9}$ and $40$ epochs in total. We set temperature hyper-parameters $\mathbf{t}=1$ for $\ell_{reg}$ in~\eqref{eq:l_in} and $\tau=0.10$ in~\eqref{eq:CoroCL} for all the experiments.
Our RPL block $f_{\theta_{\text{rpl}}^a}(.)$ is formed with the ASPP layers, followed by a layer $f_{\theta_{\text{rpl}}^b}(.)$
that expands the channels to be the same as the input of segmentation head $f_{\phi_{\text{seg}}}(.)$  (from $256$ to $304$). The projection layer $f_{\theta_\text{proj}}(.)$ for contrastive learning is represented by a
one-layer perceptron. 
Supplementary Sec.\textcolor{red}{A} provides the details about the RPL architecture, and Supplementary Sec.\textcolor{red}{C} contains more implementation details.\\
\noindent \textbf{Number of parameters in OoD detector.} The total number of RPL parameters for training is $\mathbf{30.89}${M}, which is significantly lower than the parameters in Meta-OoD~\cite{chan2021entropy} and DenseHybrid~\cite{grcic2022densehybrid}, which are $\mathbf{\sim137.11}${M}.\\
\noindent \textbf{Evaluation metrics.} Following~\cite{tian2021pixel, jung2021standardized,chan2021entropy}, we compute the receiver operating characteristics (AuROC), the area under precision recall curve (AuPRC), false positive rate at a true positive rate (FPR) of 95\% and F1 score to evaluate our approach. We utilise mean Intersection Over Union (mIoU) to measure inlier segmentation performance, following the common practice in segmentation~\cite{chen2017deeplab,wang2021exploring,wang2021dense,zhao2017pyramid}. 

\begin{figure}[t!]
    \centering
    \includegraphics[width=\linewidth]{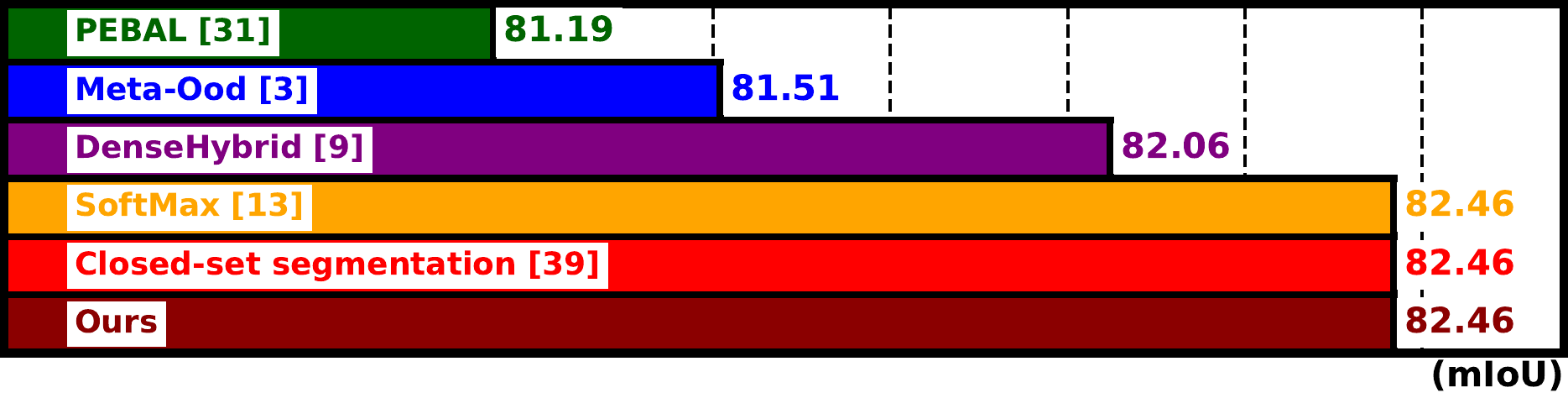}
    \vspace{-20 pt}
    \caption{Cityscapes \textbf{test set} results 
    based on \textbf{closed-set segmentation model} (i.e., without considering the OoD detection) with \textbf{single-scale} sliding evaluation.} 
    \label{fig:inlier}
    \vspace{-15 pt}
\end{figure}
\textit{We evaluate approaches for all the validation/test sets with one consistent model to simulate the real-world challenging scenario. All  methods are based on the same segmentation architecture for fair competition.}


\subsection{Comparison with SOTA Methods}
In this section, we compare our approach with others for both inlier and pixel-wise OoD detection. 
We first evaluate our approach on all  \textbf{validation sets} with AP and FPR in~\cref{tab:valid}, where we follow~\cite{tian2021pixel} and measure AuROC on Fishyscapes~\cite{blum2021fishyscapes} and the F1 score on SMIYC~\cite{chan2021segmentmeifyoucan} based on DeepLabv3+ architecture. 
The results show that the previous SOTA methods~\cite{chan2021entropy,tian2021pixel,grcic2022densehybrid} do not produce consistent performance in various contexts, where 
PEBAL~\cite{tian2021pixel} performs well in Fishyscapes~\cite{blum2021fishyscapes}, but fail in the SMIYC-Anomaly and SMIYC-Obstacle from~\cite{chan2021segmentmeifyoucan}. In contrast, Meta-OoD~\cite{chan2021entropy} yields stable results in SMIYC~\cite{chan2021segmentmeifyoucan}, while its performance degrades in Fishyscapes~\cite{blum2021fishyscapes}. 
Our results are stable across all validation sets, achieving SOTA performance for all the measurements. 
Notably, we surpass the previous SOTA PEBAL~\cite{tian2021pixel} with 0.67\%, 2.24\% FPR and 0.38\%, 11.8\% AuPRC in FS-Static and FS-L\&F, respectively. Meanwhile, we outperform Meta-OoD~\cite{chan2021entropy} with 10.25\%, 0.32\% FPR and 8.42\%, 2.77\% AuPRC in SMIYC~\cite{chan2021segmentmeifyoucan}. \cref{tab:road} compares our results with SOTA methods on RoadAnomaly, which is one of the most challenging datasets. Our approach shows the best performance with improvements of 10.54\%, 9.24\%, in FPR and AuPRC comparing with PEBAL~\cite{tian2021pixel}.

To further show the efficacy of our approach, we test it on the official black-box \textbf{test sets} in~\cref{tab:test}, which shows the results on Fishyscapes\protect\footnote{\url{https://fishyscapes.com/results}} and SMIYC\protect\footnote{\url{https://segmentmeifyoucan.com/leaderboard}}.  
We achieve the best performance by a large margin on the Fishyscapes-Static compared with the previous SOTAs. Specifically, our results is 1.21\% FPR, 3.58\% AuPRC better than PEBAL, and 4.99\% FPR, 23.69\% AuPRC better than Densehybrid in FS-Static. 
Our approach also yields around 4\% to 6\%  FPR and 10\% AuPRC improvements in FS-L\&F, demonstrating its effectiveness in detecting unseen anomalies in urban driving situations. 
However, our approach shows a slightly worse (1.6\%) AuPRC, compared with GMMSeg~\cite{liang2022gmmseg}, which can be explained by its external GMM. Otherwise, our results are substantially better than GMMSeg, particularly on RoadAnomaly (see Tab.~\ref{tab:road}). 
Our result also reaches SOTA on the SMIYC test set, except for the slightly worse AuPRC (in SMIYC-Anomaly) compared with Meta-OoD~\cite{chan2021entropy}, which failed in FS-L\&F with 35.14\% FPR and 29.96\% AuPRC. Also, our approach's lowest FPR in SMIYC shows its effectiveness. 
Considering the overall average FPR and AP results, our approach outperforms previous SOTA methods~\cite{tian2021pixel,chan2021entropy} by a significant margin. \\
\indent The experiments based on fine-tuning with the extra training sets (including Vistas~\cite{neuhold2017mapillary}, Wilddash2~\cite{zendel2018wilddash}) are shown in the Supplementary Sec.\textcolor{red}{B}, where our method shows better results than the SOTAs under same setup.

The performance of the inlier categories is another essential measurement of a model's efficacy, so we show the \textbf{in-distribution segmentation accuracy} on Cityscapes in~\cref{fig:inlier}, following~\cite{jung2021standardized,di2021pixel,grcic2022densehybrid}. 
The results based on \textbf{single-scale sliding evaluation} and show that PEBAL~\cite{tian2021pixel} decreases 1.27\% mIoU, and Meta-OoD~\cite{chan2021entropy} decreases 0.95\% mIoU, while our approach and other 
OoD methods that freeze networks (i.e., Synboost~\cite{di2021pixel}) maintain the same accuracy for inlier categories.

\begin{figure}[t!]
    \centering
    \vspace{-2pt}
    \begin{subfigure}[b]{0.495\linewidth}
         \centering    
            \includegraphics[width=1.\linewidth, height=2.5cm]{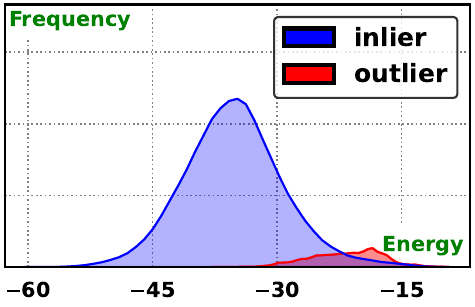}
        \caption{Hinge Energy Loss~\cite{liu2020energy}}
         \label{fig:hg_loss}
         \vspace{-7pt}
    \end{subfigure}
    \begin{subfigure}[b]{0.495\linewidth}
        \centering    
            \includegraphics[width=1.\linewidth, height=2.5cm]{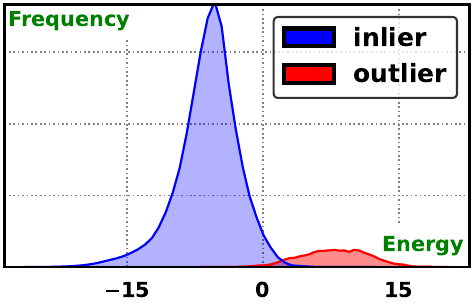}
        \caption{Positive Energy Loss (ours)}
         \label{fig:pe_loss}
         \vspace{-7pt}
     \end{subfigure}
     \caption{\textbf{Energy distribution} of hinge energy loss (a) and positive energy loss (b) on
     \textit{SMIYC (obstacle)} validation set. }
    \vspace{-12 pt}
     \label{fig:dist-diff}
\end{figure}

\begin{table*}[t!]
\caption{\textbf{Ablation study for our approach}. The datasets Static, L\&F are from Fishyscapes validation set and Anomaly, Obstacle are from SMIYC. The first two rows show baseline results from the hinge energy loss~\cite{tian2021pixel} and the entropy maximisation~\cite{chan2021entropy}. 
PE denotes the positive energy loss from~\eqref{eq:l_out}, DS denotes the dis-similarity regularisation $\ell_{\text{reg}}$ in~\eqref{eq:l_in}. CoroCL is for the context-robust contrastive learning, and we directly optimise the RPL output $f_{{\theta}_\text{rpl}}$ (instead of~\eqref{eq:outlier}) in  \textcolor{darkgray}{\textbf{the bottom row}}.}
\vspace{-5 pt}
\label{tab:ablation}
\centering
\resizebox{\linewidth}{!}{
\begin{tabular}{!{\vrule width 1.5pt}c|c!{\vrule width 1.5pt}ccc!{\vrule width 1.5pt}cc|cc|cc|cc|cc!{\vrule width 1.5pt}} 
\specialrule{1.5pt}{0pt}{0pt}
\multirow{2}{*}{Entropy~\cite{chan2021entropy}} & \multirow{2}{*}{Energy~\cite{tian2021pixel}} & \multirow{2}{*}{PE} & \multirow{2}{*}{DS} & \multirow{2}{*}{CoroCL} & \multicolumn{2}{c|}{Static} & \multicolumn{2}{c|}{L\&F} & \multicolumn{2}{c|}{Anomaly} & \multicolumn{2}{c|}{Obstacle} & \multicolumn{2}{c!{\vrule width 1.5pt}}{RoadAnomaly}  \\ 
\cline{6-15}
                                 &                                &                              &                               &                       & FPR  & AuPRC                   & FPR  & AuPRC               & FPR   & AuPRC                      & FPR  & AuPRC                        & FPR   & AuPRC                      \\ 
\specialrule{1.5pt}{0pt}{0pt}

\rule{0pt}{10pt}
                                 \textcolor{red}{\CheckmarkBold}         &                       &                              &                               &                       & 1.78 & 86.88                   & 5.04 & 52.10               & 27.59 & 60.76                      & 2.50 & 80.85                        & 28.63 & 49.22                      \\ 	
                    &             \textcolor{red}{\CheckmarkBold}                        &                              &                               &                       & 1.65 & 89.08                   & 6.64 & 51.47               & 28.23 & 70.18                      & 1.57 & 71.40                        & 34.39 & 52.66  \\            

\specialrule{1.5pt}{0pt}{0pt}
                                 &                                & \textcolor{red}{\CheckmarkBold}                     &                               &                       & 1.55 & 88.05                   & 4.52 & 56.85               & 26.58 & 73.66                      & 0.51 & 92.36                        & 31.96 & 57.28                      \\
                                 &                                & \textcolor{red}{\CheckmarkBold}                     & \textcolor{red}{\CheckmarkBold}                      &                       & 1.30 & 91.16                   & 3.79 & 63.72               & 25.65 & 76.43                      & 0.42 & 93.25                        & 30.66 & 63.02                      \\
                                 &                                & \textcolor{red}{\CheckmarkBold}                     & \textcolor{red}{\CheckmarkBold}                      & \textcolor{red}{\CheckmarkBold}              &  \textbf{0.85}  & \textbf{92.46}                   & \textbf{2.52}    & \textbf{70.61}               & \textbf{7.18} & \textbf{88.55}                      & \textbf{0.09} & \textbf{96.91}                      & \textbf{17.74}    & \textbf{71.61}                       \\
\specialrule{1.5pt}{0pt}{0pt}    
\rowcolor{lightgray}
                                 &                                & \textcolor{gray}{\CheckmarkBold}                     &                               &                       & 6.92 & 54.31                   & 17.18 & 32.57               & 29.69 & 68.99                     & 0.95	& 83.08                        & 33.72	& 57.99                      \\

\specialrule{1.5pt}{0pt}{0pt}
\end{tabular}}
\vspace{-10 pt}
\end{table*}

\begin{table}[t!]
\caption{FP detection in Cityscape validation set based on threshold $t=0.0$. The results are averaged over the images, each with $\approx 1.8\times10^{6}$  
pixels (excluding the \emph{ignore} labels). } 
\label{tab:fp_detect}
\vspace{-7 pt}
\resizebox{\linewidth}{!}{\begin{tabular}{!{\vrule width 1.5pt}c|c|c|c!{\vrule width 1.5pt}} 
\specialrule{1.5pt}{0pt}{0pt}
 RPL($\alpha=.05$)+CoroCl & RPL($\alpha=.05$) & RPL ($\alpha=.10$)  & RPL ($\alpha=.15$) \\ 
\hline
  FP=1949.306   &   FP=4360.534         &    FP=10784.576      &     FP=19837.211                   \\
\specialrule{1.5pt}{0pt}{0pt}
\end{tabular}}
\end{table}

\begin{figure}[t!]
    \centering
    \begin{subfigure}[b]{0.495\linewidth}
         \centering    
            \includegraphics[width=1.\linewidth, height=3.3cm]{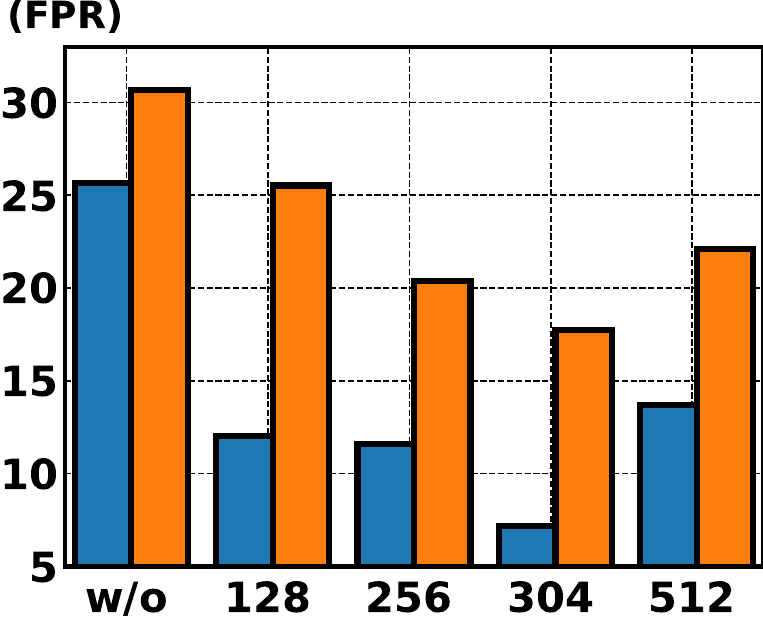}
        \caption{}
        \label{fig:embed_a}
         \vspace{-7pt}
    \end{subfigure}
    \begin{subfigure}[b]{0.495\linewidth}
        \centering    
            \includegraphics[width=1.\linewidth, height=3.3cm]{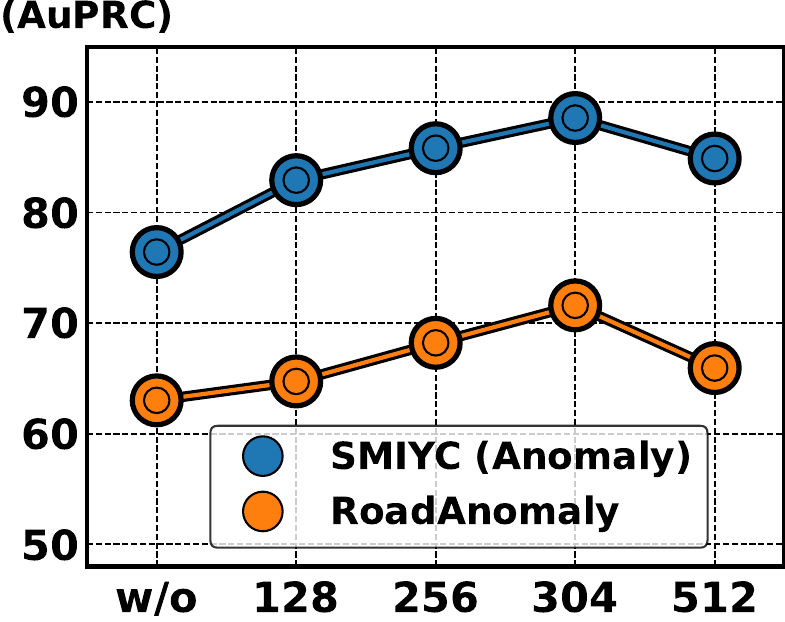}
        \caption{}
        \label{fig:embed_b}
         \vspace{-7pt}
     \end{subfigure}
     \caption{\textbf{The effect of the embedding depth} on the SMIYC-Anomaly and RoadAnomaly datasets, where x-axis is the depth number. Note that $304$ is the output of the RPL from $f_{\theta_{\text{rpl}}}$ in~\eqref{eq:outlier}. }
     \label{fig:embedding_depth}
    \vspace{-15 pt}
\end{figure}

\subsection{Ablation Studies}

\cref{tab:ablation} shows the ablation study of the RPL module, where Entropy~\cite{chan2021entropy} is the entropy maximisation (first row), and  Energy~\cite{tian2021pixel} denotes the hinge energy loss (second row). 
In RPL optimisation, we test the positive energy (PE) loss from~\eqref{eq:l_out}, the dis-similarity (DS) regularisation  in~\eqref{eq:l_in}, and the Context-robust Contrastive Learning (CoroCL) in~\eqref{eq:CoroCL}.
In general, the entropy-based loss~\cite{chan2021entropy} and the hinge-based energy loss~\cite{tian2021pixel} yield unsatisfactory performance for all the validation sets, where the former provides low AuPRC and the latter fails in small obstacle situations (SMIYC-obstacle). Our proposed PE loss improves 0.48\% FPR, 4.75\% AuPRC and 1.99\% FPR, 11.51\% AuPRC in FS-L\&F and SMIYC-Obstacle by comparing it with Entropy optimisation~\cite{chan2021entropy}. 
After adding the DS regularisation, the results improve for all validation sets, where AuPRC has large improvement, for example, it increases 3.11\%, 6.87\%  AuPRC in FS-Static and FS-L\&F, respectively. In the row before the last, CoroCL shows 18.47\% FPR, 12.12\% AuPRC improvements in SMIYC-Anomaly and 12.92\% FPR, 8.59\% AuPRC in RoadAnomaly. Lastly, we show the results of the direct optimisation of RPL as an independent classifier (i.e., without adding the RPL output back to the intermediate features of the segmentation model). Compared with direct optimisation (\textcolor{darkgray}{the bottom row}), the result with PE (the 3rd row) shows that our RPL can more effectively detect potential anomalies (e.g.,  improves 12.66\% FPR in L\& F).

\begin{table}[t!]
\centering
\caption{\textbf{Our RPL brings improvements} to previous SOTAs in Fishyscapes and SMIYC validation sets.}
\vspace{-8 pt}
\label{tab:rpl_sotas}
\resizebox{\linewidth}{27pt}{
\begin{tabular}{!{\vrule width 1.5pt}l!{\vrule width 1.5pt}c|c!{\vrule width 1.5pt}c|c!{\vrule width 1.5pt}c|c!{\vrule width 1.5pt}c|c!{\vrule width 1.5pt}} 
\specialrule{1.5pt}{0pt}{0pt}
\multicolumn{1}{!{\vrule width 1.5pt}c!{\vrule width 1.5pt}}{\multirow{2}{*}{Methods}} & \multicolumn{2}{c!{\vrule width 1.5pt}}{Static} & \multicolumn{2}{c!{\vrule width 1.5pt}}{L\&F} & \multicolumn{2}{c!{\vrule width 1.5pt}}{Anomaly} & \multicolumn{2}{c!{\vrule width 1.5pt}}{Obstacle}  \\ 
\cline{2-9}
                          & FPR   & AuPRC                & FPR   & AuPRC            & FPR   & AuPRC                & FPR  & AuPRC                   \\ 
\specialrule{1.5pt}{0pt}{0pt}
Meta-OoD~\cite{chan2021entropy}                  & 13.57 & 72.91                & 37.69 & 41.31            & 17.43 & 80.13                & 0.41 & 94.14                   \\
\textbf{RPL+Meta-OoD}              & \textbf{2.34}  & \textbf{87.98}                & \textbf{7.52}  & \textbf{55.25}            & \textbf{12.91} & \textbf{82.08}                & \textbf{0.24} & \textbf{94.47}                   \\ 
\specialrule{1.5pt}{0pt}{0pt}
PEBAL~\cite{tian2021pixel}                     & 1.52  & 92.08                & 4.76  & 58.81            & 36.74 & 53.10                & 7.92 & 10.15                   \\
\textbf{RPL+PEBAL}                 & \textbf{0.78}      &   \textbf{93.28}                   &  \textbf{3.74}     &     \textbf{63.99}             &   \textbf{17.89}    &  \textbf{78.53}                   &   \textbf{3.12}    &  \textbf{84.23}                       \\
\specialrule{1.5pt}{0pt}{0pt}
\end{tabular}}
\vspace{-10 pt}
\end{table}


The RPL module is optimised with the positive energy loss from~\eqref{eq:l_in}, which we claim to be more robust to imbalanced distribution of inliers and outliers than the hinge  loss~\cite{liu2020energy}, resulting in better OoD pixel sensitivity detection.
This is shown in~\cref{fig:dist-diff} that compares the energy distribution obtained from the hinge loss (a), and our positive energy loss (b) using our RPL module on the SMIYC-obstacle dataset, where the positive energy loss yields better separation and clustering of energy distribution.


\cref{tab:ablation} shows that CoroCL provides more improvements on SMIYC-Anomaly and RoadAnomaly, so we study it further.
Recall that CoroCL works with RPL's intermediate features, so it is important to investigate the impact of embedding depth and projector architecture.
\cref{fig:embedding_depth} shows that the best performance is observed when the embedding depth is \textbf{304}, which is better than 512 by around 5\% in AuPRC for both datasets. This result enables us to conclude that when the output dimension of the projector $f_{\theta_\text{proj}}(.)$ equals that of the RPL block $f_{\theta_\text{rpl}^b}(.)$, we  have a more effective optimisation. 
\cref{tab:fp_detect} shows the pixel-wise FP anomaly detection with energy threshold $t=0.0$, where we can note that more FPs occur if we set larger $\alpha$ during training, as the model will be more sensitive to the outliers, and CoroCL successfully reduces the FPs for the inlier scenes.
. 

Our RPL module can also be deployed to SOTA methods to improve their performance. \cref{tab:rpl_sotas} shows improvements brought by RPL for PEBAL~\cite{tian2021pixel} and Meta-OoD~\cite{chan2021entropy} on Fishycapes and SMIYC datasets.
For instance, RPL improves Meta-OoD~\cite{chan2021entropy} by around 15\% AuPRC on Static and L\&F in Fishyscapes and by over 70\% AuPRC for PEBAL~\cite{tian2021pixel} on SMIYC-Obstacle. 
\vspace{-5pt}

\section{Conclusion and Discussion}
\vspace{-5pt}
In this paper, we introduced the residual pattern learning (RPL) block that induces the closed-set segmentation model to become highly uncertain for potential anomalous regions. 
Compared with recent re-training approaches~\cite{chan2021entropy,tian2021pixel,grcic2022densehybrid}, our RPL detects potential anomalies more effectively, while causing minimal impact to the closed-set segmentation accuracy~\cite{di2021pixel, jung2021standardized, hendrycks2019scaling}.
Our positive energy loss enables the detection of small anomalies with a more effective optimisation than the one used in previous hinge-loss energy optimisation~\cite{tian2021pixel}. 
Furthermore, our proposed CoroCL is shown to be robust to context changes between training and testing images, avoiding the massive  mis-detections observed in previous anomaly detectors~\cite{tian2021pixel,chan2021entropy,grcic2022densehybrid,di2021pixel,jung2021standardized}.
A limitation of our approach is that the proposed optimisation tries  to approximate the closed-set segmentation results, without further reducing its inlier entropy.
Future work will focus on how to reduce such inlier uncertainty without affecting the sensitivity of the OoD pixels.
{\small
\bibliographystyle{ieee_fullname}
\bibliography{egbib}

\begin{thebibliography}{10}\itemsep=-1pt

\bibitem{bevandic2019simultaneous}
Petra Bevandi{\'c}, Ivan Kre{\v{s}}o, Marin Or{\v{s}}i{\'c}, and Sini{\v{s}}a
  {\v{S}}egvi{\'c}.
\newblock Simultaneous semantic segmentation and outlier detection in presence
  of domain shift.
\newblock In {\em Pattern Recognition: 41st DAGM German Conference, DAGM GCPR
  2019, Dortmund, Germany, September 10--13, 2019, Proceedings 41}, pages
  33--47. Springer, 2019.

\bibitem{blum2021fishyscapes}
Hermann Blum, Paul-Edouard Sarlin, Juan Nieto, Roland Siegwart, and Cesar
  Cadena.
\newblock The fishyscapes benchmark: Measuring blind spots in semantic
  segmentation.
\newblock {\em International Journal of Computer Vision}, 129(11):3119--3135,
  2021.

\bibitem{chan2021segmentmeifyoucan}
Robin Chan, Krzysztof Lis, Svenja Uhlemeyer, Hermann Blum, Sina Honari, Roland
  Siegwart, Mathieu Salzmann, Pascal Fua, and Matthias Rottmann.
\newblock Segmentmeifyoucan: A benchmark for anomaly segmentation.
\newblock {\em arXiv preprint arXiv:2104.14812}, 2021.

\bibitem{chan2021entropy}
Robin Chan, Matthias Rottmann, and Hanno Gottschalk.
\newblock Entropy maximization and meta classification for out-of-distribution
  detection in semantic segmentation.
\newblock In {\em Proceedings of the ieee/cvf international conference on
  computer vision}, pages 5128--5137, 2021.

\bibitem{chen2017deeplab}
Liang-Chieh Chen, George Papandreou, Iasonas Kokkinos, Kevin Murphy, and Alan~L
  Yuille.
\newblock Deeplab: Semantic image segmentation with deep convolutional nets,
  atrous convolution, and fully connected crfs.
\newblock {\em IEEE transactions on pattern analysis and machine intelligence},
  40(4):834--848, 2017.

\bibitem{chen2017rethinking}
Liang-Chieh Chen, George Papandreou, Florian Schroff, and Hartwig Adam.
\newblock Rethinking atrous convolution for semantic image segmentation.
\newblock {\em arXiv preprint arXiv:1706.05587}, 2017.

\bibitem{chen2018encoder}
Liang-Chieh Chen, Yukun Zhu, George Papandreou, Florian Schroff, and Hartwig
  Adam.
\newblock Encoder-decoder with atrous separable convolution for semantic image
  segmentation.
\newblock In {\em Proceedings of the European conference on computer vision
  (ECCV)}, pages 801--818, 2018.

\bibitem{cordts2016cityscapes}
Marius Cordts, Mohamed Omran, Sebastian Ramos, Timo Rehfeld, Markus Enzweiler,
  Rodrigo Benenson, Uwe Franke, Stefan Roth, and Bernt Schiele.
\newblock The cityscapes dataset for semantic urban scene understanding.
\newblock In {\em Proceedings of the IEEE conference on computer vision and
  pattern recognition}, pages 3213--3223, 2016.

\bibitem{di2021pixel}
Giancarlo Di~Biase, Hermann Blum, Roland Siegwart, and Cesar Cadena.
\newblock Pixel-wise anomaly detection in complex driving scenes.
\newblock In {\em Proceedings of the IEEE/CVF conference on computer vision and
  pattern recognition}, pages 16918--16927, 2021.

\bibitem{grcic2021dense}
Matej Grci{\'c}, Petra Bevandi{\'c}, and Sini{\v{s}}a {\v{S}}egvi{\'c}.
\newblock Dense anomaly detection by robust learning on synthetic negative
  data.
\newblock {\em arXiv preprint arXiv:2112.12833}, 2021.

\bibitem{grcic2022densehybrid}
Matej Grci{\'c}, Petra Bevandi{\'c}, and Sini{\v{s}}a {\v{S}}egvi{\'c}.
\newblock Densehybrid: Hybrid anomaly detection for dense open-set recognition.
\newblock {\em arXiv preprint arXiv:2207.02606}, 2022.

\bibitem{hao2020brief}
Shijie Hao, Yuan Zhou, and Yanrong Guo.
\newblock A brief survey on semantic segmentation with deep learning.
\newblock {\em Neurocomputing}, 406:302--321, 2020.

\bibitem{he2015delving}
Kaiming He, Xiangyu Zhang, Shaoqing Ren, and Jian Sun.
\newblock Delving deep into rectifiers: Surpassing human-level performance on
  imagenet classification.
\newblock In {\em Proceedings of the IEEE international conference on computer
  vision}, pages 1026--1034, 2015.

\bibitem{he2016deep}
Kaiming He, Xiangyu Zhang, Shaoqing Ren, and Jian Sun.
\newblock Deep residual learning for image recognition.
\newblock In {\em Proceedings of the IEEE conference on computer vision and
  pattern recognition}, pages 770--778, 2016.

\bibitem{hendrycks2019scaling}
Dan Hendrycks, Steven Basart, Mantas Mazeika, Mohammadreza Mostajabi, Jacob
  Steinhardt, and Dawn Song.
\newblock Scaling out-of-distribution detection for real-world settings.
\newblock {\em arXiv preprint arXiv:1911.11132}, 2019.

\bibitem{hendrycks2016baseline}
Dan Hendrycks and Kevin Gimpel.
\newblock A baseline for detecting misclassified and out-of-distribution
  examples in neural networks.
\newblock {\em arXiv preprint arXiv:1610.02136}, 2016.

\bibitem{hendrycks2018deep}
Dan Hendrycks, Mantas Mazeika, and Thomas Dietterich.
\newblock Deep anomaly detection with outlier exposure.
\newblock {\em arXiv preprint arXiv:1812.04606}, 2018.

\bibitem{hong2022goss}
Jie Hong, Weihao Li, Junlin Han, Jiyang Zheng, Pengfei Fang, Mehrtash Harandi,
  and Lars Petersson.
\newblock Goss: Towards generalized open-set semantic segmentation.
\newblock {\em arXiv preprint arXiv:2203.12116}, 2022.

\bibitem{jung2021standardized}
Sanghun Jung, Jungsoo Lee, Daehoon Gwak, Sungha Choi, and Jaegul Choo.
\newblock Standardized max logits: A simple yet effective approach for
  identifying unexpected road obstacles in urban-scene segmentation.
\newblock In {\em Proceedings of the IEEE/CVF International Conference on
  Computer Vision}, pages 15425--15434, 2021.

\bibitem{khosla2020supervised}
Prannay Khosla, Piotr Teterwak, Chen Wang, Aaron Sarna, Yonglong Tian, Phillip
  Isola, Aaron Maschinot, Ce Liu, and Dilip Krishnan.
\newblock Supervised contrastive learning.
\newblock {\em Advances in Neural Information Processing Systems},
  33:18661--18673, 2020.

\bibitem{krevso2020efficient}
Ivan Kre{\v{s}}o, Josip Krapac, and Sini{\v{s}}a {\v{S}}egvi{\'c}.
\newblock Efficient ladder-style densenets for semantic segmentation of large
  images.
\newblock {\em IEEE Transactions on Intelligent Transportation Systems},
  22(8):4951--4961, 2020.

\bibitem{lakshminarayanan2017simple}
Balaji Lakshminarayanan, Alexander Pritzel, and Charles Blundell.
\newblock Simple and scalable predictive uncertainty estimation using deep
  ensembles.
\newblock {\em Advances in neural information processing systems}, 30, 2017.

\bibitem{lee2018simple}
Kimin Lee, Kibok Lee, Honglak Lee, and Jinwoo Shin.
\newblock A simple unified framework for detecting out-of-distribution samples
  and adversarial attacks.
\newblock {\em Advances in neural information processing systems}, 31, 2018.

\bibitem{li2022targeted}
Tianhong Li, Peng Cao, Yuan Yuan, Lijie Fan, Yuzhe Yang, Rogerio~S Feris, Piotr
  Indyk, and Dina Katabi.
\newblock Targeted supervised contrastive learning for long-tailed recognition.
\newblock In {\em Proceedings of the IEEE/CVF Conference on Computer Vision and
  Pattern Recognition}, pages 6918--6928, 2022.

\bibitem{liang2022gmmseg}
Chen Liang, Wenguan Wang, Jiaxu Miao, and Yi Yang.
\newblock Gmmseg: Gaussian mixture based generative semantic segmentation
  models.
\newblock {\em arXiv preprint arXiv:2210.02025}, 2022.

\bibitem{lin2017feature}
Tsung-Yi Lin, Piotr Doll{\'a}r, Ross Girshick, Kaiming He, Bharath Hariharan,
  and Serge Belongie.
\newblock Feature pyramid networks for object detection.
\newblock In {\em Proceedings of the IEEE conference on computer vision and
  pattern recognition}, pages 2117--2125, 2017.

\bibitem{lin2014microsoft}
Tsung-Yi Lin, Michael Maire, Serge Belongie, James Hays, Pietro Perona, Deva
  Ramanan, Piotr Doll{\'a}r, and C~Lawrence Zitnick.
\newblock Microsoft coco: Common objects in context.
\newblock In {\em European conference on computer vision}, pages 740--755.
  Springer, 2014.

\bibitem{lis2019detecting}
Krzysztof Lis, Krishna Nakka, Pascal Fua, and Mathieu Salzmann.
\newblock Detecting the unexpected via image resynthesis.
\newblock In {\em Proceedings of the IEEE/CVF International Conference on
  Computer Vision}, pages 2152--2161, 2019.

\bibitem{liu2020energy}
Weitang Liu, Xiaoyun Wang, John Owens, and Yixuan Li.
\newblock Energy-based out-of-distribution detection.
\newblock {\em Advances in Neural Information Processing Systems}, 2020.

\bibitem{liu2019deep}
Ziyin Liu, Zhikang Wang, Paul~Pu Liang, Russ~R Salakhutdinov, Louis-Philippe
  Morency, and Masahito Ueda.
\newblock Deep gamblers: Learning to abstain with portfolio theory.
\newblock {\em Advances in Neural Information Processing Systems}, 32, 2019.

\bibitem{long2015fully}
Jonathan Long, Evan Shelhamer, and Trevor Darrell.
\newblock Fully convolutional networks for semantic segmentation.
\newblock In {\em Proceedings of the IEEE conference on computer vision and
  pattern recognition}, pages 3431--3440, 2015.

\bibitem{mukhoti2018evaluating}
Jishnu Mukhoti and Yarin Gal.
\newblock Evaluating bayesian deep learning methods for semantic segmentation.
\newblock {\em arXiv preprint arXiv:1811.12709}, 2018.

\bibitem{neal2012bayesian}
Radford~M Neal.
\newblock {\em Bayesian learning for neural networks}, volume 118.
\newblock Springer Science \& Business Media, 2012.

\bibitem{neuhold2017mapillary}
Gerhard Neuhold, Tobias Ollmann, Samuel Rota~Bulo, and Peter Kontschieder.
\newblock The mapillary vistas dataset for semantic understanding of street
  scenes.
\newblock In {\em Proceedings of the IEEE international conference on computer
  vision}, pages 4990--4999, 2017.

\bibitem{pinggera2016lost}
Peter Pinggera, Sebastian Ramos, Stefan Gehrig, Uwe Franke, Carsten Rother, and
  Rudolf Mester.
\newblock Lost and found: detecting small road hazards for self-driving
  vehicles.
\newblock In {\em 2016 IEEE/RSJ International Conference on Intelligent Robots
  and Systems (IROS)}, pages 1099--1106. IEEE, 2016.

\bibitem{ronneberger2015u}
Olaf Ronneberger, Philipp Fischer, and Thomas Brox.
\newblock U-net: Convolutional networks for biomedical image segmentation.
\newblock In {\em International Conference on Medical image computing and
  computer-assisted intervention}, pages 234--241. Springer, 2015.

\bibitem{tian2021pixel}
Yu Tian, Yuyuan Liu, Guansong Pang, Fengbei Liu, Yuanhong Chen, and Gustavo
  Carneiro.
\newblock Pixel-wise energy-biased abstention learning for anomaly segmentation
  on complex urban driving scenes.
\newblock {\em arXiv preprint arXiv:2111.12264}, 2021.

\bibitem{tian2020makes}
Yonglong Tian, Chen Sun, Ben Poole, Dilip Krishnan, Cordelia Schmid, and
  Phillip Isola.
\newblock What makes for good views for contrastive learning?
\newblock {\em Advances in Neural Information Processing Systems},
  33:6827--6839, 2020.

\bibitem{wang2018high}
Ting-Chun Wang, Ming-Yu Liu, Jun-Yan Zhu, Andrew Tao, Jan Kautz, and Bryan
  Catanzaro.
\newblock High-resolution image synthesis and semantic manipulation with
  conditional gans.
\newblock In {\em Proceedings of the IEEE conference on computer vision and
  pattern recognition}, pages 8798--8807, 2018.

\bibitem{wang2021exploring}
Wenguan Wang, Tianfei Zhou, Fisher Yu, Jifeng Dai, Ender Konukoglu, and Luc
  Van~Gool.
\newblock Exploring cross-image pixel contrast for semantic segmentation.
\newblock In {\em Proceedings of the IEEE/CVF International Conference on
  Computer Vision}, pages 7303--7313, 2021.

\bibitem{wang2021dense}
Xinlong Wang, Rufeng Zhang, Chunhua Shen, Tao Kong, and Lei Li.
\newblock Dense contrastive learning for self-supervised visual pre-training.
\newblock In {\em Proceedings of the IEEE/CVF Conference on Computer Vision and
  Pattern Recognition}, pages 3024--3033, 2021.

\bibitem{xia2020synthesize}
Yingda Xia, Yi Zhang, Fengze Liu, Wei Shen, and Alan Yuille.
\newblock Synthesize then compare: Detecting failures and anomalies for
  semantic segmentation.
\newblock In {\em Proceedings of the European Conference on Computer Vision
  (ECCV)}, 2020.

\bibitem{yuan2020object}
Yuhui Yuan, Xilin Chen, and Jingdong Wang.
\newblock Object-contextual representations for semantic segmentation.
\newblock In {\em European conference on computer vision}, pages 173--190.
  Springer, 2020.

\bibitem{yurtsever2020survey}
Ekim Yurtsever, Jacob Lambert, Alexander Carballo, and Kazuya Takeda.
\newblock A survey of autonomous driving: Common practices and emerging
  technologies.
\newblock {\em IEEE access}, 8:58443--58469, 2020.

\bibitem{zendel2018wilddash}
Oliver Zendel, Katrin Honauer, Markus Murschitz, Daniel Steininger, and
  Gustavo~Fernandez Dominguez.
\newblock Wilddash-creating hazard-aware benchmarks.
\newblock In {\em Proceedings of the European Conference on Computer Vision
  (ECCV)}, pages 402--416, 2018.

\bibitem{zhao2017pyramid}
Hengshuang Zhao, Jianping Shi, Xiaojuan Qi, Xiaogang Wang, and Jiaya Jia.
\newblock Pyramid scene parsing network.
\newblock In {\em Proceedings of the IEEE conference on computer vision and
  pattern recognition}, pages 2881--2890, 2017.

\bibitem{zhu2019improving}
Yi Zhu, Karan Sapra, Fitsum~A Reda, Kevin~J Shih, Shawn Newsam, Andrew Tao, and
  Bryan Catanzaro.
\newblock Improving semantic segmentation via video propagation and label
  relaxation.
\newblock In {\em Proceedings of the IEEE/CVF Conference on Computer Vision and
  Pattern Recognition}, pages 8856--8865, 2019.

\end{thebibliography}
}

\setcounter{equation}{0}
\setcounter{figure}{0}
\setcounter{table}{0}
\setcounter{page}{1}
\setcounter{section}{0}

\twocolumn[{%
\renewcommand\twocolumn[1][]{#1}%
\maketitle
\begin{center}
    \textbf{\Large Supplementary Material of Residual Pattern Learning for Pixel-wise Out-of-Distribution Detection in Semantic Segmentation}
    \vspace{20pt}
    \centering
    \captionsetup{type=figure}
    \includegraphics[width=\textwidth]{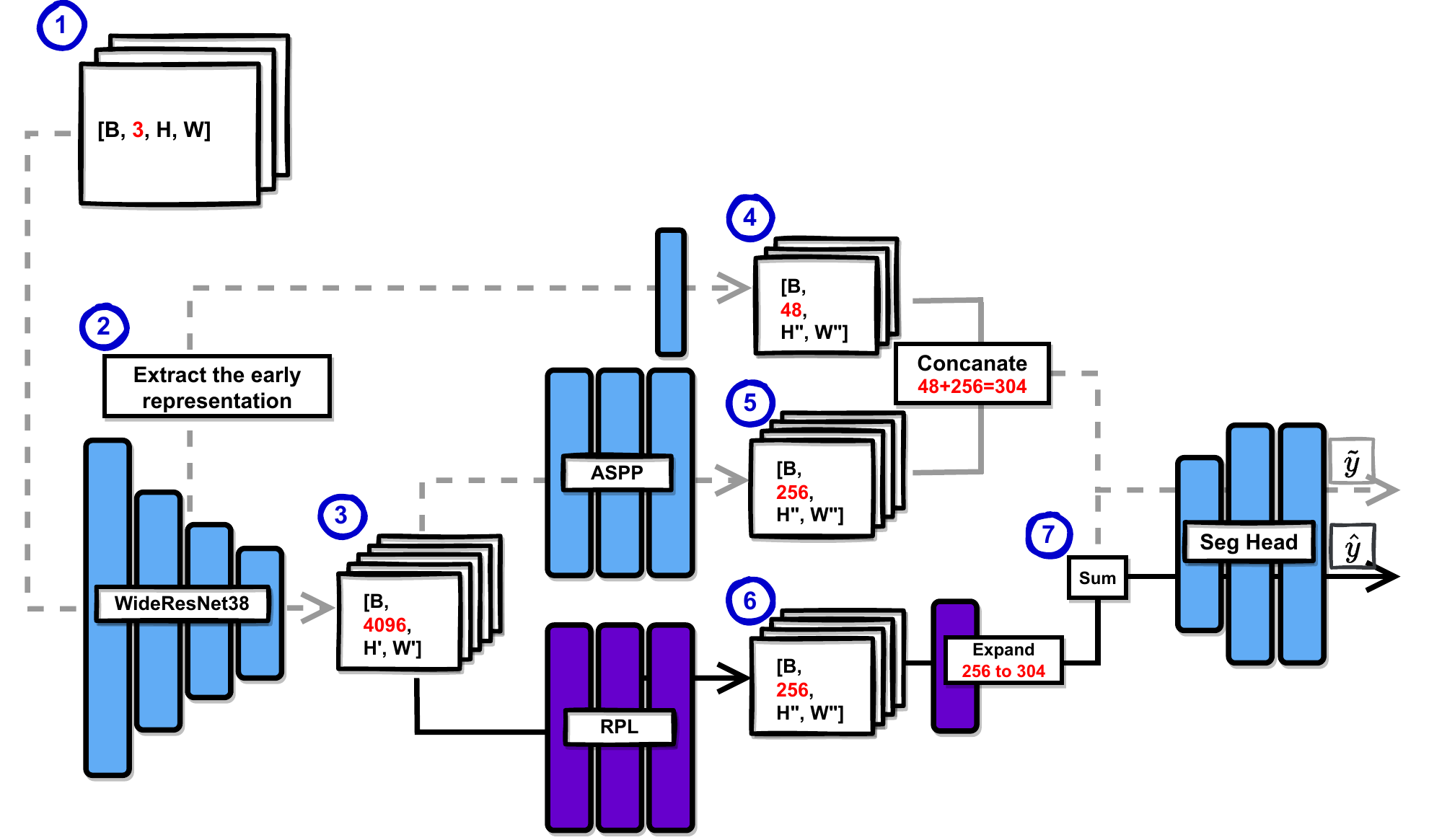}
    \caption{\textbf{The detailed workflow of the RPL module.} The \textcolor{cyan}{blue blocks} denote the convolutional layers in the original closed-set segmentation models, the \textcolor{violet}{violet blocks} represent our RPL module, and the \textcolor{gray}{dashed line (``- - -")} means the blocks that are processed without requiring training. We produce the pseudo label ($\Tilde{\mathbf{y}}$) via the path of \textcolor{blue}{\protect\circled{1}} $\rightarrow$ \textcolor{blue}{\protect\circled{2}} $\rightarrow$ \textcolor{blue}{\protect\circled{3}} $\rightarrow$ \textcolor{blue}{\protect\circled{5}} $\cup$ \textcolor{blue}{\protect\circled{4}} $\rightarrow$ $\Tilde{\mathbf{y}}$, and we produce prediction ($\hat{\mathbf{y}}$) via \textcolor{blue}{\protect\circled{1}} $\rightarrow$ \textcolor{blue}{\protect\circled{3}} $\rightarrow$ \textcolor{blue}{\protect\circled{6}} $\rightarrow$ \textcolor{blue}{\protect\circled{7}} + (\textcolor{blue}{\protect\circled{5}} $\cup$ \textcolor{blue}{\protect\circled{4}}) $\rightarrow$ $\hat{\mathbf{y}}$}.
    \vspace{10pt}
    \label{fig:rpl_1} 
\end{center}
}]

\maketitle


\begin{figure*}[t!]
    \centering
    \includegraphics[width=\linewidth]{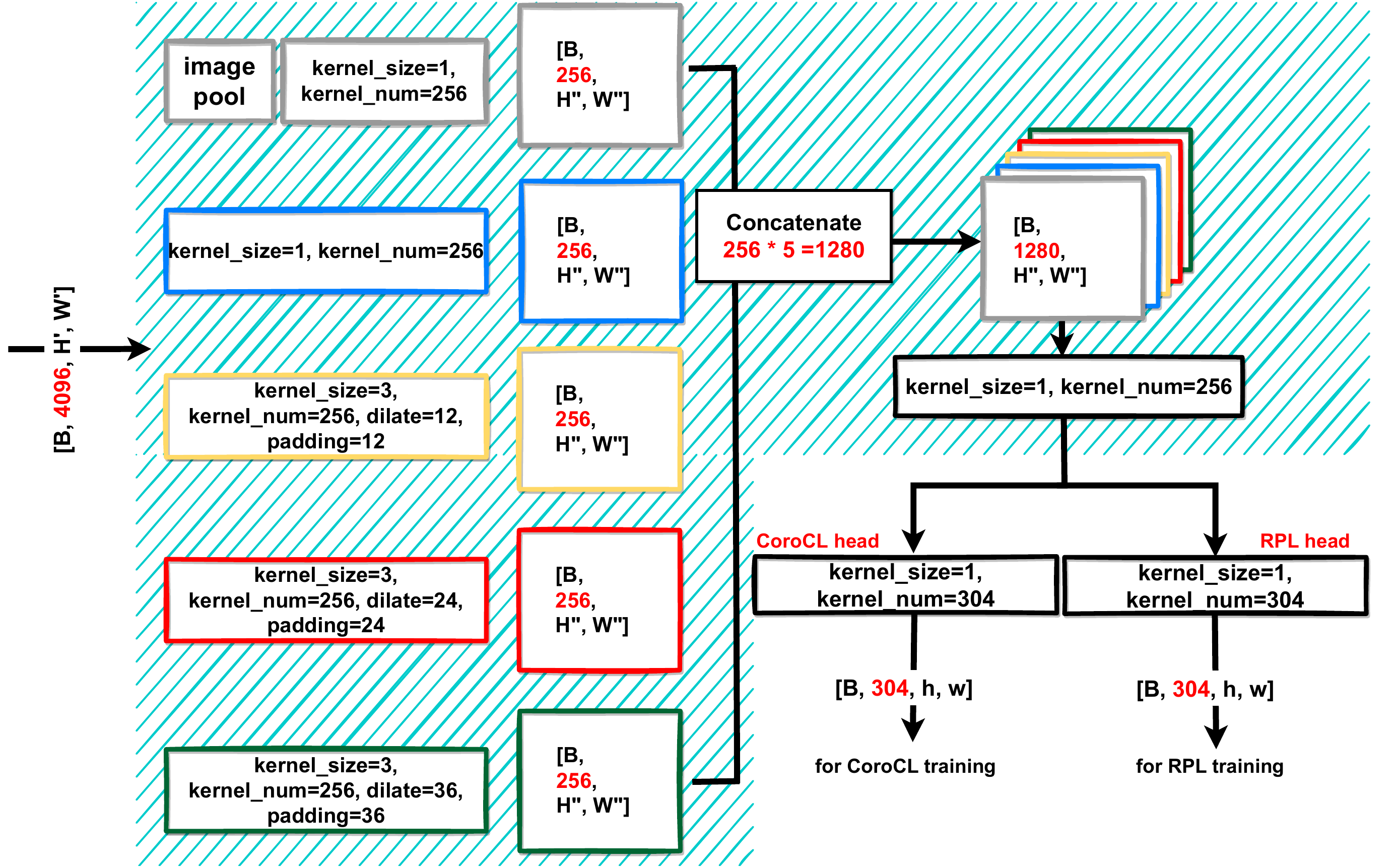}
    \caption{\textbf{The detailed architecture of RPL and CoroCL.} Our proposed RPL firstly encodes the incoming features from the segmentation network into a set features extracted from different dilated rates and concatenate them together. After being processed by the following convolutional layer, RPL will output the results for CoroCL optimisation (main paper Eq.~\textcolor{red}{(7)}) and segmentation head (main paper  Eq.~\textcolor{red}{(3)}) based on two separate heads. Note: the region inside the \textcolor{darkcyan}{cyan} region is motivated from ASPP~\cite{chen2017rethinking,chen2018encoder}.} 
    \label{fig:detailed_arch}
    \vspace{-10pt}
\end{figure*}

\begin{figure}[ht!]
    \centering
    \includegraphics[width=\linewidth]{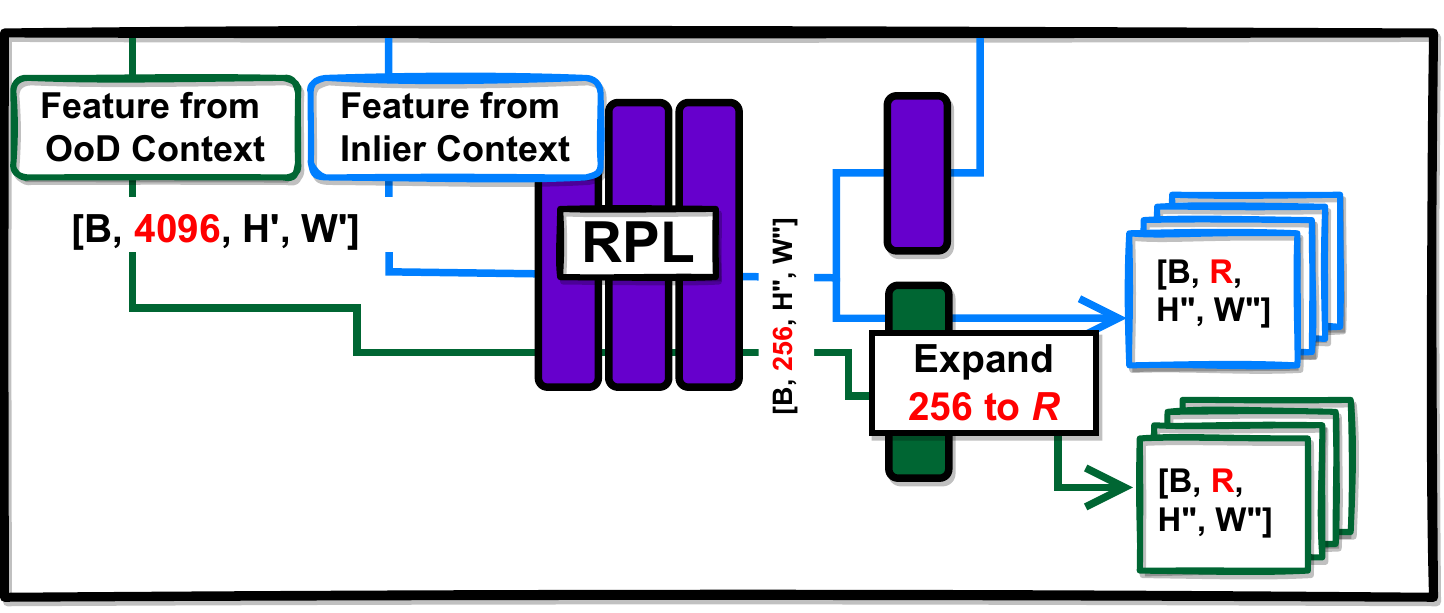}
    \caption{\textbf{The detailed workflow of the CoroCL.} The \textcolor{blue}{blue lines} denote the forward pass with the inlier context features (i.e., based on the input of $\mathbf{x}^{oe}$), while the \textcolor{teal}{green lines} represent the OoD context features (i.e., based on $\mathbf{x}^{out}$).}
    \label{fig:cl_2}
    \vspace{-15pt}
\end{figure}

\section{The Architecture of Residual Pattern Learning (RPL) and DeepLabV3+}
The RPL module is externally attached to the closed-set segmentation network that assists in deciding the potential anomalies, where we utilise DeeplabV3+~\cite{chen2018encoder} as our architecture. As shown in~\cref{fig:rpl_1}, the batch (\textbf{B}) of RGB-based images under height (\textbf{H}) and width (\textbf{W}) in \textcolor{blue}{\circled{1}} will be fed to the FCN encoder network (e.g., WiderResNet38) first to produce the feature map with \textbf{4096} channels in \textcolor{blue}{\circled{3}}. This feature map will then go through the Atrous Spatial Pyramid Pooling (ASPP) layers and RPL module to produce the outputs under the same resolution (i.e., \textbf{256} channels) in \textcolor{blue}{\circled{5}} and \textcolor{blue}{\circled{6}}, respectively. After that, the representation extracted from shallow layers (in \textcolor{blue}{\circled{2}}) will go through a convolutional layer to produce the feature maps in \textcolor{blue}{\circled{4}} that are concatenated with \textcolor{blue}{\circled{5}}. Then 
the combined feature map will be fed into the final classifier (Seg. head) to produce $\Tilde{\mathbf{y}}$. The feature map in \textcolor{blue}{\circled{6}} will be processed by the following convolutional layer to expand the channels to \textbf{304}, which are added to the intermediate feature map from the original segmentation model in \textcolor{blue}{\circled{7}}. Finally, such feature map with a potential anomaly will be classified to produce $\hat{\mathbf{y}}$. During training, we utilise $H=700$, $W=700$ and $B=8$ in stage \textcolor{blue}{\circled{1}}, to produce $H'=W'=88$ in stage \textcolor{blue}{\circled{3}} and $H"=W"=350$ in \textcolor{blue}{\circled{4}} \textcolor{blue}{\circled{5}} \textcolor{blue}{\circled{6}}. Finally, the Seg. head will produce $\Tilde{\mathbf{y}}$, $\hat{\mathbf{y}}$ with shape $8\times 19 \times 700 \times 700$ based on the  bilinear upsampling, where $19$ is the closed-set (i.e., Cityscapes~\cite{cordts2016cityscapes}) categories.

\subsection{The Architecture of RPL with Context-Robust Contrastive Learning (CoroCL)}
On top of the RPL module, we propose CoroCL to generalise the OoD detector for various open-world contexts, as demonstrated in~\cref{fig:cl_2}. CoroCL pulls the embedding features that belong to the same class (i.e., both are OoD or inliers) closer and pushes apart those embeddings from different classes (i.e., one is inlier and the other is the outlier, or vice versa). We extract those embeddings based on an extra convolutional layer (also known as the ”projector”) via the intermediate features from inlier and OoD contexts, where the projector expands the features from 256 channels to the $R$ depth of the embedding features. In our experiments, $R=304$ shows the best performance which is demonstrated in  Fig.~\textcolor{red}{6} of the main paper.

\begin{table*}[t!]
\centering
\caption{\textbf{Comparing with SOTAs on Fihsyscapes and SMIYC test benchmarks}\textcolor{red}{\protect\footnotemark[1]}$^,$\textcolor{red}{\protect\footnotemark[2]} with extra datasets \cite{neuhold2017mapillary, zendel2018wilddash}. Our results are in \textbf{bold}, and the $\textcolor{gray}{gray}$ row shows the method~\cite{grcic2021dense} that utilises a post-processing to narrow the anomaly detection area.}
\vspace{-5pt}
\label{tab:extra_data}
\resizebox{\linewidth}{!}{%
\begin{tabular}{!{\vrule width 1.5pt}r|c!{\vrule width 1.5pt}cc|cc|cc|cc!{\vrule width 1.5pt}} 
\specialrule{1.5pt}{0pt}{0pt}
\multicolumn{1}{!{\vrule width 1.5pt}c|}{\multirow{3}{*}{Methods}} & \multirow{3}{*}{Anomaly Detection Area} & \multicolumn{4}{c|}{Fishyscapes (test)}                             & \multicolumn{4}{c!{\vrule width 1.5pt}}{SMIYC (test)}                                    \\ 
\cline{3-10}
                         &                                      & \multicolumn{2}{c|}{Static} & \multicolumn{2}{c|}{Lost\&Found} & \multicolumn{2}{c|}{AnomalyTrack} & \multicolumn{2}{c!{\vrule width 1.5pt}}{ObstacleTrack}  \\ 
\cline{3-10}
                         &                                      & FPR & AuPRC                 & FPR & AuPRC                    & FPR & AuPRC                  & FPR & AuPRC                    \\

\specialrule{1.5pt}{0pt}{0pt}
NFlowJS~\cite{grcic2021dense}                  & whole image                                   &    15.41 &   52.12                    &  8.98   &   39.36                        &  
34.71   &  56.92                      &  
0.41   &  85.55                        \\
DenseHybrid~\cite{grcic2022densehybrid}              & whole image                                   &  -   & -                      & -    &   -                       &  9.81   &     77.96                   &   0.24  &   87.08                       \\
\hline
Ours                     & whole image                                   &  \textbf{0.53}   &   \textbf{95.80}                    &  \textbf{2.24}   &   \textbf{59.43}                       &   	\textbf{6.22}   &   \textbf{90.78}                     &  \textbf{0.40}   &    \textbf{88.61}                      \\
\rowcolor{lightgray}
NFlowJS (w/ GF)~\cite{grcic2021dense}               & road/sidewalks pixels                                   &  100   &  50.11                     & 1.96    &  69.43
                        &  -   &  -                      &  -   &  -                        \\

\specialrule{1.5pt}{0pt}{0pt}
\end{tabular}
}
\end{table*}
\vspace{-5pt}
\subsection{The detailed architecture of RPL and CoroCL}
As shown in~\cref{fig:detailed_arch}, we design our proposed RPL module based on the Atrous Spatial Pyramid Pooling (ASPP)~\cite{chen2018encoder,chen2017rethinking} block in~\cite{zhu2019improving}, followed by one convolutional head for CoroCL and one for RPL. During training, the incoming feature (with \textbf{4096} channels) will go through a set of convolutional layers that have different dilation rates which produce a set of features that are concatenated to form the feature map with depth \textbf{1280}. There is one more convolutional layer to extract the information from such concatenated feature map and reduce the channels to \textbf{256}. Finally, the heads of CoroCL and RPL will produce the outputs with \textbf{304} depth for training.

\begin{table*}[t!]
\centering
\caption{\textbf{Improvements for different backbones} on Fihsyscapes, SMIYC and RoadAnomaly validation sets. ``Before" represents the pixel-wise anomaly detection performance based on the closed-set segmentation model, while ``After" denotes the results after the training of RPL with CoroCL. We use \textcolor{red}{red} to represent a decrease and \textcolor{teal}{green} to represent an increase in the ``Improve" row and the results reported in the main paper are in \textbf{boldface}.}
\vspace{-5pt}
\label{tab:backbones}
\renewcommand{\arraystretch}{1.2}
\resizebox{\linewidth}{!}{
\begin{tabular}{!{\vrule width 1.5pt}c|c!{\vrule width 1.5pt}ccc|ccc!{\vrule width 1.5pt}ccc|ccc!{\vrule width 1.5pt}ccc!{\vrule width 1.5pt}} 
\specialrule{1.5pt}{0pt}{0pt}
\multicolumn{2}{!{\vrule width 1.5pt}c!{\vrule width 1.5pt}}{\multirow{3}{*}{Backbone}} & \multicolumn{6}{c!{\vrule width 1.5pt}}{Fishyscapes}                      & \multicolumn{6}{c!{\vrule width 1.5pt}}{SMIYC}                                   & \multicolumn{3}{c!{\vrule width 1.5pt}}{\multirow{2}{*}{RoadAnomaly}}  \\ 
\cline{3-14}
\multicolumn{2}{!{\vrule width 1.5pt}c!{\vrule width 1.5pt}}{}                          & \multicolumn{3}{c|}{Static} & \multicolumn{3}{c!{\vrule width 1.5pt}}{L\&F} & \multicolumn{3}{c|}{Anomaly} & \multicolumn{3}{c!{\vrule width 1.5pt}}{Obstacle} & \multicolumn{3}{c!{\vrule width 1.5pt}}{}                              \\ 
\cline{3-17}
\multicolumn{2}{!{\vrule width 1.5pt}c!{\vrule width 1.5pt}}{}                          & FPR$\downarrow$  & AP$\uparrow$  & ROC$\uparrow$         & FPR$\downarrow$  & AP$\uparrow$ & ROC$\uparrow$     & FPR$\downarrow$  & AP$\uparrow$ & ROC$\uparrow$          & FPR$\downarrow$  & AP$\uparrow$ & ROC$\uparrow$           & FPR$\downarrow$  & AP$\uparrow$ & ROC$\uparrow$                                \\ 

\specialrule{1.5pt}{0pt}{0pt}
\multirow{3}{*}{MobileNet}     & Before         &  47.94  &  17.45     &   88.84            & 42.87    & 6.32       & 90.56                 & 46.19    &  51.92     &  87.17               &  6.86   &  52.18     &  98.32             &  67.81   &   20.27    &  73.76                              \\
& After          &  18.64   & 74.42       &  96.89             & 20.77    &  48.54     & 96.59          & 26.74    &   63.52    & 91.43                & 3.26    & 80.07      & 99.23                &  38.11   &  62.49     & 91.72                                     \\ 
& Improve          &  \textcolor{red}{29.30}   & \textcolor{teal}{57.27}       &  \textcolor{teal}{8.05}             & \textcolor{red}{22.10}    &  \textcolor{teal}{42.22}     & \textcolor{teal}{6.03}          & \textcolor{red}{19.45}    &   \textcolor{teal}{11.60}    & \textcolor{teal}{4.26}                & \textcolor{red}{3.60}    & \textcolor{teal}{27.89}      & \textcolor{teal}{0.91}                &  \textcolor{red}{29.70}   &  \textcolor{teal}{42.22}     & \textcolor{teal}{17.96}                                     \\ 
                               
\hline
\multirow{3}{*}{ResNet50}     & Before              & 46.66    &   28.64    &  89.01        & 42.04    & 10.15      &  91.24    & 65.75    & 46.46      &  81.07              &  6.55    & 49.12     &  91.33               & 67.61    &  22.08     &   72.78                                   \\
                               & After          & 5.69    &     87.27  &    99.07           & 16.78    &  49.92      &  97.78         & 22.51    &  72.18     &     94.08           & 2.62    &74.40       &     99.42            &  26.18   &  63.96     &   93.24                                   \\ 
& Improve          & \textcolor{red}{40.97}    &     \textcolor{teal}{58.63}  &    \textcolor{teal}{10.06}           & \textcolor{red}{25.26}     &  \textcolor{teal}{39.77 }      &  \textcolor{teal}{6.54}         & \textcolor{red}{43.24}    &  \textcolor{teal}{25.72}     &     \textcolor{teal}{13.01}           & \textcolor{red}{3.93}    & \textcolor{teal}{25.28}       &     \textcolor{teal}{8.09}            &   \textcolor{red}{41.43}    &  \textcolor{teal}{41.88}     &  \textcolor{teal}{20.46}                                   \\ 
\hline
\multirow{3}{*}{ResNet101}     & Before         & 42.85    & 30.15      &  90.16             & 38.07    &   24.57    &  92.36         & 44.92    & 53.91      &  86.50              &  23.75   & 13.30     &  93.78               & 66.21    &  24.05     &   77.25                                   \\
& After          &   1.61  & 89.88      &  99.14             & 8.82     & 60.08      & 98.84          & 15.13    &  74.83     &   95.14           & 2.46    &  78.78     &  99.66                 & 24.54    &  65.42     &   94.24                                   \\ 
& Improve          &   \textcolor{red}{41.24}  & \textcolor{teal}{59.73}      &  \textcolor{teal}{8.98}             & \textcolor{red}{29.25}    & \textcolor{teal}{35.51}      & \textcolor{teal}{6.48}         &  \textcolor{red}{29.79}   & \textcolor{teal}{20.92}      &   \textcolor{teal}{9.64}             & \textcolor{red}{21.29}    &  \textcolor{teal}{65.48}     &   \textcolor{teal}{5.88}              &  \textcolor{red}{41.67}   &  \textcolor{teal}{41.37}     &   \textcolor{teal}{16.99}                                   \\ 
\hline
\multirow{3}{*}{WiderResNet38} & Before         &   17.78  &  41.68     &  95.90             &  41.78   & 16.05      &  93.72         &   67.75  &  44.54     &  80.26              &  4.50   &   34.44    &  99.67               &  69.99   & 19.95      &  73.61                                    \\
                               & After           &  \textbf{0.85}   &  \textbf{92.46}     &  \textbf{99.73}             & \textbf{2.52}   & \textbf{70.61}      &  \textbf{99.39}         &  \textbf{7.18}   &  \textbf{88.55}     &  \textbf{98.06}              &  \textbf{0.09}   & \textbf{96.91}     &  \textbf{99.97}               & \textbf{17.74}    &  \textbf{71.61}     & \textbf{95.72}                                     \\  
                               & Improve           & \textcolor{red}{16.93}    &  \textcolor{teal}{50.78}     &  \textcolor{teal}{3.83}             & \textcolor{red}{39.26}    & \textcolor{teal}{54.56}      &  \textcolor{teal}{5.67}         &  \textcolor{red}{60.57}   &  \textcolor{teal}{44.01}     &  \textcolor{teal}{9.80}              &  \textcolor{red}{4.41}   & \textcolor{teal}{62.47}     &  \textcolor{teal}{1.30}                & \textcolor{red}{52.25}     &  \textcolor{teal}{51.66}     & \textcolor{teal}{22.11}                                     \\  
\specialrule{1.5pt}{0pt}{0pt}
\end{tabular}}
\vspace{-10pt}
\end{table*}

\section{Experiments with Extra Training Set}
\noindent\textbf{Dataset descriptions.} The NFlowJS~\cite{grcic2021dense} and Densehybrid~\cite{grcic2022densehybrid} have additional experimental setups that fine-tune the OoD detector to  extra training sets, including Vistas~\cite{neuhold2017mapillary} and Wilddash2~\cite{zendel2018wilddash}. Vistas~\cite{neuhold2017mapillary} contain $20,000$ images from real-world driving scenes with high resolution ($2592 \times 1944$ pixels) and $66$ categories of finely-annotated pixel-wise labels. Similarly, Wilddash2~\cite{zendel2018wilddash} is another driving scene dataset containing $4,255$ images with $80$ categories in total, where each image has $1920\times1080$ pixels. Given that the experimental setup presented in our submitted main paper only utilises Cityscapes~\cite{cordts2016cityscapes} (i.e., $29,75$) images, fine-tuning the OoD detector with those extra training sets enables better robustness to hard inliers.  \\
\noindent\textbf{Results from Fishyscape\protect\footnote{\url{https://fishyscapes.com/results}} and SMIYC\protect\footnote{\url{https://segmentmeifyoucan.com/leaderboard}}.} To enable a fair comparison, we follow \cite{grcic2022densehybrid, grcic2021dense} to fine-tune our RPL module with $10$ epochs for both Vistas~\cite{neuhold2017mapillary} and Wilddash2~\cite{zendel2018wilddash}. \cref{tab:extra_data} shows that our method outperforms other approaches under the same setup. For example, our AuPRC results are 12.82\% and 1.53\% higher than Densehybrid~\cite{grcic2022densehybrid}  on SMIYC-Anomaly and SMIYC-Obstacle subsets, respectively.\\
\noindent The \textit{Ground-Focus (GF) post-process} in the \textcolor{gray}{last row} of~\cref{tab:extra_data} merges all road and sidewalk pixels to a common "ground" class by creating a convex hull that encapsulates all such pixels. During inference, all the pixels outside this hull will be ignored, producing significant improvements for Fishyscapes-Lost\&Found dataset. However, real-world anomalies (e.g., birds) might not be located on the "ground", reducing its practicability. For example, Fishyscapes-Static has anomalies outside the road that are never detected, leading to unsatisfactory performance. In addition, the inaccurate prediction of the road/sidewalks categories also results in the misdetection of the anomalies.  
\begin{table*}[t!]
\caption{\textbf{Ablations for CoroCL on Fishyscapes, SMIYC and RoadAnomaly validation sets.} We define the OoD, inlier with \{\OodCoco,\InlierCoco\} in COCO context and \{\OodCity, \InlierCity\} in city context. The best performance are in bold.}
\label{tab:ablation_proj}

\vspace{-5pt}
\centering
\renewcommand{\arraystretch}{1.25}
\resizebox{\linewidth}{!}{
\begin{tabular}{!{\vrule width 1.5pt}c|c!{\vrule width 1.5pt}ccc|ccc!{\vrule width 1.5pt}ccc|ccc!{\vrule width 1.5pt}ccc!{\vrule width 1.5pt}} 
\specialrule{1.5pt}{0pt}{0pt}
\multirow{3}{*}{\textbf{Anchor}} & \multirow{3}{*}{\textbf{Contrastive}} & \multicolumn{6}{c!{\vrule width 1.5pt}}{\textbf{Fishyscapes}}                      & \multicolumn{6}{c!{\vrule width 1.5pt}}{\textbf{SMIYC}}                                   & \multicolumn{3}{c|}{\multirow{2}{*}{\textbf{RoadAnomaly}}}  \\ 
\cline{3-14}
                             &                                   & \multicolumn{3}{c|}{Static} & \multicolumn{3}{c!{\vrule width 1.5pt}}{L\&F} & \multicolumn{3}{c|}{Anomaly} & \multicolumn{3}{c!{\vrule width 1.5pt}}{Obstacle} & \multicolumn{3}{c|}{}                              \\ 
\cline{3-17}
                             &                                   & FPR & AP & ROC         & FPR & AP & ROC     & FPR & AP & ROC          & FPR & AP & ROC           & FPR & AP & ROC                                \\ 
\specialrule{1.5pt}{0pt}{0pt}

                              \InlierCity \OodCity    &  \InlierCity \OodCity                             &  1.43  &    91.23   & 99.47              & 4.26     &    67.37   &  98.10         & 21.91    &    77.76   &  94.91              & 6.37    & 92.13       &  99.81               & 25.84    & 62.61      &    93.70                                  \\
                         \InlierCoco \OodCoco    &  \InlierCoco \OodCoco  & 1.70     & 88.72      & 99.57              & 6.97     & 56.84      & 98.70          & 9.41    & 86.62      &97.57                & 0.11     & 95.92      & 99.94                &  21.0   &   69.89    &   95.70                                   \\  
                         
                          \InlierCity \OodCity    &    \InlierCoco \InlierCity \OodCity                                             &  1.79  &    89.90   &    99.52           & 3.74     & 68.17      & 98.95          & 11.31    &     86.37  & 97.27               & 0.29    & 94.31      & 99.93                & 26.78    & 66.91      & 93.87                                     \\              
                          
                         
                         \InlierCity \OodCity    &   \InlierCity \OodCity \InlierCoco \OodCoco                              &  \textbf{0.85}   &  \textbf{92.46}     &  \textbf{99.73}             & \textbf{2.52}    & \textbf{70.61}      &  \textbf{99.39}         &  \textbf{7.18}   &  \textbf{88.55}     &  \textbf{98.06 }             &  \textbf{0.09}   & \textbf{96.91}     &  \textbf{99.97}               & \textbf{17.74}    &  \textbf{71.61}     & \textbf{95.72}                                     \\
                         
                         \InlierCity \OodCity \InlierCoco \OodCoco    &   \InlierCity \OodCity \InlierCoco \OodCoco                                &  1.57   &  90.24     & 99.58              & 4.90    &  60.72     & 98.53           & 15.29    & 82.94      &   97.18             & 0.10      & 96.58      & 99.96                &  19.62   & 68.97       &  95.44                                    \\

\specialrule{1.5pt}{0pt}{0pt}
\end{tabular}}
\end{table*}
\section{More Implementation Details}
We provide more implementation details in this section. \textbf{In the training of RPL}, we partially load the parameters from the pre-trained ASPP block in DeepLabV3+~\cite{chen2018encoder,chen2017rethinking} to the main RPL module, as they share the same architecture. We initialise the convolutional head for RPL based on~\cite{he2015delving} and we apply $10$ times the learning rate (with $7.5e^{-4}$) to the head compared with other convolutional layers that are trained. The images from Cityscape~\cite{cordts2016cityscapes} are randomly cropped with $700 \times 700$ resolution, while the COCO~\cite{lin2014microsoft} images are randomly scaled with ratio in $\{.1, .125, .25, .5, .75\}$ and then mixed with the city images based on outlier exposure (OE)~\cite{tian2021pixel}. Meanwhile, we copy the vanilla COCO images based on padding or centre cropping to the same resolution of $700 \times 700$ as city images for contrastive learning. \textbf{In the training of CoroCL}, we concatenate the context images from COCO and Cityscapes and extract the embeddings of both contexts via single forward propagation. 
We randomly choose \textbf{512} pixel-wise samples from both inlier and OoD in city and COCO contexts to perform CoroCL based on the Eq. \textcolor{red}{(7)} (from the main paper), where $\tau=0.10$ for all the experiments.

We train the RPL module with one Tesla V100 16GB and RPL+CoroCL with one RTX A6000, as the contrastive learning needs more GPU memory. Following previous works~\cite{khosla2020supervised, wang2021exploring}, we discard the projector head after training and directly utilise RPL outputs to induce the closed-segmentation to produce high-uncertainty in potential anomalous regions.

\begin{table}[t!]
\caption{\textbf{The impact of the projector architecture} on the SMIYC-Anomaly and RoadAnomaly datasets.}
\vspace{-8 pt}
\label{tab:projector_arch}
\resizebox{\linewidth}{27pt}{
\begin{tabular}{!{\vrule width 1.5pt}l!{\vrule width 1.5pt}c|c|c!{\vrule width 1.5pt}c|c|c!{\vrule width 1.5pt}} 
\specialrule{1.5pt}{0pt}{0pt}
\multicolumn{1}{!{\vrule width 1.5pt}c!{\vrule width 1.5pt}}{\multirow{2}{*}{Architecture}} & \multicolumn{3}{c!{\vrule width 1.5pt}}{SMIYC-Anomaly} & \multicolumn{3}{c!{\vrule width 1.5pt}}{RoadAnomaly}  \\ 
\cline{2-7}
                              & FPR   & AuPRC & AuROC              & FPR   & AuPRC & AuROC             \\ 
\specialrule{1.5pt}{0pt}{0pt}
2 layers (w/o BN)             & 14.56 & 82.08 & 95.51              & 21.94 & 62.59 & 94.47             \\
2 layers (w/ BN)              & 13.72 & 83.24 & 95.95              & 21.11 & 63.81 & 94.53             \\ 
\hline
\textbf{single-layer}                       & \textbf{7.18}  & \textbf{88.55} & \textbf{98.06}              & \textbf{17.74} & \textbf{71.61} & \textbf{95.72}             \\
\specialrule{1.5pt}{0pt}{0pt}
\end{tabular}}
\vspace{-10pt}
\end{table}

\section{Results from Different Backbones}
\cref{tab:backbones} displays the results of our approach with different backbones, while we measure them based on the area under the receiver operating characteristics (AuROC), average precision (AP), and false positive rate at a true positive rate of 95\% (FPR). We report the closed-set segmentation performance in ``Before" and our performance in ``After", while the  improvements in all backbones demonstrate the generalisation of our approach. For example, our approach improves the performance by 29.70\%, 41.43\% and 52.25\% FPR in the RoadAnomaly validation set for MobileNet, ResNet50 and WiderResNet38, respectively.

\begin{figure}[t!]
    \centering
    \begin{subfigure}[b]{0.3\linewidth}
    \includegraphics[width=\textwidth]{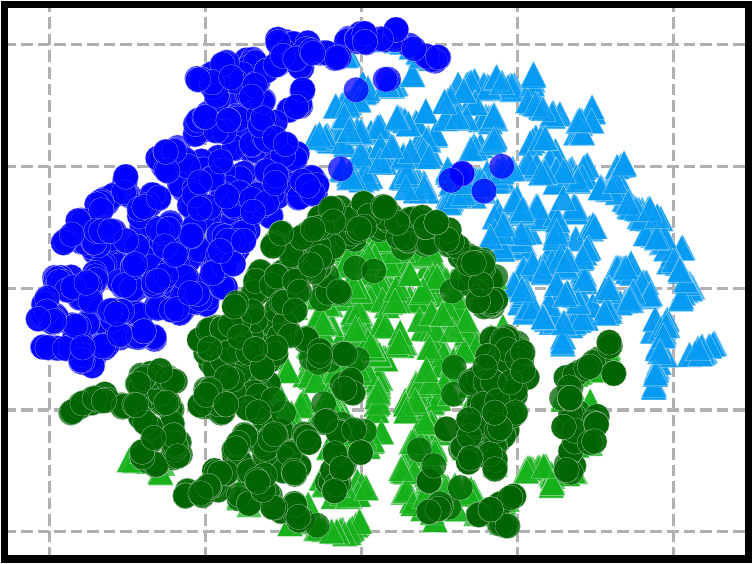} \hfill
    \includegraphics[width=\textwidth]{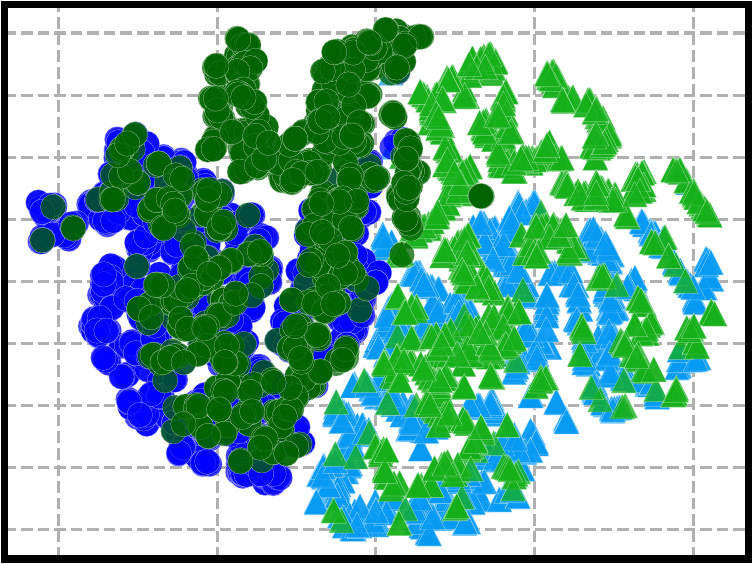}
    \vspace{-18pt}
    \caption*{case (a)}
    \end{subfigure}
    \begin{subfigure}[b]{0.3\linewidth}
    \includegraphics[width=\linewidth]{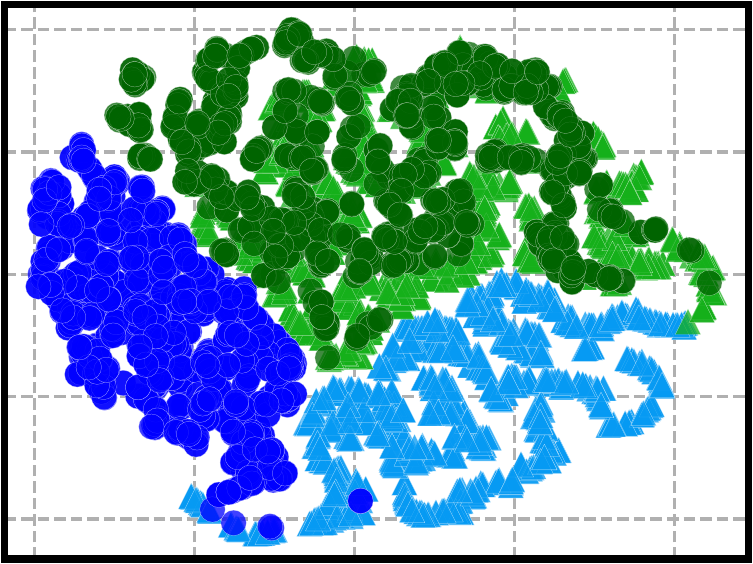} \hfill
    \includegraphics[width=\linewidth]{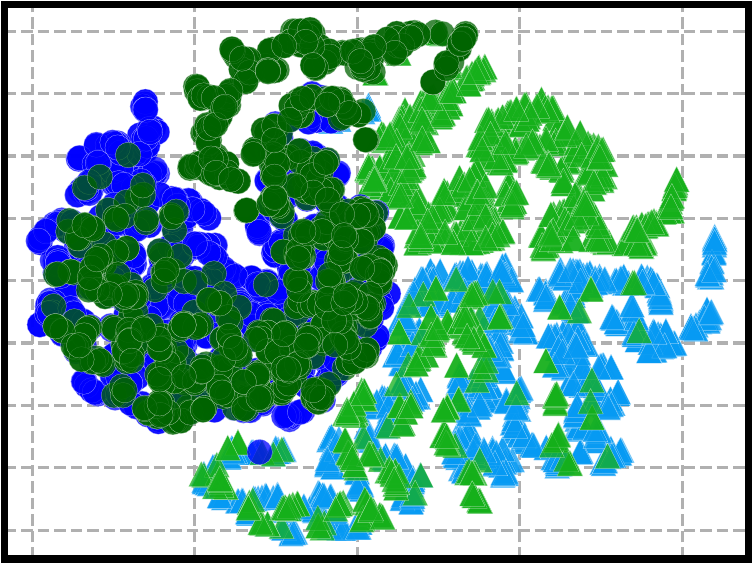} 
    \vspace{-18pt}
    \caption*{case (b)}
    \end{subfigure}
    \begin{subfigure}[b]{0.3\linewidth}
    \includegraphics[width=\linewidth]{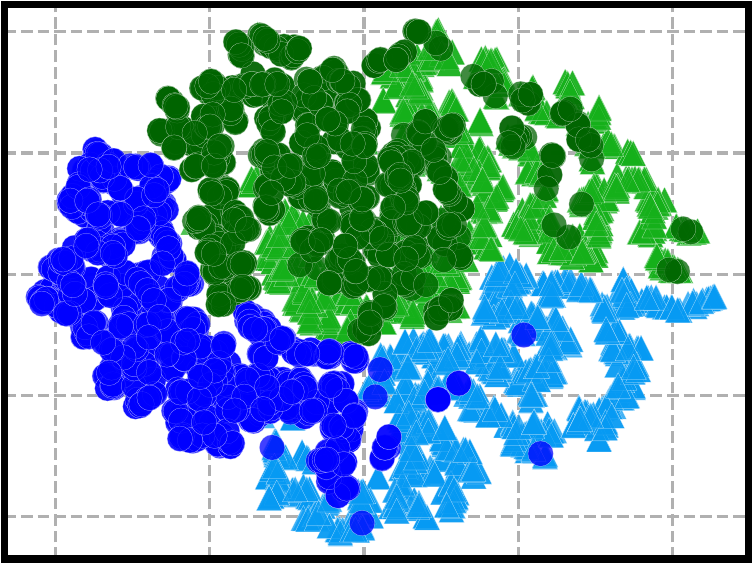} \hfill
    \includegraphics[width=\linewidth]{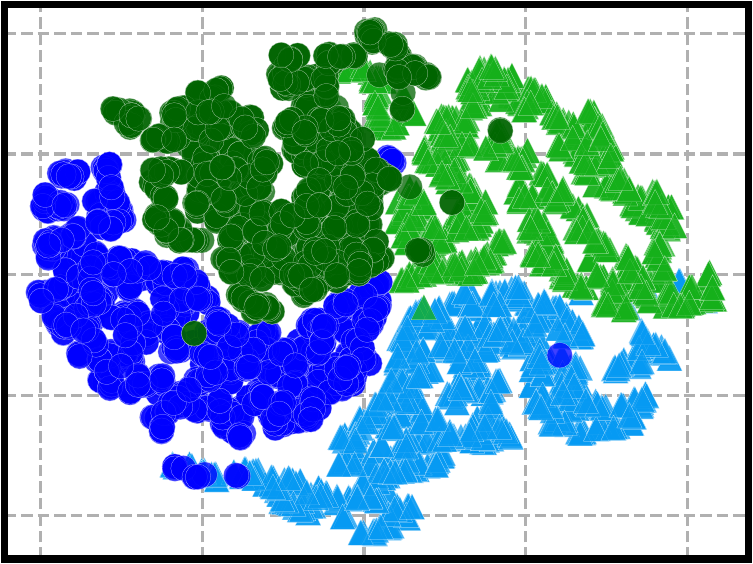} 
    \vspace{-18pt}
    \caption*{case (c)}
    \end{subfigure}
    
    \caption*{Fig \textcolor{red}{$\dagger$}: T-SNE visualisation of RPL outputs \textbf{w/o CoroCL (first row)} and \textbf{with CoroCL (second row)}. Each column uses the same images, where city and non-city contexts \ul{inliers} are \mycircle{city_inlier} and \mycircle{ood_inlier}, while city and non-city contexts \ul{outliers} are \mytriangle{city_outlier} and \mytriangle{ood_outlier}. Better viewed in zoomed-in mode.}
    \vspace{-15pt}
\end{figure}

\begin{figure*}
\begin{subfigure}[b]{0.49\linewidth}
    \centering
    \includegraphics[width=.492\textwidth, height=2.5cm]{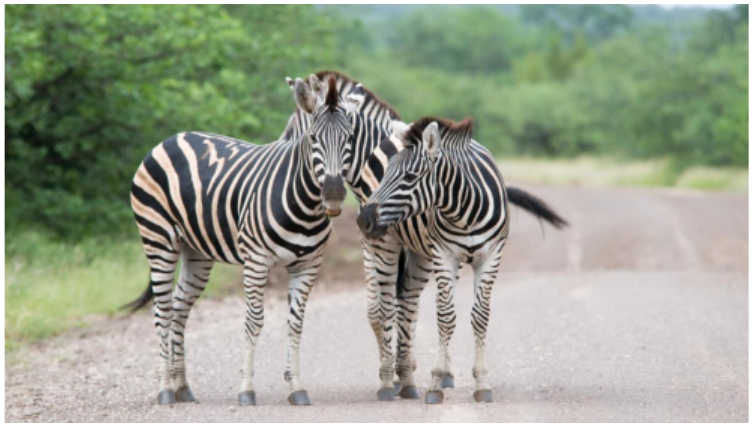} \hfill
    \includegraphics[width=.492\textwidth, height=2.5cm]{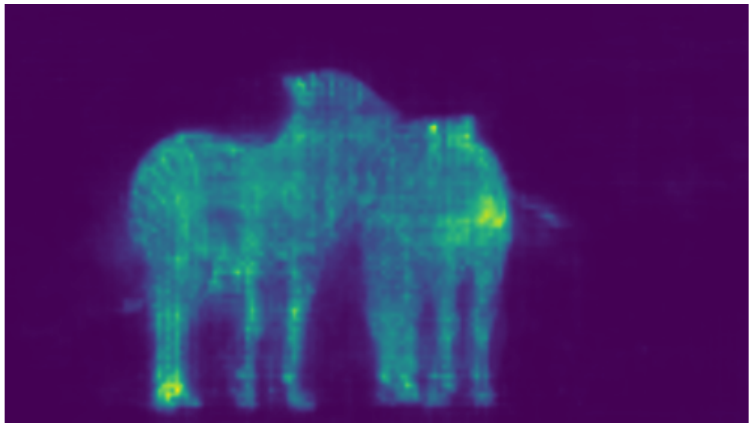} 
        \includegraphics[width=.492\textwidth, height=2.5cm]{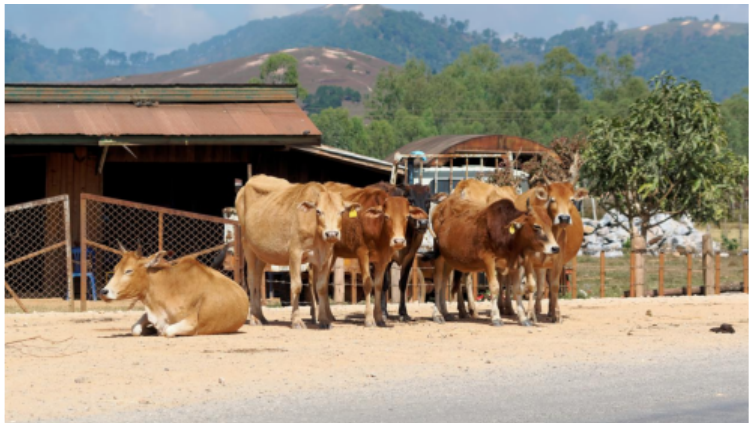} \hfill
    \includegraphics[width=.492\textwidth, height=2.5cm]{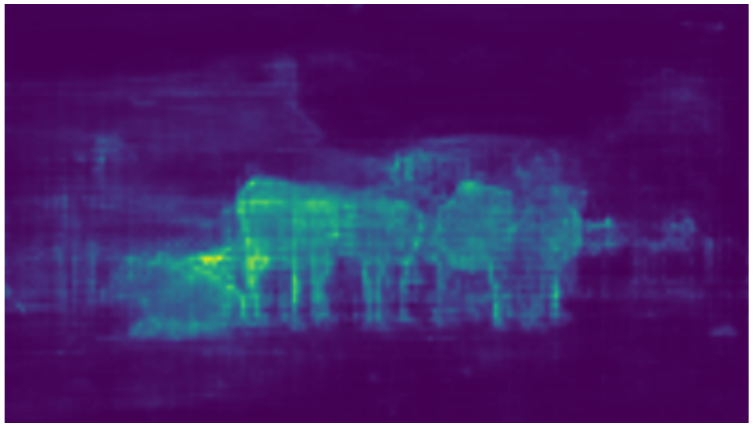} 
    \includegraphics[width=.492\textwidth, height=2.5cm]{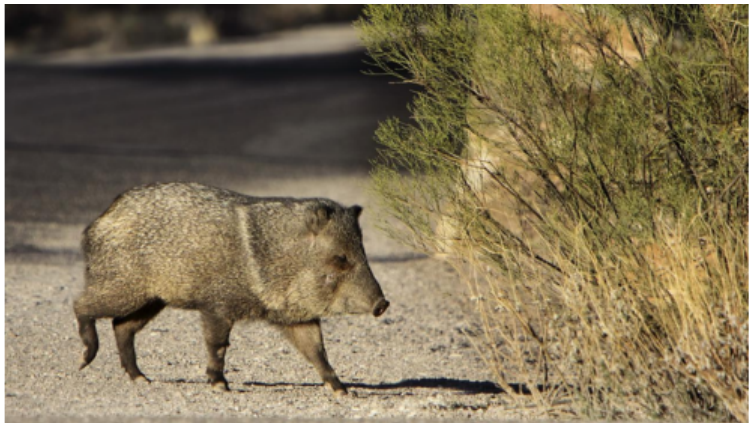} \hfill
    \includegraphics[width=.492\textwidth, height=2.5cm]{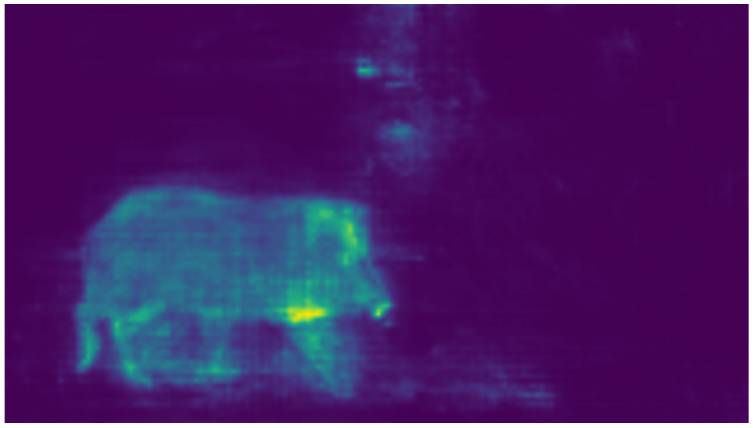} 
\end{subfigure}
\begin{subfigure}[b]{0.49\linewidth}
    \centering
    \includegraphics[width=.492\textwidth, height=2.5cm]{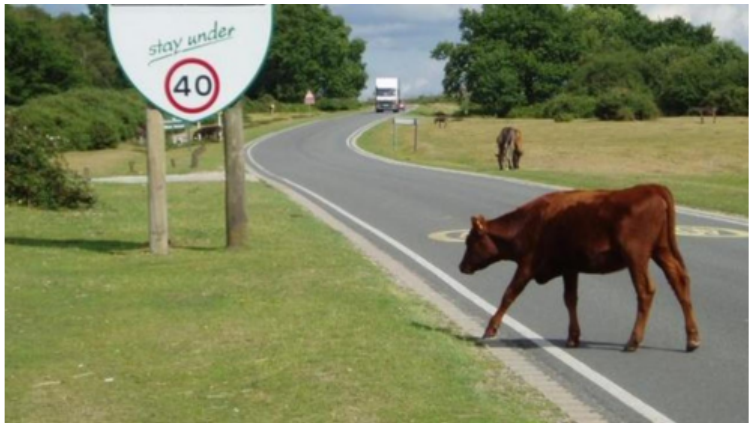} \hfill
    \includegraphics[width=.492\textwidth, height=2.5cm]{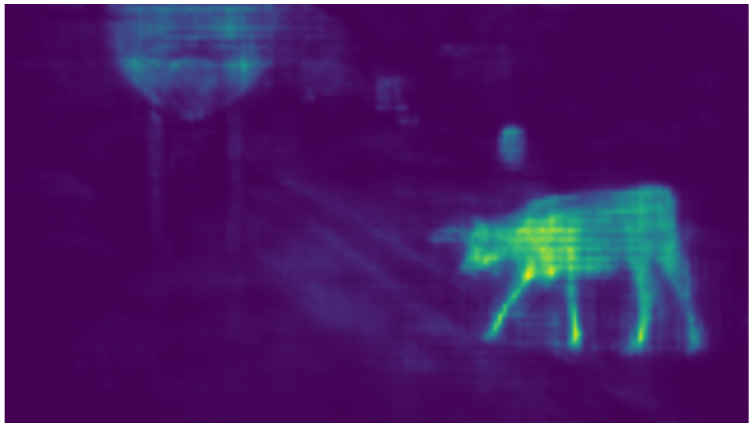} 
    \includegraphics[width=.492\textwidth, height=2.5cm]{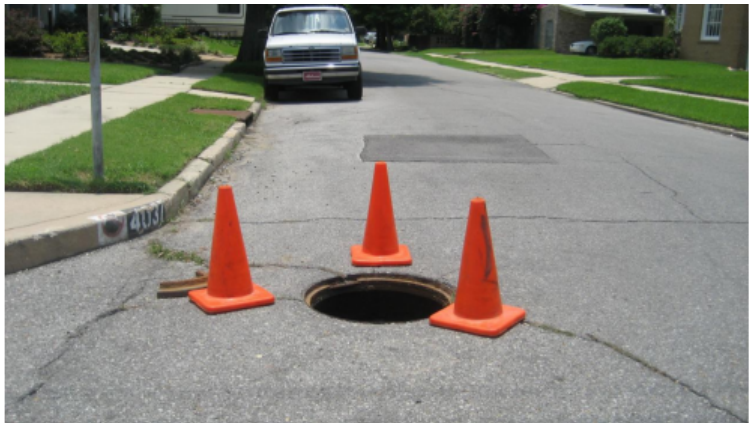} \hfill
    \includegraphics[width=.492\textwidth, height=2.5cm]{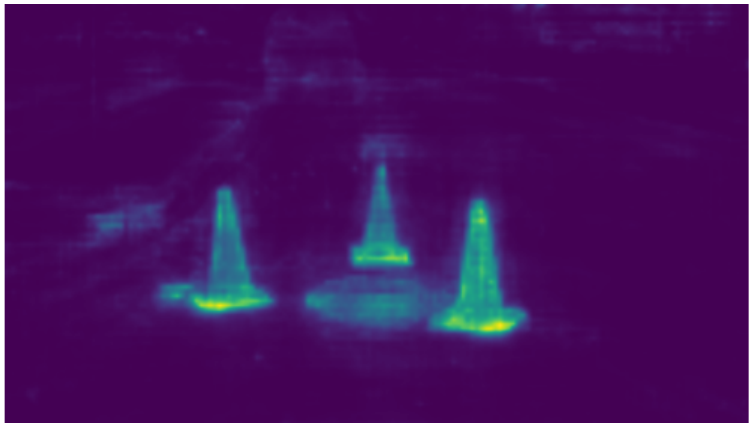}  
    \includegraphics[width=.492\textwidth, height=2.5cm]{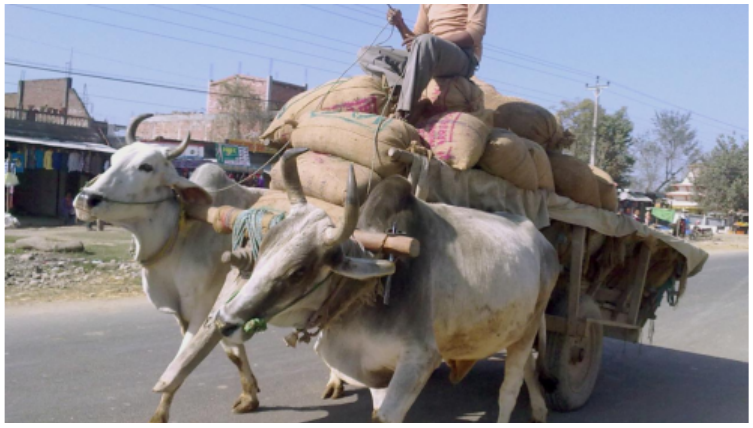} \hfill
    \includegraphics[width=.492\textwidth, height=2.5cm]{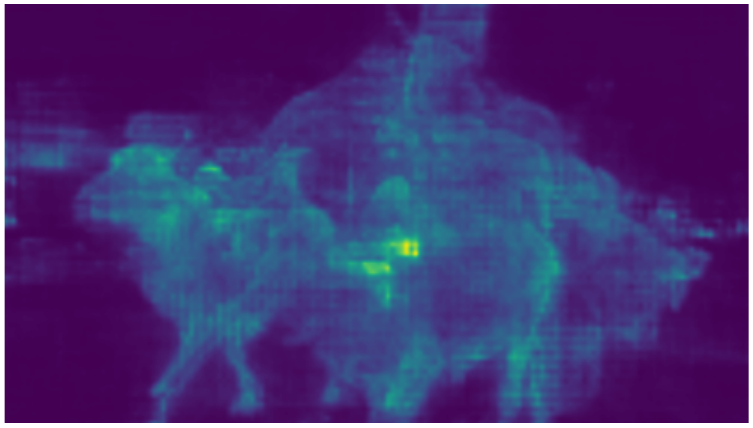} 
\end{subfigure}
\vspace{-5pt}
\caption{The self-attention results of the learned residual pattern feature from RoadAnomaly~\cite{lis2019detecting}.}
\vspace{-15pt}
\label{fig:rpl}
\end{figure*}

\section{More Details of CoroCL}
\noindent \textbf{Impact of Projector Architecture.} Differently from previous contrastive learning methods~\cite{wang2021dense,wang2021exploring}, we find that a projector with a single-layer performs better than with a multi-layer, as shown in~\cref{tab:projector_arch}, which may be due to a better robustness to overfitting given the smaller number of layers. \\
\noindent \textbf{Construction of Anchor and Contrastive Sets}
We implement the ablation of the anchor and contrastive sets based on AuROC, AP and false positive rate at a true positive rate of 95\% (FPR) in~\cref{tab:ablation_proj}.
The pixel-wise embedding samples in our training have OoD (\OodCoco) and inlier (\InlierCoco) in the COCO context and OoD (\OodCity) and inlier (\InlierCity) in the city context, as shown in Fig. \textcolor{red}{2} in the main paper. The choice of the samples that build the anchor and contrastive sets will heavily impact the final performance. For example, using city context samples \{\InlierCity,\OodCity\} for both anchor and contrastive sets (in the first row) yields great performance in Fishyscapes (e.g., 91.23\% AP in Static and 67.37\% AP in L\&F) but the poor performance in SMIYC. On the contrary, using COCO context samples \{\InlierCoco,\OodCoco\} for both anchor and contrastive sets (in the second row) improves the results by 12.5\% and 6.26\% FPR in both Anomaly and Obstacle of SMIYC but demonstrates worse performance in Fishyscapes. The best performance is observed when we use \{\InlierCity,\OodCity\} to construct anchor set and \{\InlierCity, \OodCity, \InlierCoco, \OodCoco\} to be the contrastive set, which achieves the reported performance in the main paper.
Compared with our results (in the fourth row of Tab. \textcolor{red}{3.}), the last row additionally enforces \InlierCoco $\rightarrow$ $\leftarrow$ \InlierCoco and \InlierCoco $\leftarrow$ $\rightarrow$ \OodCoco (based on Eq. \textcolor{red}{7} in the main paper). Due to the training and validation datasets based on the driving scenes, we suspect that the optimisation applied to the daily natural images (i.e., COCO contexts) damages the convergence of our approach, which yields unsatisfactory performance. \\
\textbf{T-SNE visulaisation.} As shown in Fig. \textcolor{red}{$\dagger$}, we apply T-SNE on the outputs of RPL block for both city and other context images. Using the same images (each column), we randomly sample $\mathbf{4000}$ pixel-wise RPL embeddings. We observe the RPL results \textbf{w/o CoroCL (first row)} can only separate anomalies in city contexts (\mycircle{city_inlier} and \mytriangle{city_outlier}), but fail in non-city contexts (\mycircle{ood_inlier} and \mytriangle{ood_outlier}). \textbf{CoroCL (second row)} clusters the inliers from various scenes while pushing the outliers away, independently of city/non-city contexts.

\begin{table}[t!]
\caption{Results on StreetHazards w/ LDN-121 net based on the closed-set checkpoint and evaluation code on {\small \url{https://github.com/matejgrcic/DenseHybrid}}. The Closed/Open sets are measured by mean IoU and we use \textit{energy} to compute anomaly score. $*$ denotes the results from pretrained inlier model for both our approach and~\cite{grcic2022densehybrid}.}
\vspace{-3pt}
\label{tab:exp_street}
\resizebox{\linewidth}{!}{
\begin{tabular}{|c|c|c|c|c|c|c|}
\hline
\multirow{2}{*}{Method} & \multicolumn{3}{c|}{Anomaly Detection} & \multirow{2}{*}{Closed-set} & \multicolumn{2}{c|}{Open-set} \\
\cline{2-4} \cline{6-7}
                        & FPR         & AP         & AuC        &                             & (t5)          & (t6)         \\
\hline
LDN-121*                 &  15.6           &  16.7          &  95.1          &  \textbf{65.0}                           &   39.3           &  44.5   \\
DenseH~\cite{grcic2022densehybrid}                 &   13.0           &  30.2           &  95.6          &  63.0                           &   46.1            &  45.3   \\
\rowcolor{LightCyan}
\textbf{RPL}                  &  \textbf{8.22}           &  \textbf{31.15}          &  \textbf{97.19}          &   \textbf{65.0}                          & \textbf{58.14}             & \textbf{54.38} \\    
\hline
\end{tabular}}
\vspace{-10pt}
\end{table}

\section{Generalization and results in StreetHazard} \vspace{-3pt}
RPL can be easily adopted by other FCN-based architectures by attaching the RPL module before the pixel-wise classifier head. 
For example, we easily attach the RPL module to the LDN-121 segmentation model, with results in Tab~\cref{tab:exp_street}. Based on same pre-trained checkpoint, the RPL's results outperform the previous SOTA Densehybrid~\cite{grcic2022densehybrid} with 4.8\% improvement in FPR and over 10\% mIoU in Open-set evaluation.

\section{Visualisations of Residual Patterns}\vspace{-3pt}
The learned residual patterns from the RPL feature map  {\small$\mathbf{r}\in\mathbb{R}^{304 \times H \times W}$} can be visualised via self-attention and the pytorch-like code is in below: \\
\texttt{\footnotesize{torch.einsum('abc,bca->bc', r, r.permute(1,2,0))}}. \\
The visualisations of such learned residual patterns are shown in~\cref{fig:rpl}, where the anomaly objects are highlighted.

\section{More Visualisations of OoD Maps} 
\begin{figure*}

\begin{subfigure}[b]{0.49\linewidth}
    \centering
    \includegraphics[width=.49\textwidth, height=2.5cm]{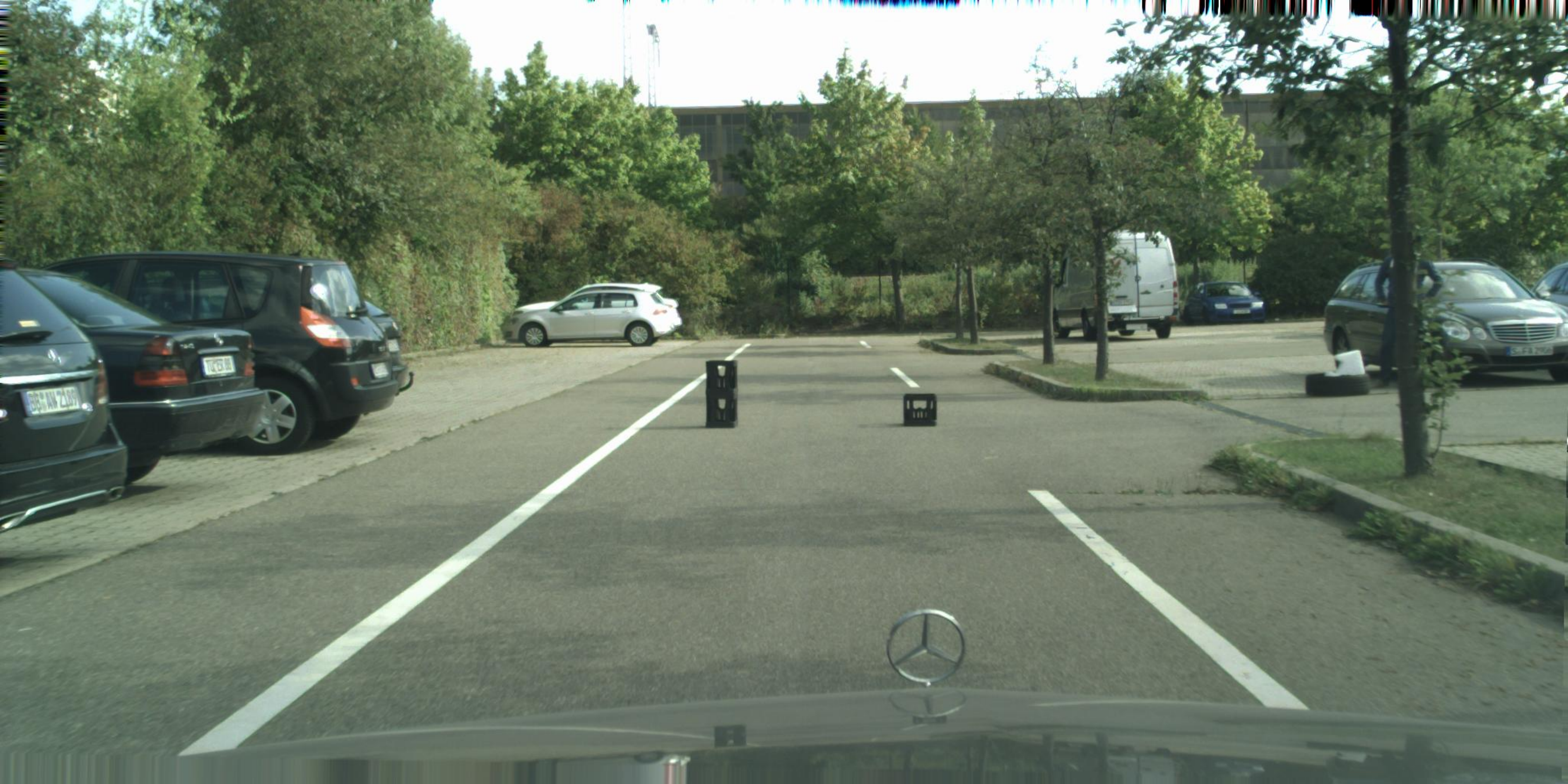} \hfill
    \includegraphics[width=.49\textwidth, height=2.5cm]{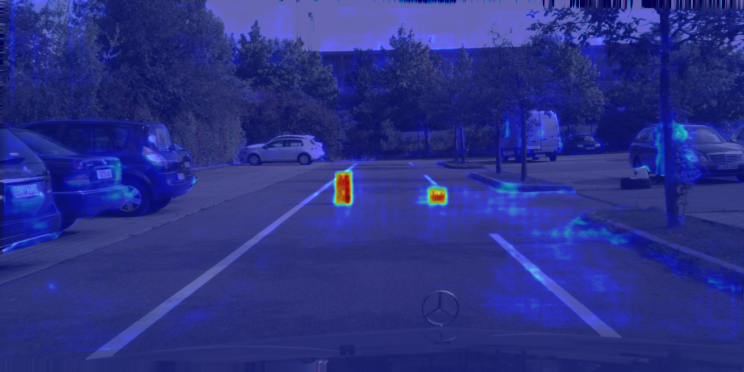}
    \includegraphics[width=.49\textwidth, height=2.5cm]{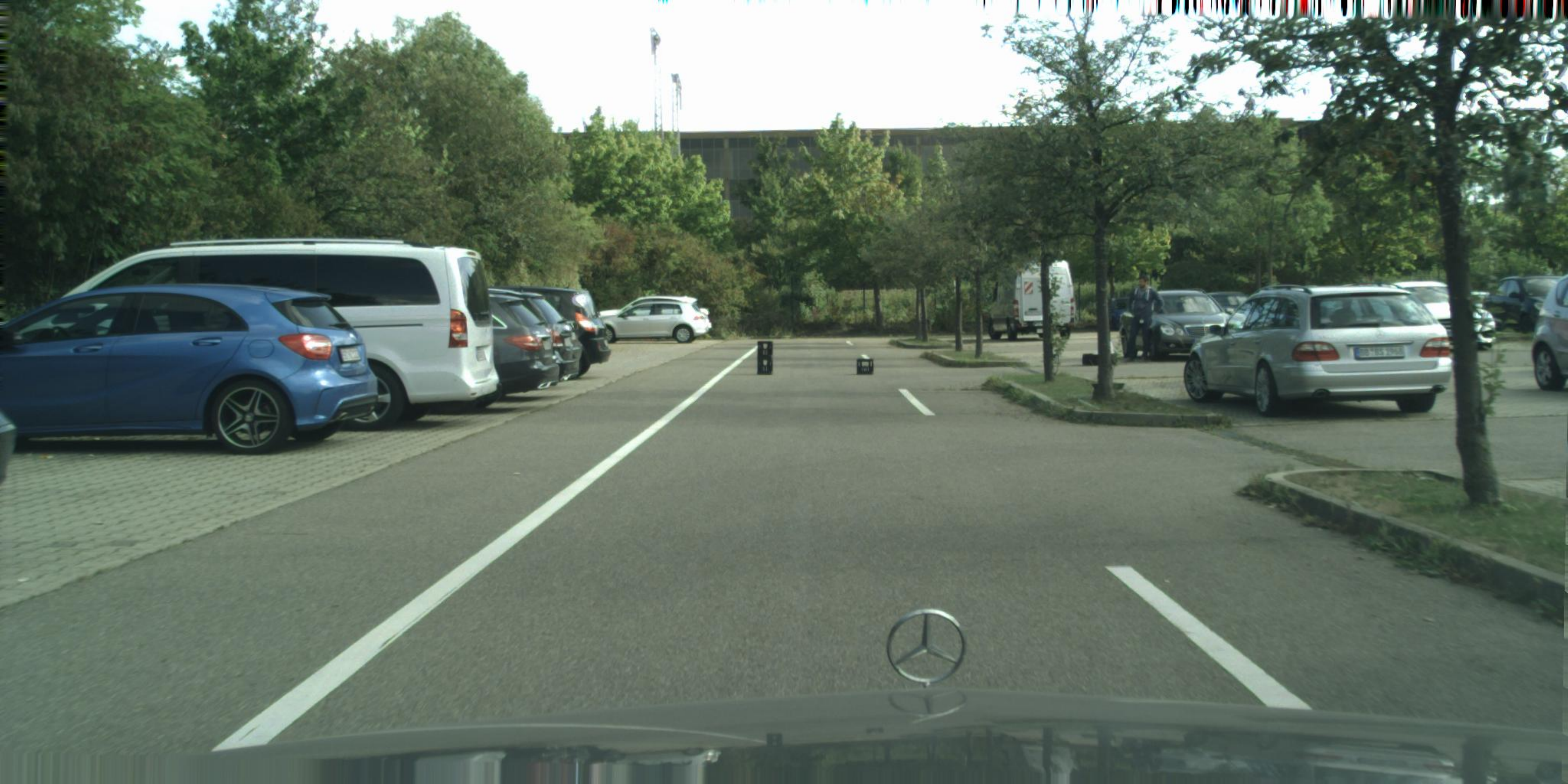} \hfill
    \includegraphics[width=.49\textwidth, height=2.5cm]{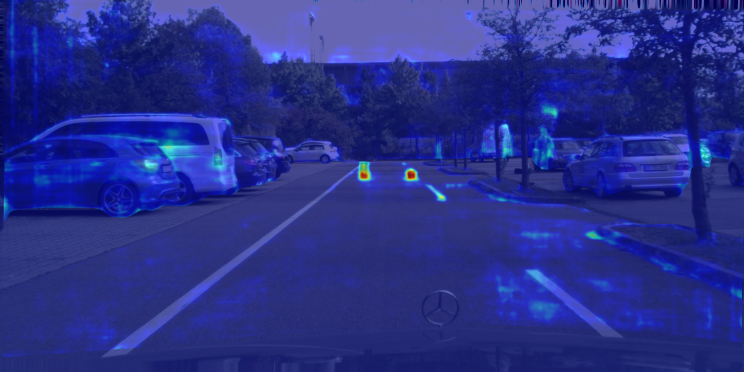}
    \includegraphics[width=.49\textwidth, height=2.5cm]{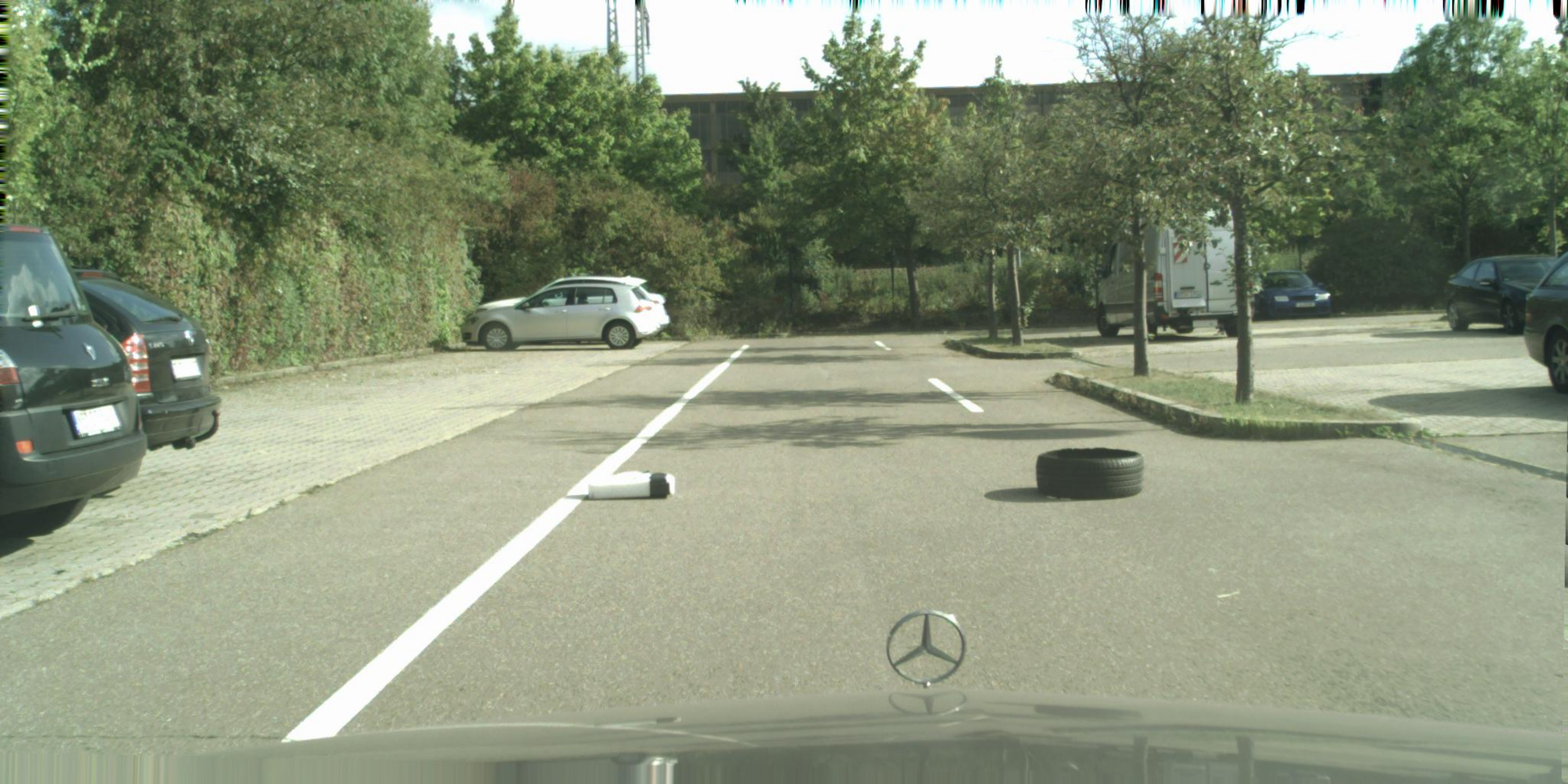} \hfill
    \includegraphics[width=.49\textwidth, height=2.5cm]{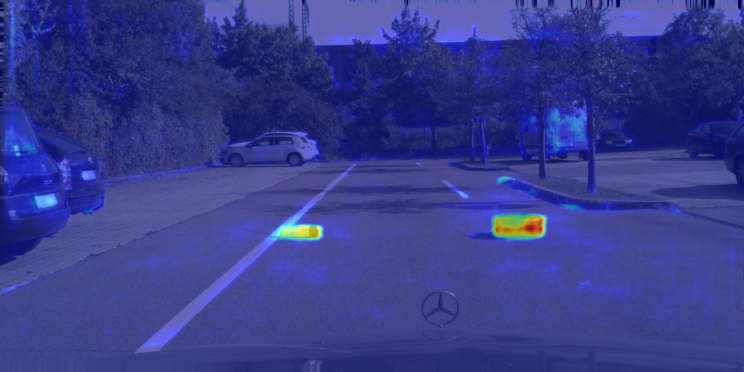}
    \includegraphics[width=.49\textwidth, height=2.5cm]{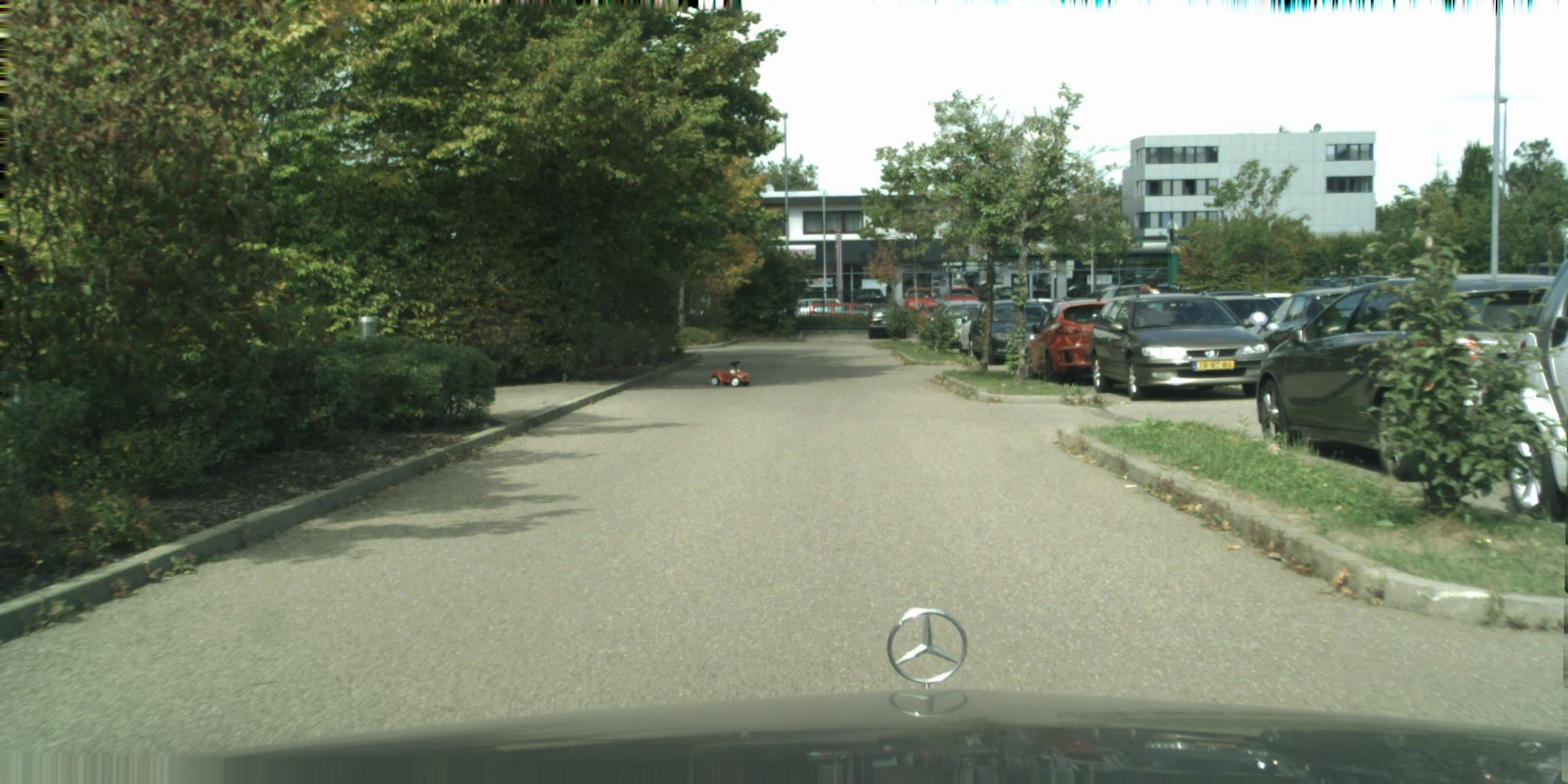} \hfill
    \includegraphics[width=.49\textwidth, height=2.5cm]{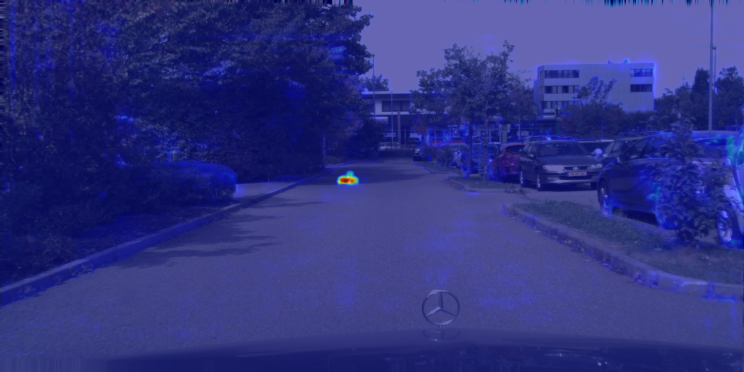}
    \includegraphics[width=.49\textwidth, height=2.5cm]{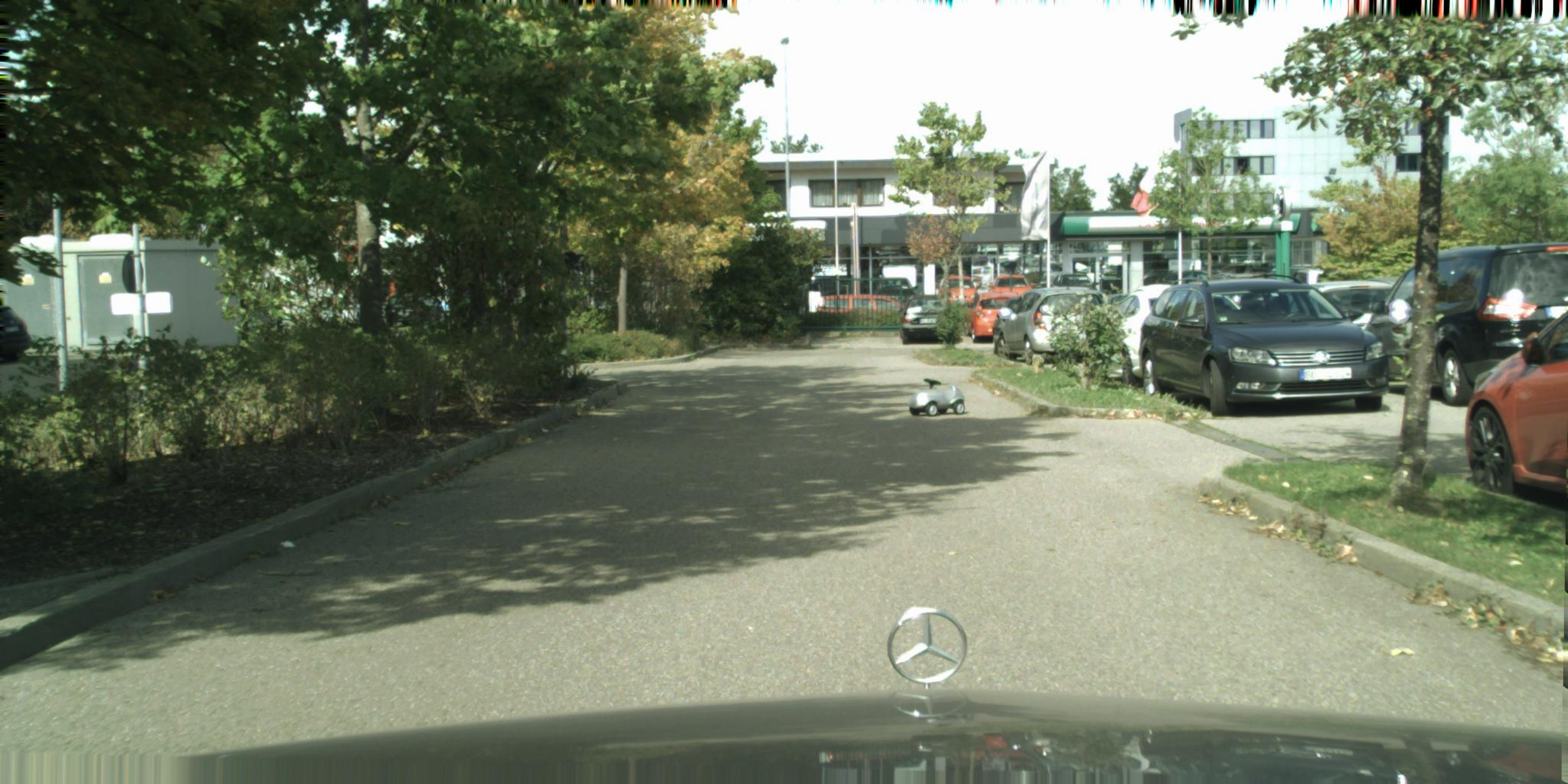} \hfill
    \includegraphics[width=.49\textwidth, height=2.5cm]{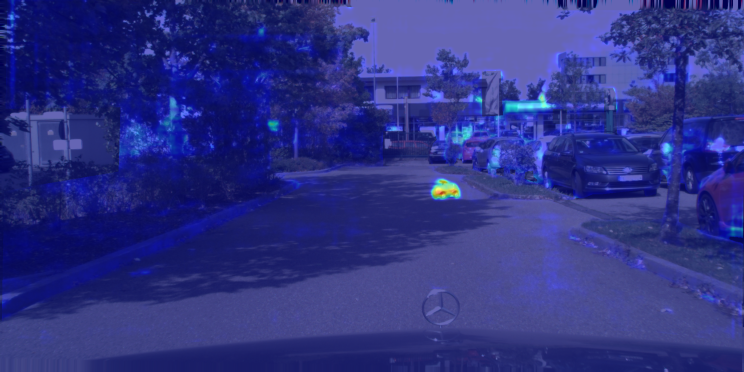}
    \includegraphics[width=.49\textwidth, height=2.5cm]{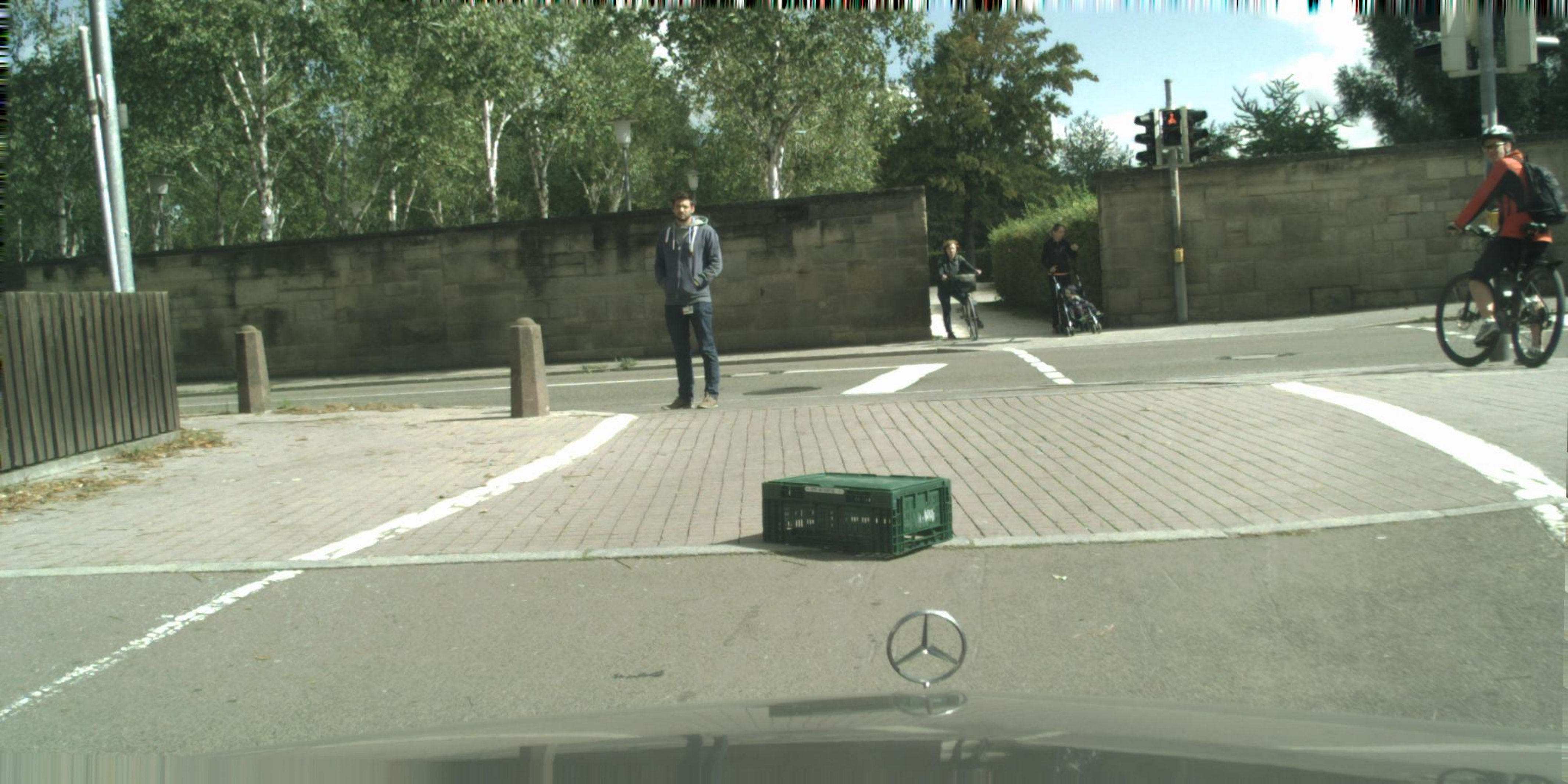} \hfill
    \includegraphics[width=.49\textwidth, height=2.5cm]{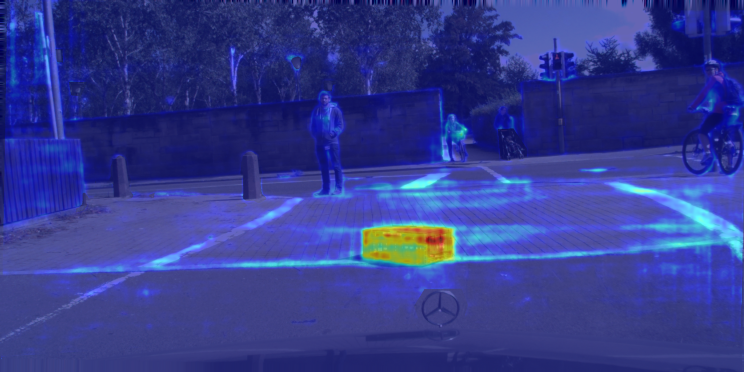}
    \includegraphics[width=.49\textwidth, height=2.5cm]{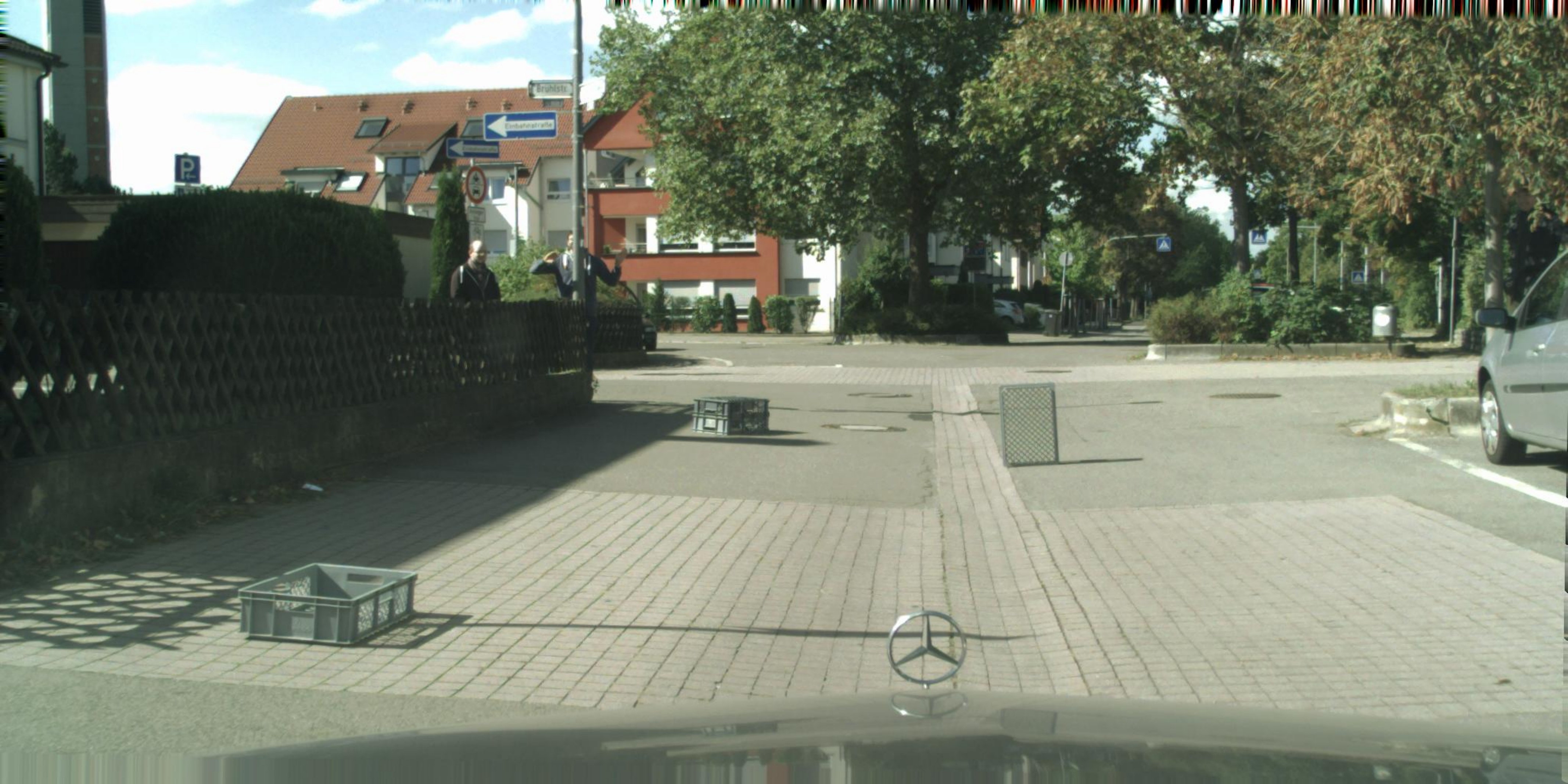} \hfill
    \includegraphics[width=.49\textwidth, height=2.5cm]{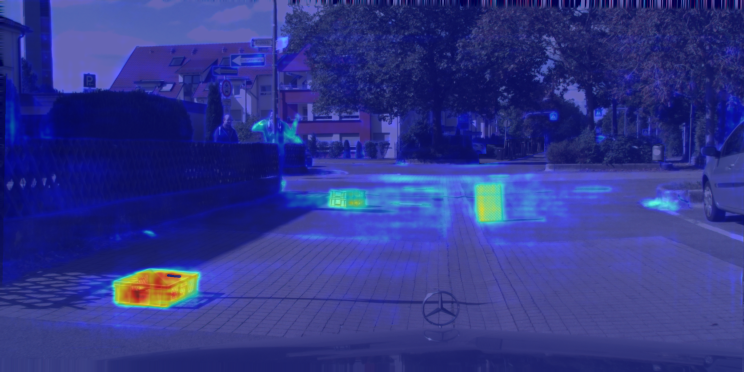}
    \includegraphics[width=.49\textwidth, height=2.5cm]{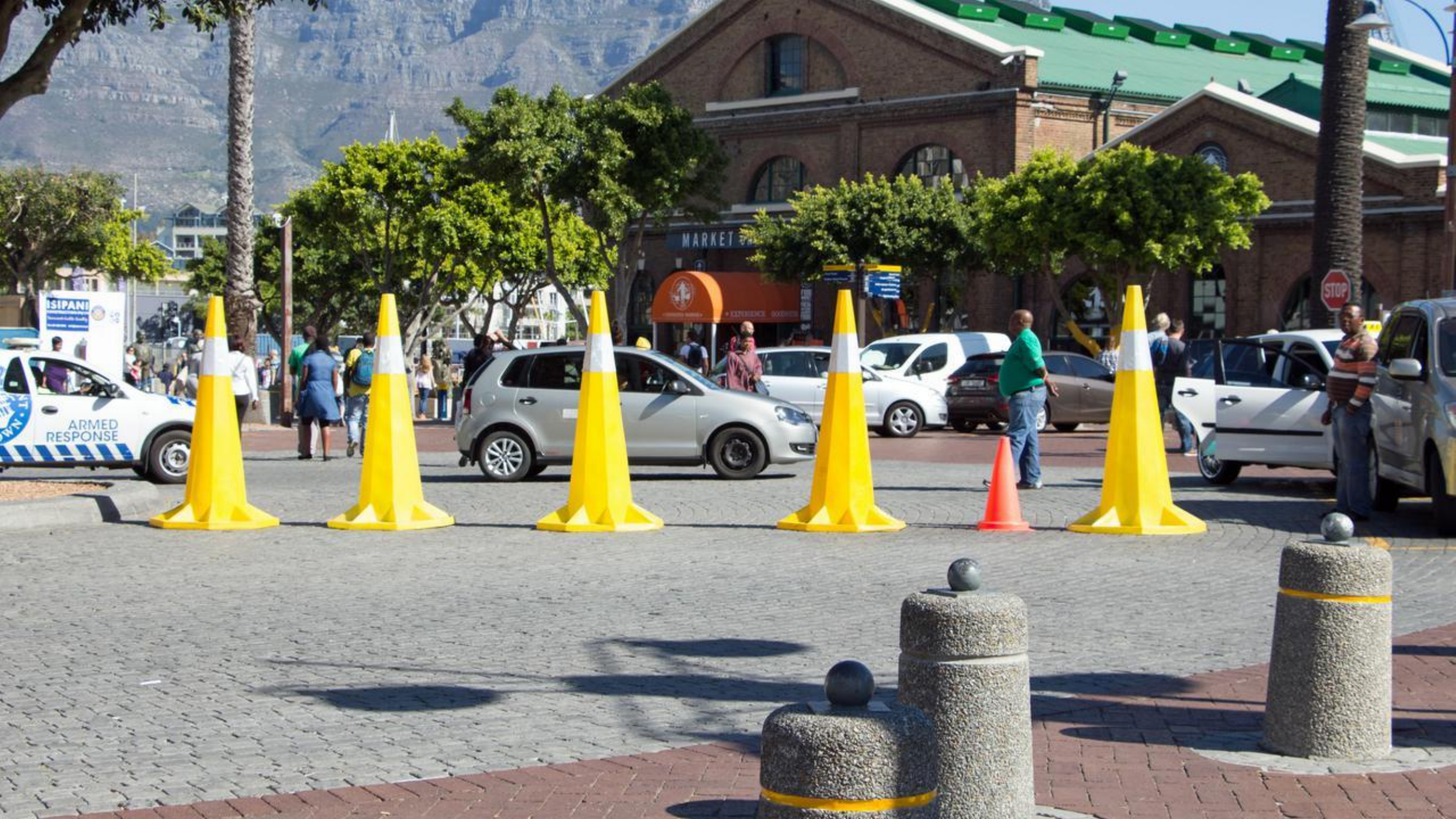} \hfill
    \includegraphics[width=.49\textwidth, height=2.5cm]{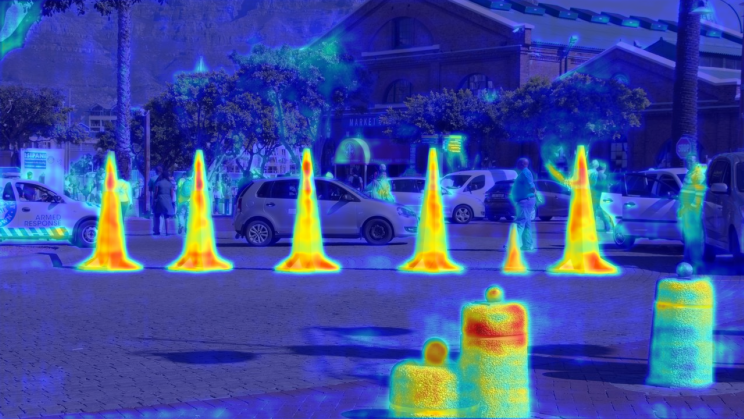}
    \includegraphics[width=.49\textwidth, height=2.5cm]{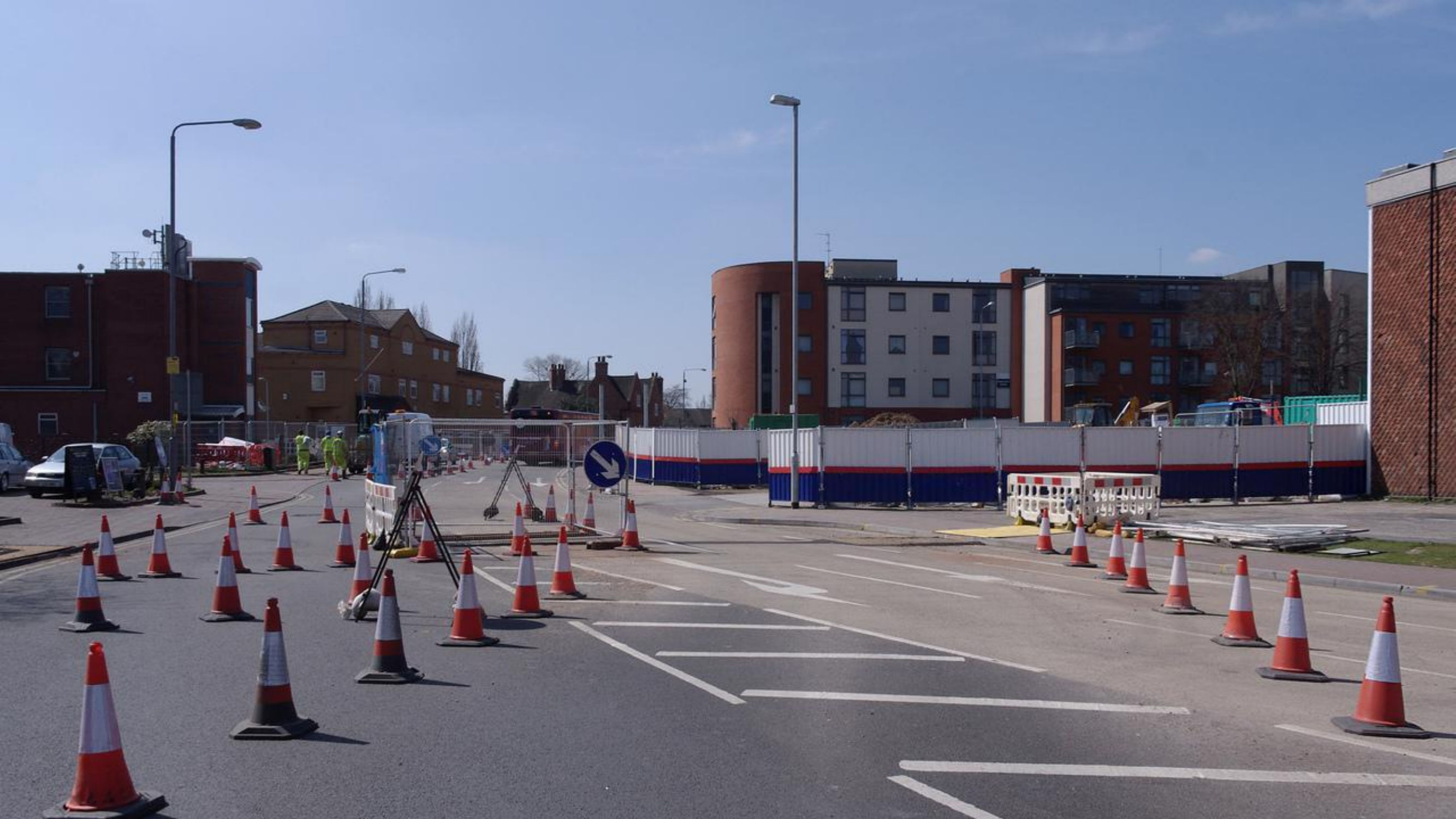} \hfill
    \includegraphics[width=.49\textwidth, height=2.5cm]{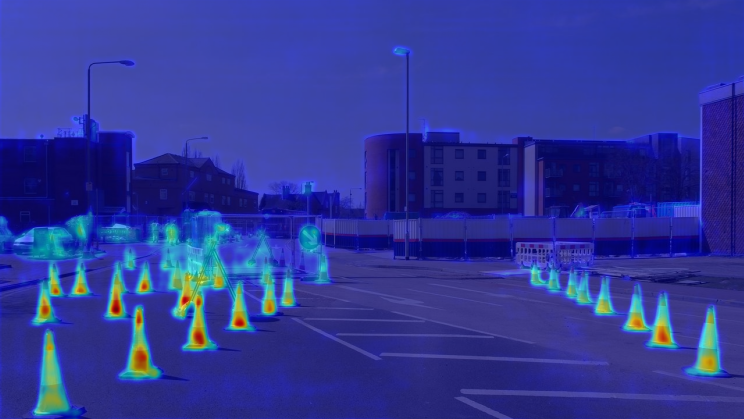}
    \caption{City Context Scenes}
    \label{fig:my_label}
\end{subfigure}
\begin{subfigure}[b]{0.49\linewidth}
    \centering
    \includegraphics[width=.49\textwidth, height=2.5cm]{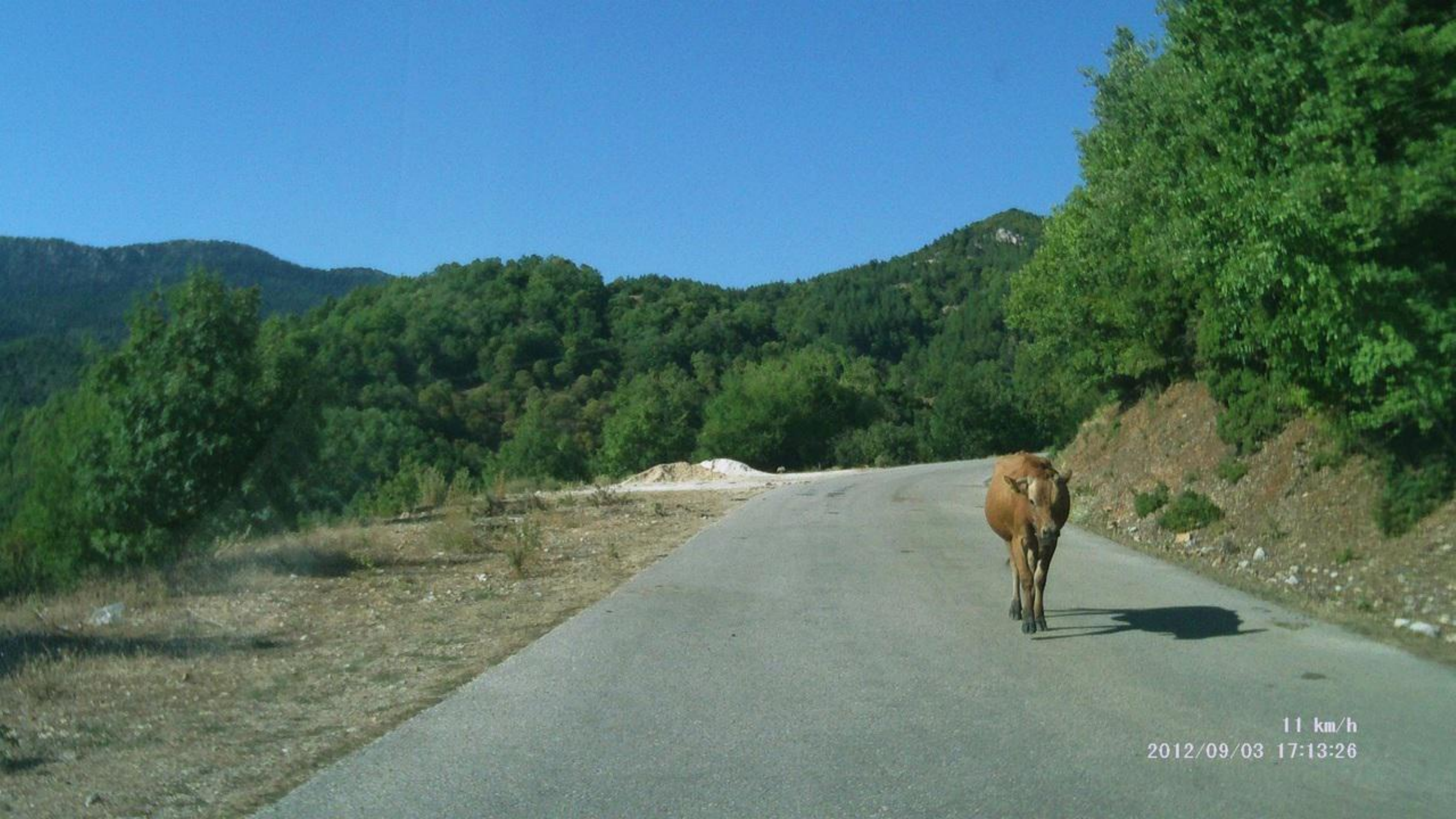} \hfill
    \includegraphics[width=.49\textwidth, height=2.5cm]{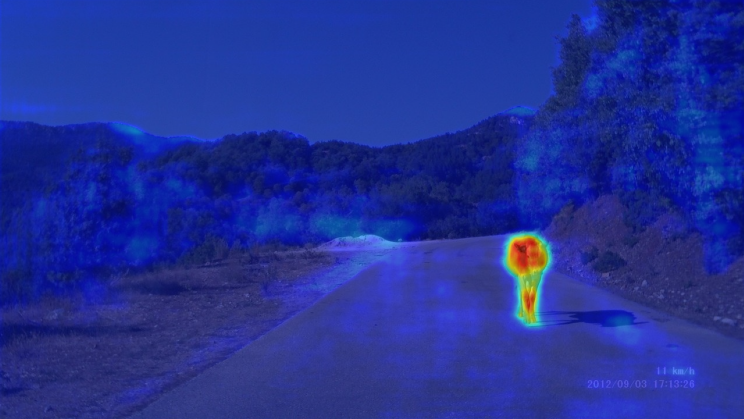}
    
    \includegraphics[width=.49\textwidth, height=2.5cm]{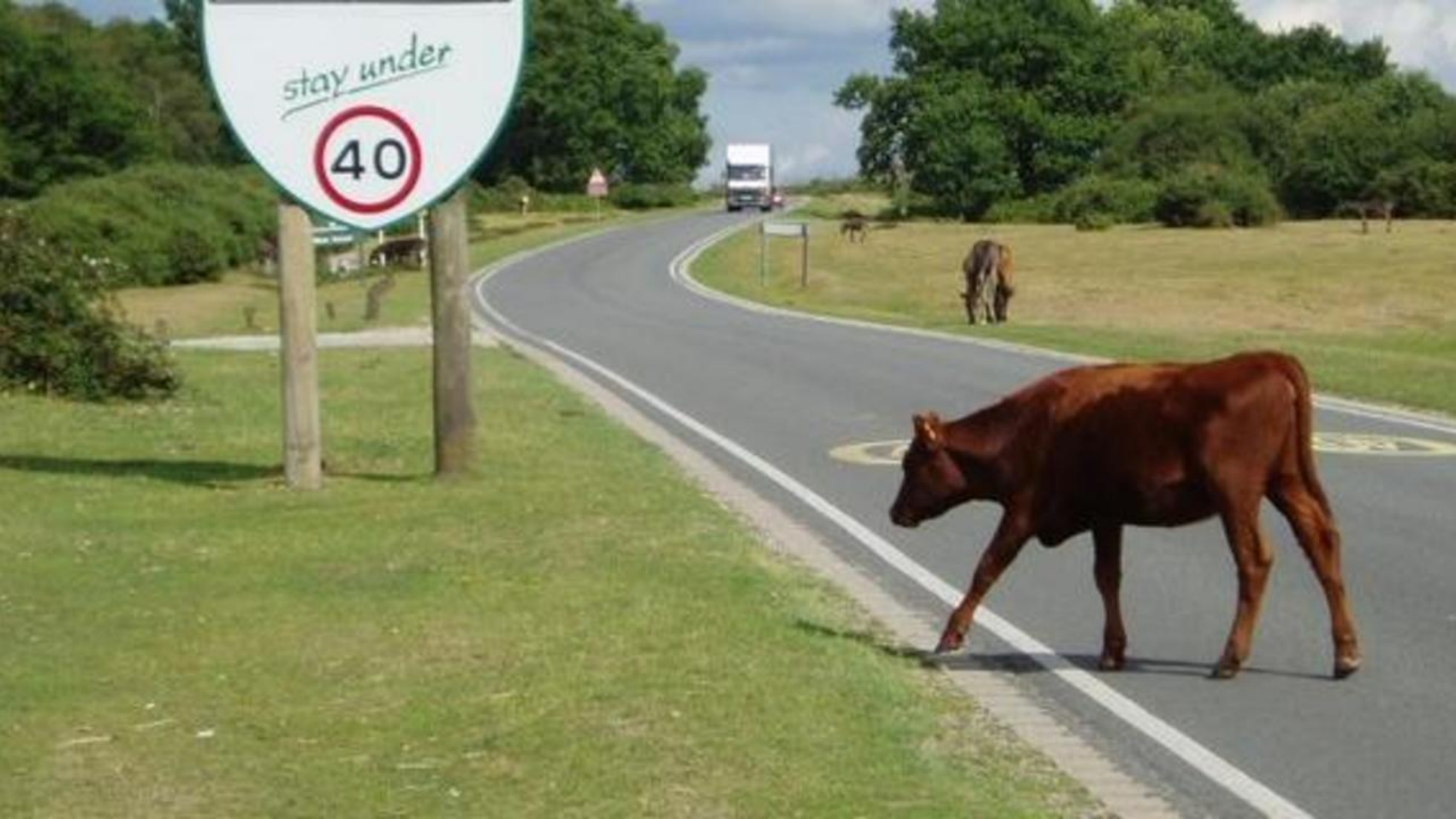} \hfill
    \includegraphics[width=.49\textwidth, height=2.5cm]{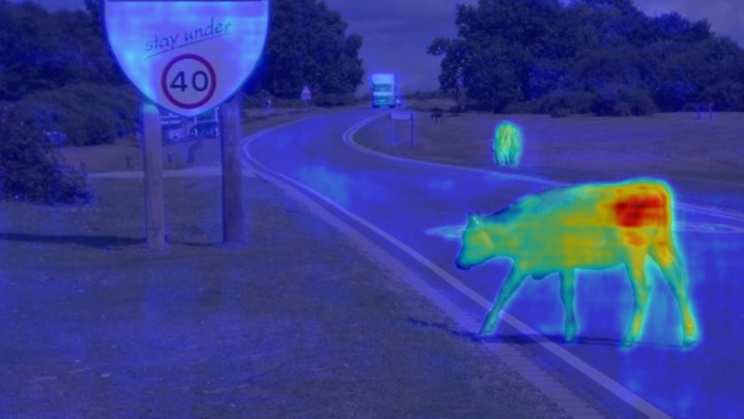}
    
    \includegraphics[width=.49\textwidth, height=2.5cm]{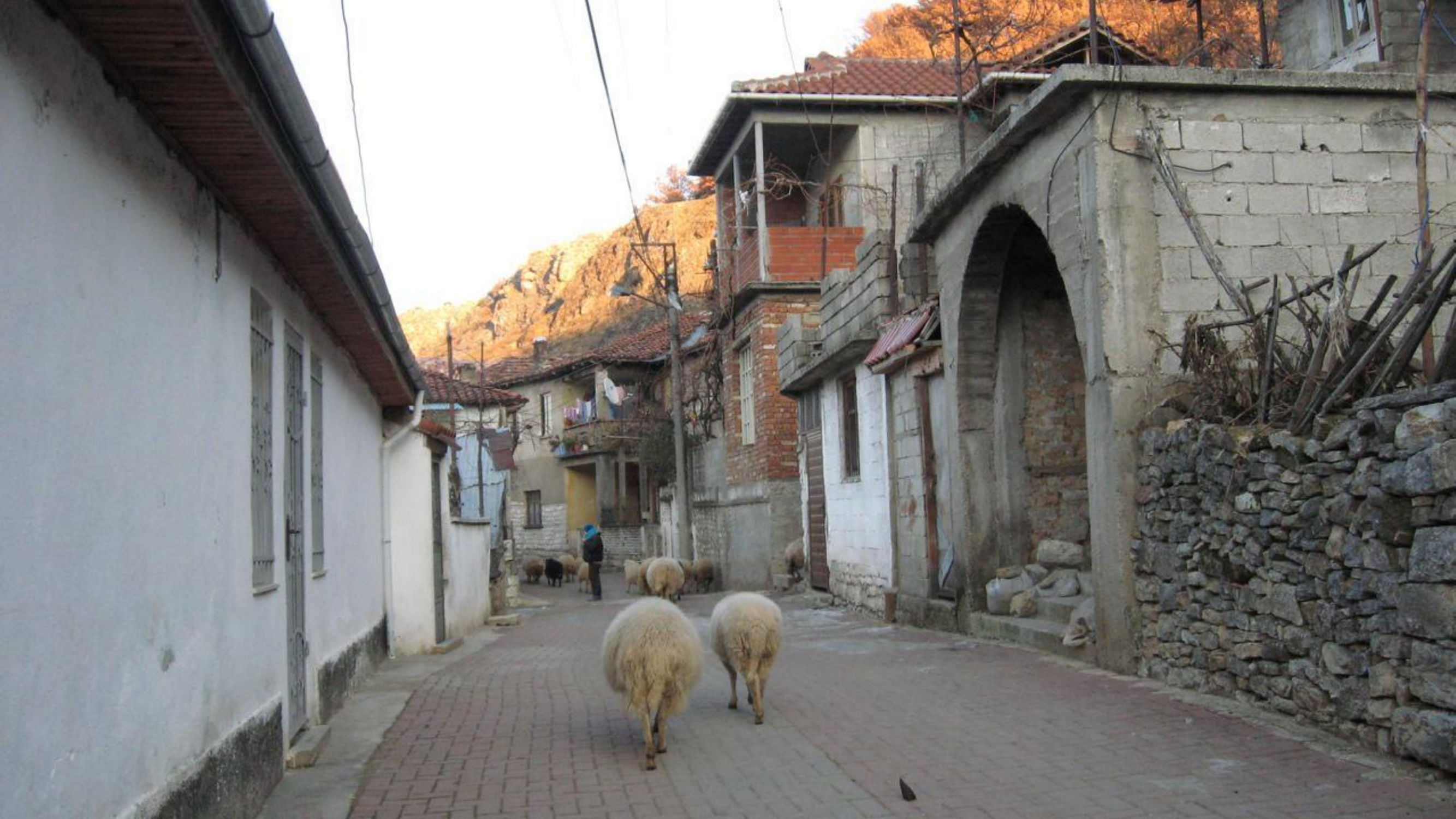} \hfill
    \includegraphics[width=.49\textwidth, height=2.5cm]{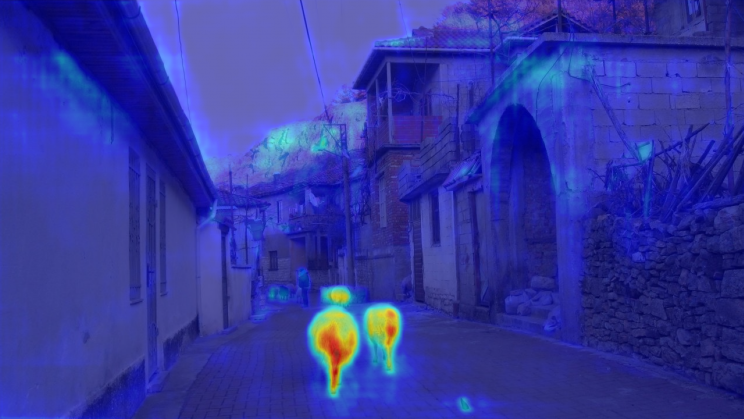}
    
    \includegraphics[width=.49\textwidth, height=2.5cm]{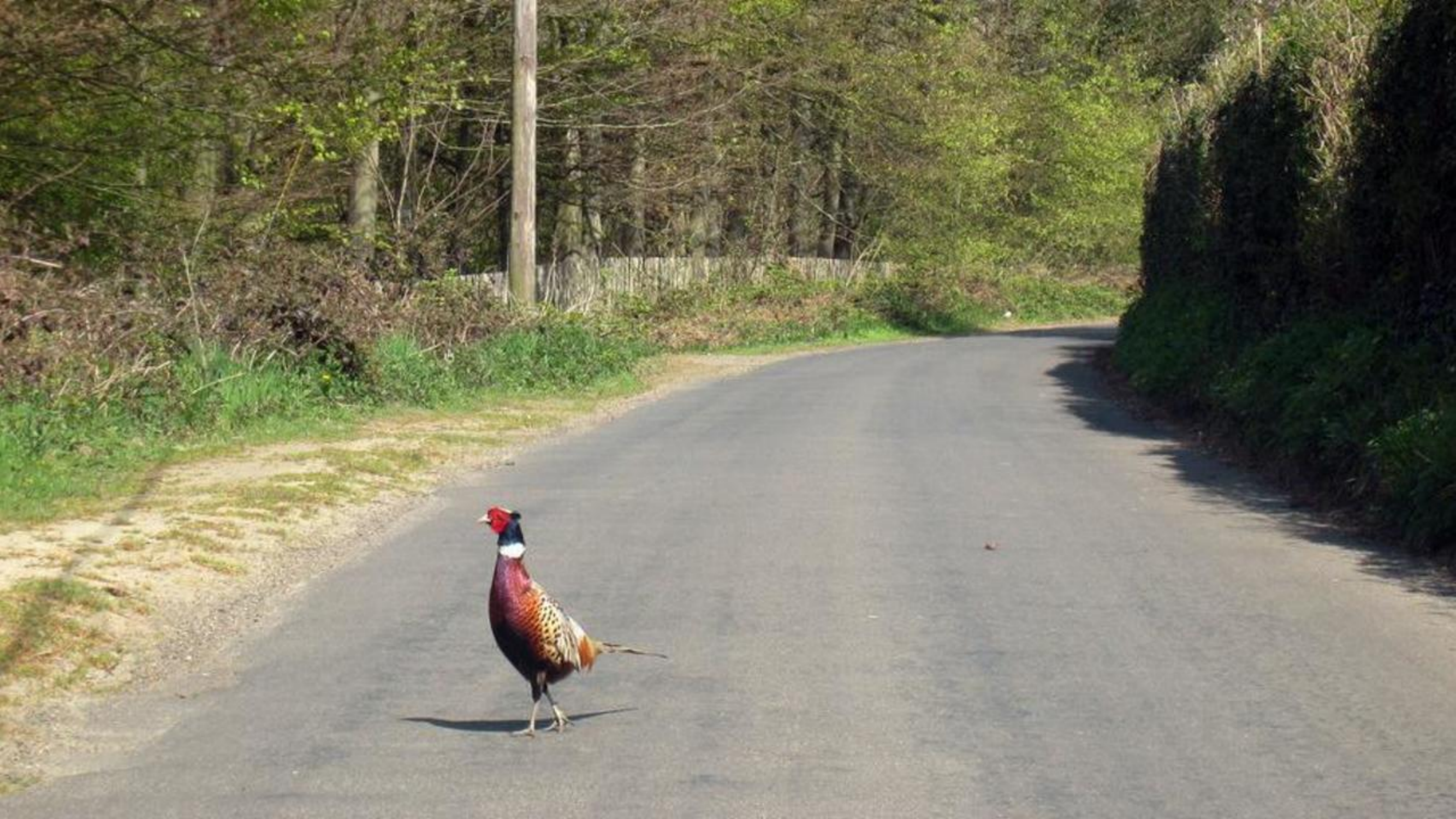} \hfill
    \includegraphics[width=.49\textwidth, height=2.5cm]{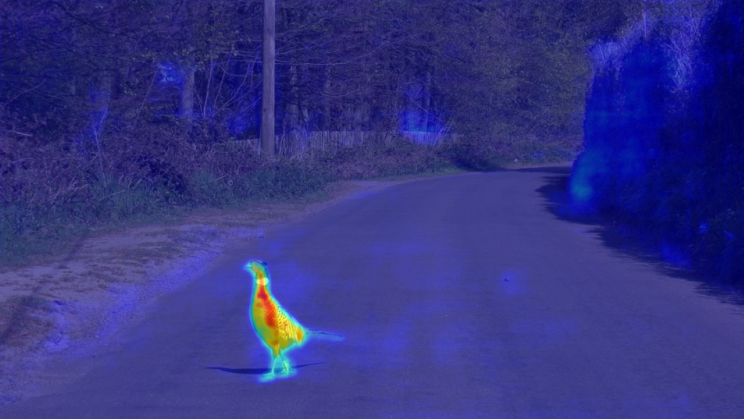}
    
    \includegraphics[width=.49\textwidth, height=2.5cm]{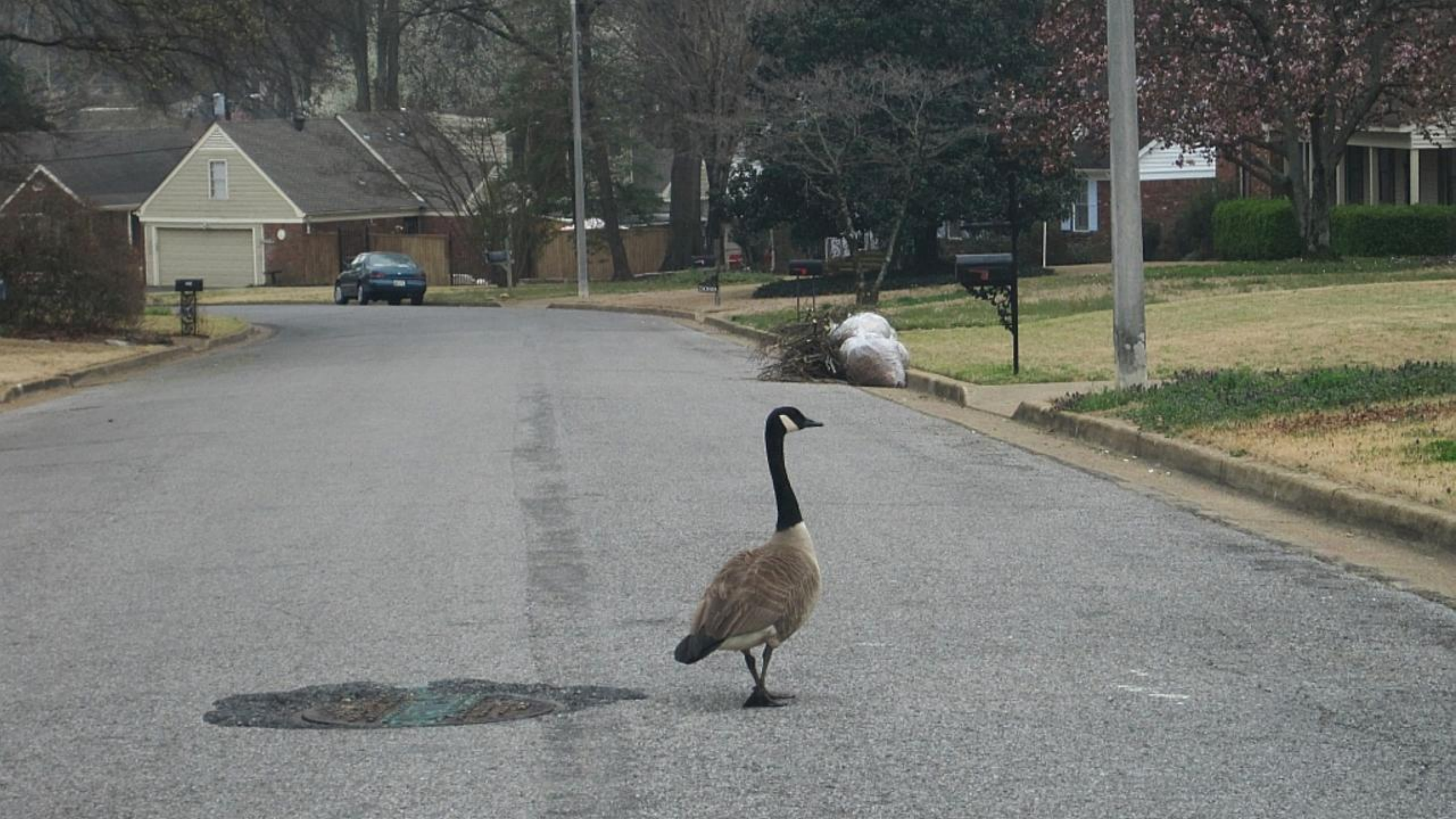} \hfill
    \includegraphics[width=.49\textwidth, height=2.5cm]{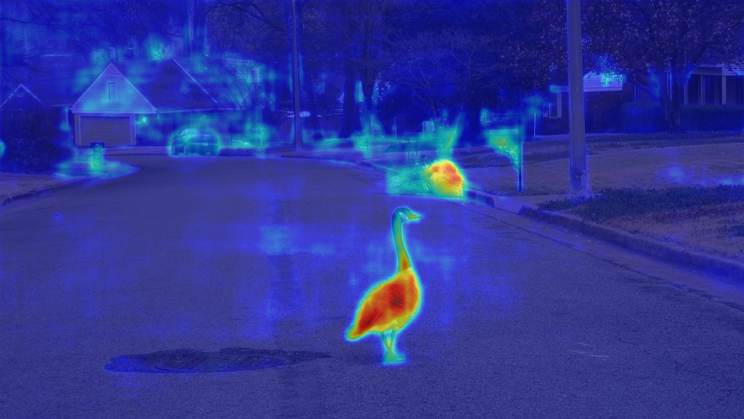}
    
    \includegraphics[width=.49\textwidth, height=2.5cm]{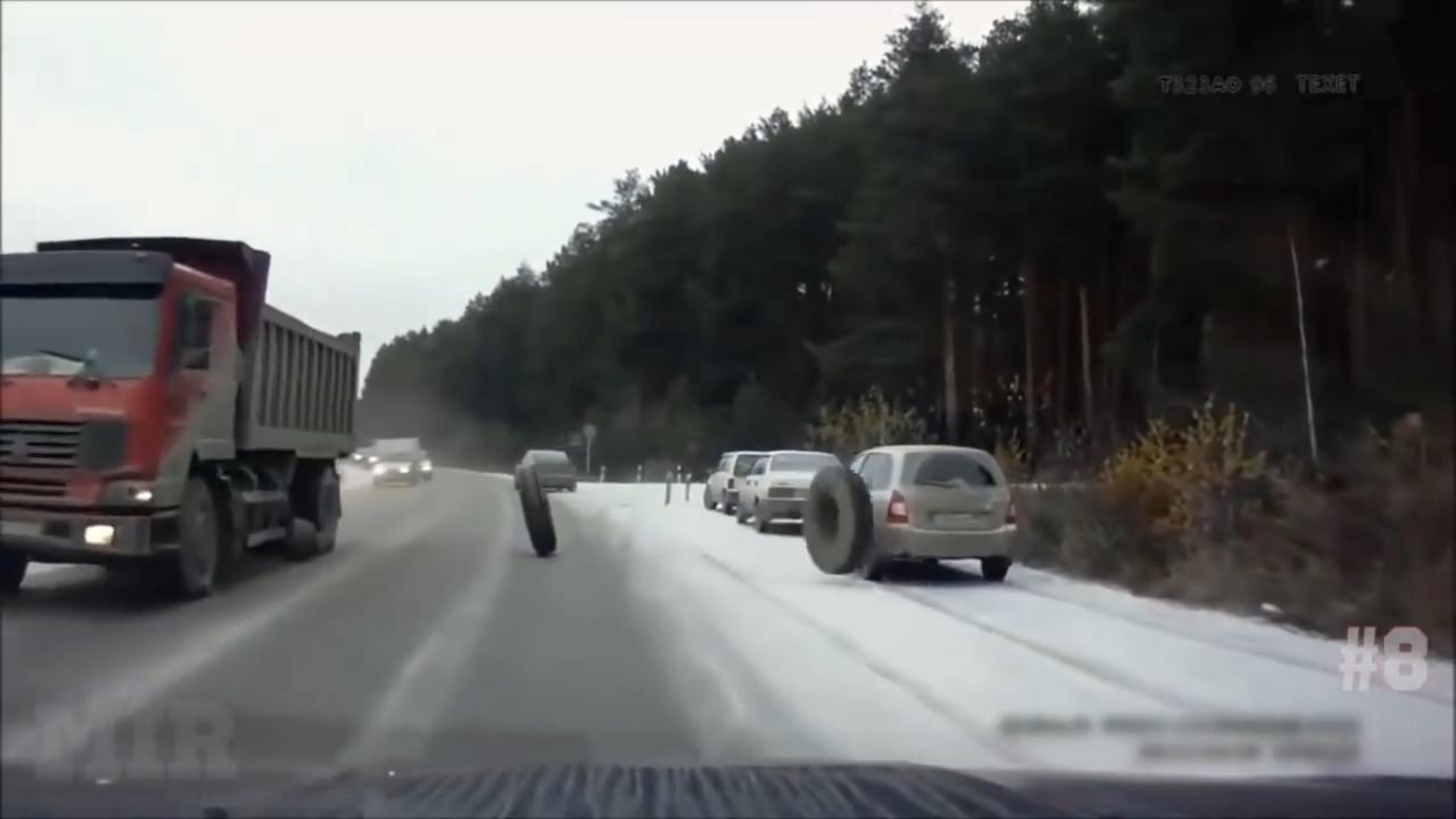} \hfill
    \includegraphics[width=.49\textwidth, height=2.5cm]{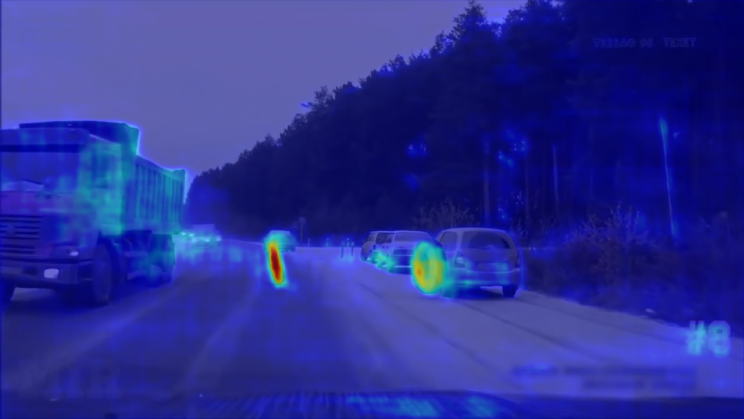}

    \includegraphics[width=.49\textwidth, height=2.5cm]{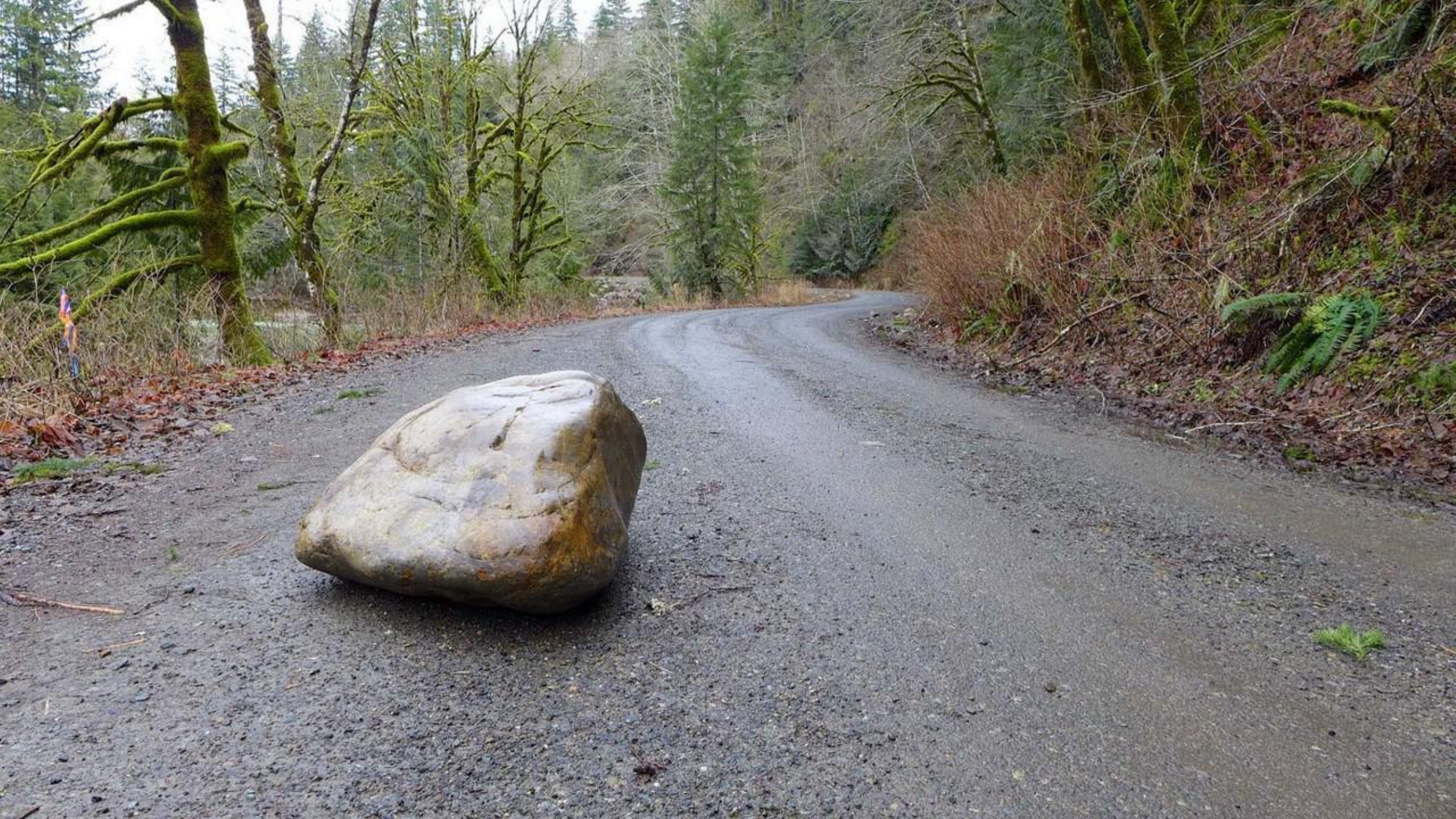} \hfill
    \includegraphics[width=.49\textwidth, height=2.5cm]{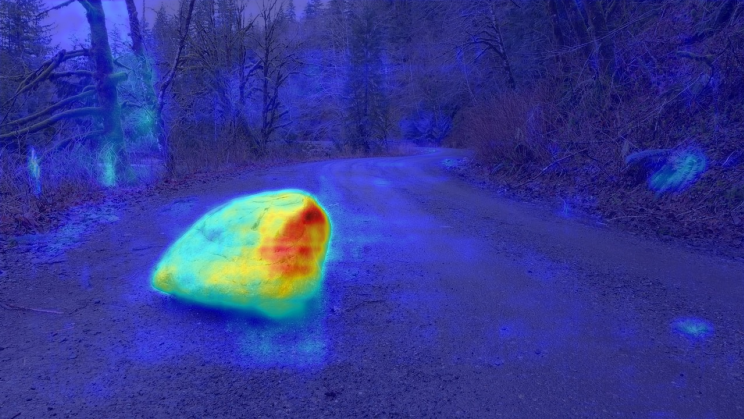}

    \includegraphics[width=.49\textwidth, height=2.5cm]{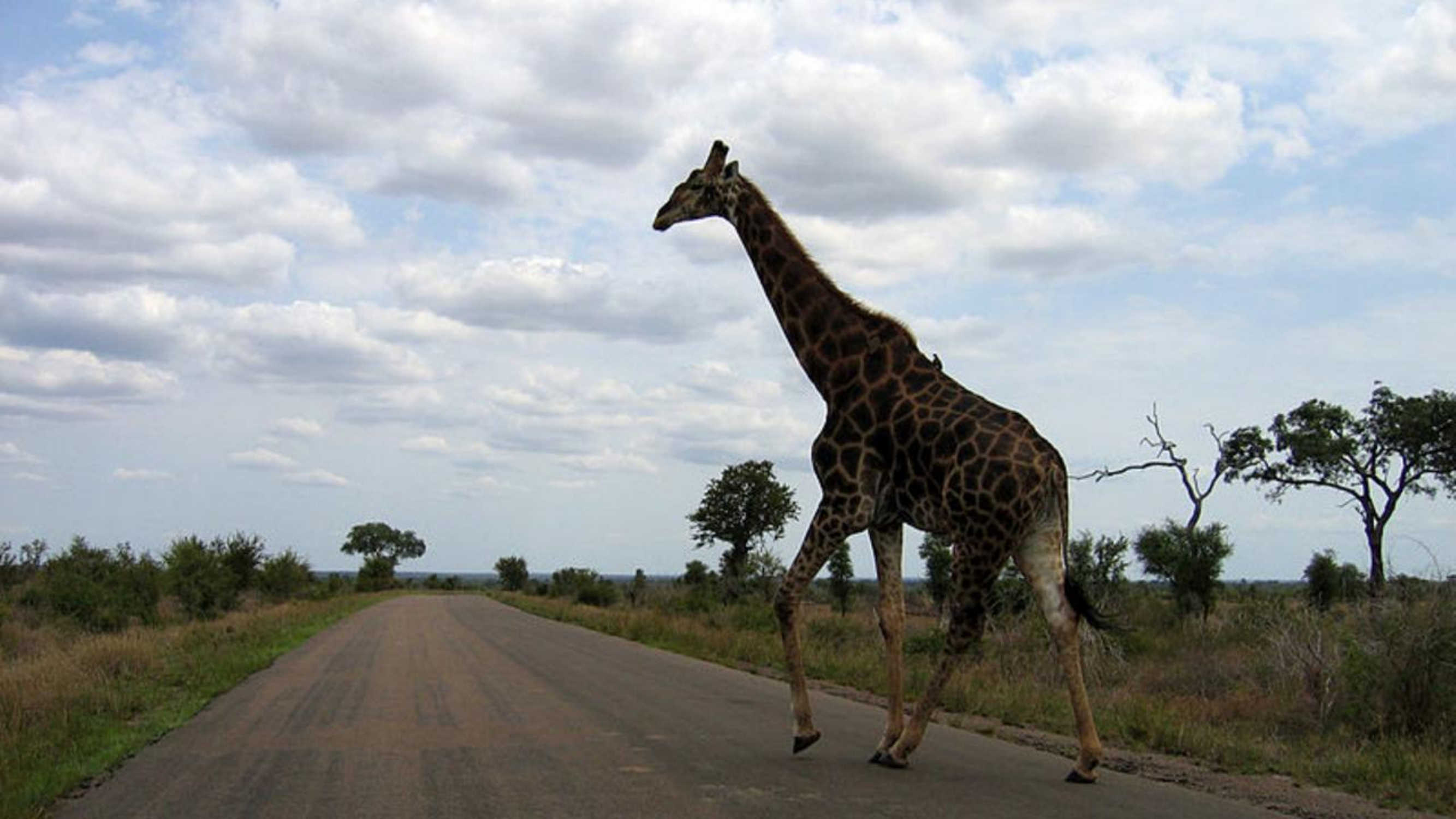} \hfill
    \includegraphics[width=.49\textwidth, height=2.5cm]{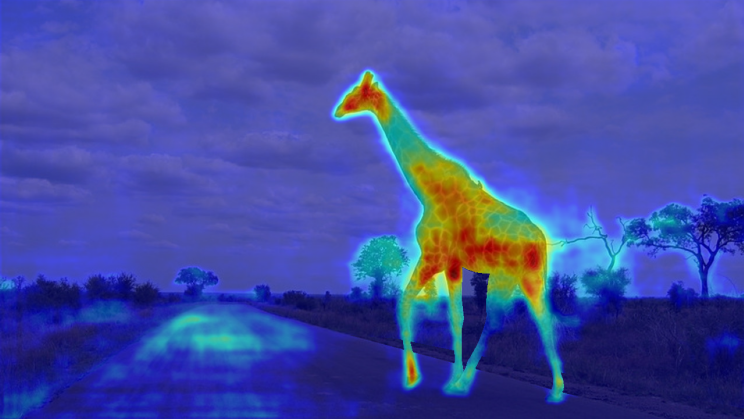}
    
    \includegraphics[width=.49\textwidth, height=2.5cm]{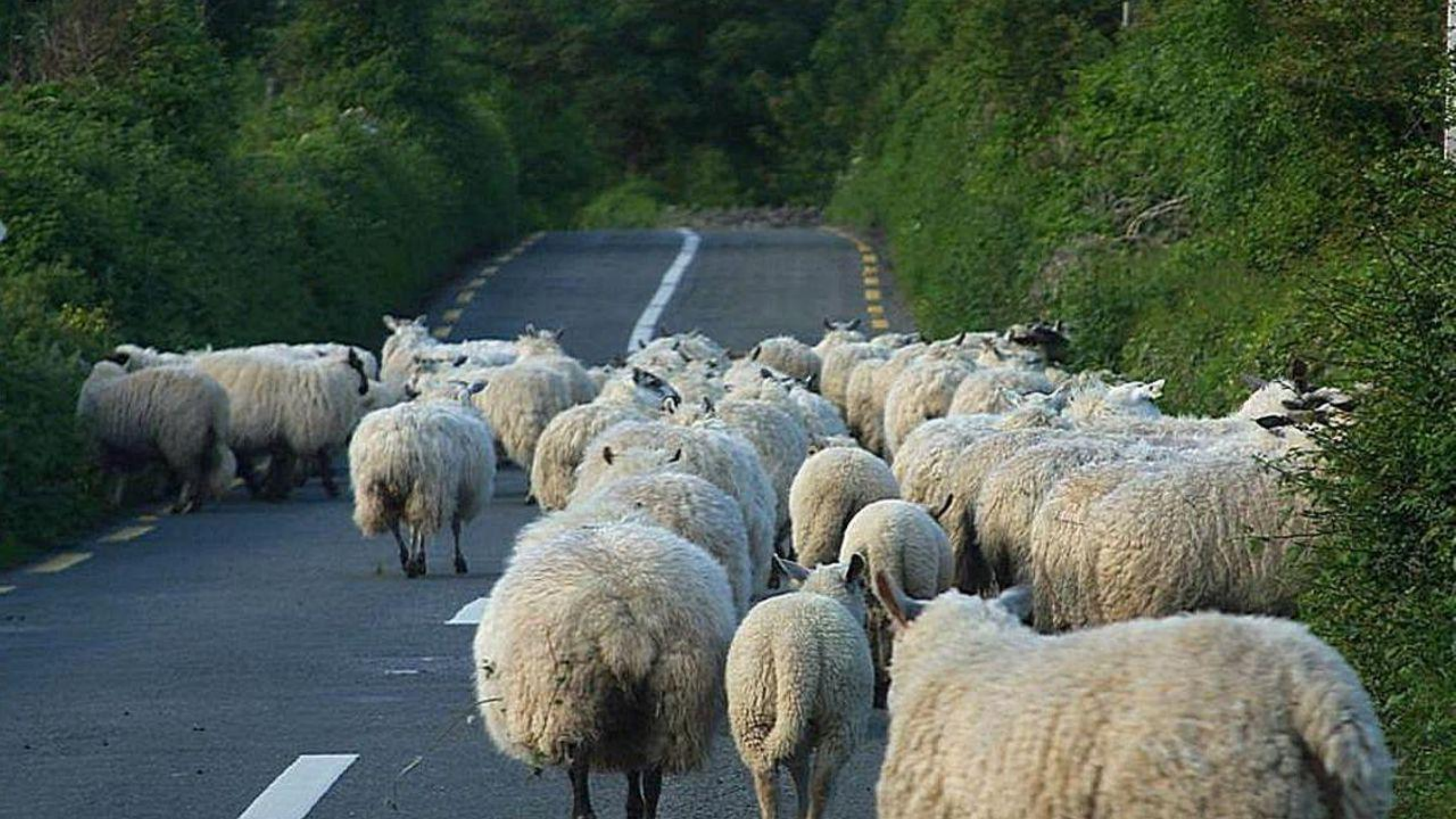} \hfill
    \includegraphics[width=.49\textwidth, height=2.5cm]{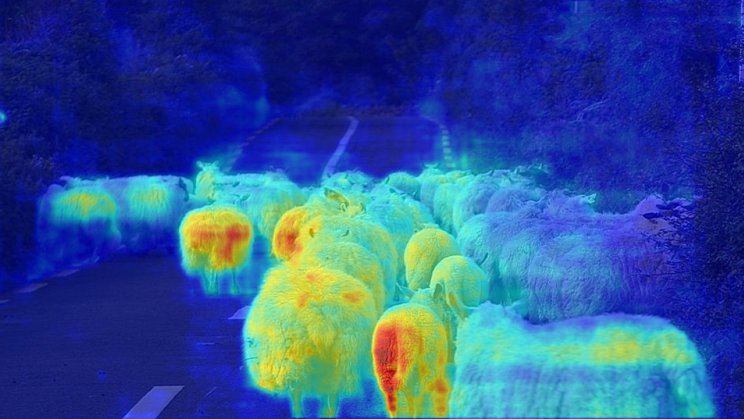}
    
    \caption{Country Context Scenes}
    \label{fig:my_label}
\end{subfigure}
\caption{\textbf{More visualisations} for our method in different contexts.}
\label{fig:sup_anomaly_vis}
\end{figure*}

\vspace{-3pt}
Fig. \ref{fig:sup_anomaly_vis} shows the anomaly segmentation visualisation results of our method. The results indicate that our method successfully detects and segments anomaly objects in different scenarios, including various hard anomalies. Rows 1-5 of the city scenes demonstrate that our method can detect small and distant anomalies well, while rows 8 and 9 show the robustness of our method to many anomalous objects, which are successfully detected and segmented. The country context scene results show that our method can accurately segment hard objects. Rows 1-5 and 8 also show that our method accurately segments OoD animals. Row 9 of the country scene is a hard challenge for anomaly segmentation because most of the pixels in the image are anomalies, and it is difficult for previous methods to identify such large-scale anomalies. Nevertheless, our method can still detect the anomalies on the road in this hard case.


\clearpage

\end{document}